%% file: main.tex
\documentclass{article}
\PassOptionsToPackage{numbers}{natbib}

\usepackage[main, final]{neurips_2026}

\usepackage{amsmath}
\usepackage{amssymb}
\usepackage{amsthm}
\usepackage{graphicx}
\usepackage{placeins}
\usepackage{xcolor}

\definecolor{goalred}{HTML}{DD2222}
\definecolor{startgreen}{HTML}{22CC22}

\newcommand{\red}[1]{\textcolor{goalred}{#1}}
\newcommand{\green}[1]{\textcolor{startgreen}{#1}}

\usepackage[pagebackref=true]{hyperref} 
\renewcommand*{\backref}[1]{} 
\renewcommand*{\backrefalt}[4]{%
  \ifcase #1 
    (Not cited.)%
  \or 
    (Cited on page #2.)%
  \else 
    (Cited on pages #2.)%
  \fi%
}

\usepackage{booktabs} 
\usepackage{enumitem}
\usepackage{wrapfig}

\usepackage{tikz}
\usetikzlibrary{arrows.meta,positioning,decorations.pathmorphing,calc,backgrounds,fit,shapes,decorations.markings}

\usepackage[breakable, skins, theorems]{tcolorbox}

\definecolor{maincolor}{RGB}{0,113,178}
\definecolor{mainbg}{RGB}{232,243,252}

\newcounter{thmcounter}
\renewcommand{\thethmcounter}{\arabic{thmcounter}}

\makeatletter
\newcommand{\thmlabel}[1]{\protected@edef\@currentlabel{\thethmcounter}\label{#1}}
\makeatother

\tcbset{
    boxstyle/.style={
        enhanced, breakable,
        colback=mainbg, colframe=maincolor, coltitle=white,
        boxrule=0.8pt, arc=2pt,
        fonttitle=\bfseries,
        before upper={\setlength{\parskip}{4pt}},
        top=4pt, bottom=4pt, left=6pt, right=6pt,
    }
}

\newtcolorbox[use counter=thmcounter, number within=]{theorembox}[1][]{%
    boxstyle,
    title={Theorem~\arabic{thmcounter}\ifx\\#1\\\else\ (#1)\fi},
}

\newtcolorbox[use counter=thmcounter, number within=]{corollarybox}[1][]{%
    boxstyle,
    title={Corollary~\arabic{thmcounter}\ifx\\#1\\\else\ (#1)\fi},
}

\newtcolorbox{theoremrestate}[1][]{%
    boxstyle,
    title={#1},
}

\newtcolorbox{assumptionbox}[1][]{%
    boxstyle,
    title={Assumptions\ifx\\#1\\\else\ (#1)\fi},
}

\newtcolorbox{mainresultbox}[1][]{%
    boxstyle, boxrule=1.2pt,
    title={Main Result\ifx\\#1\\\else\ (#1)\fi},
}

\title{When Does LeJEPA Learn a World Model?}

\author{%
  David Klindt \\
  Cold Spring Harbor Laboratory\\
  \texttt{klindt@cshl.edu}
  \And
  Yann LeCun \\
  New York University\\
  \texttt{yann.lecun@nyu.edu}
  \And
  Randall Balestriero\\
  Brown University\\
  \texttt{randall\_balestriero@brown.edu}
}

\begin{document}

\maketitle

\vspace{-10pt}
\begin{abstract}
\vspace{-10pt}
A representation that scrambles the true degrees of freedom of the world cannot support reliable planning or compositional generalization. We prove that LeJEPA (alignment plus Gaussian regularization) linearly recovers the world's latent variables from nonlinear observations, a property known as \textit{linear identifiability}, in a class of worlds where latents evolve under stationary, additive-noise transitions. Our main result is that among all such worlds, the Gaussian is the \textit{unique} latent distribution for which this guarantee holds. The forward direction rests on a spectral decomposition in which each degree of nonlinearity is strictly penalized by alignment, making the linear map the optimum; the converse rules out every non-Gaussian alternative. We further prove an \textit{approximate identifiability} result where the guarantee degrades gracefully, and show that linear, orthogonal identifiability enables \textit{optimal latent-space planning}. We validate the theory with experiments ranging from 2D examples to 4096-dimensional latents, including ablations, pixel-based robotic control, and natural images from ImageNet-X. Our theory turns an empirically successful recipe into a mathematical guarantee, providing the foundation for building World Models that provably recover the structure of the world.
\end{abstract}

\vspace{-15pt}
\section{Introduction}
\vspace{-10pt}
The promise of self-supervised learning (SSL) is that we can learn useful representations of the world without labeled data, just by observing and predicting. Joint-Embedding Predictive Architectures \citep[JEPAs,][]{lecun2022} pursue this vision by training a representation to produce similar embeddings for related views of the same input, while a regularizer prevents the representation from collapsing to a trivial constant \citep{balestriero2025lejepa, bardes2022}. The resulting representations have proven remarkably effective across image~\citep{assran2023ijepa}, video~\citep{bardes2024vjepa, assran2025vjepa2}, and latent-space planning~\citep{sobal2025pldm, zhou2025dinowm, maes2026leworldmodel}. However, a deeper question remains:
\emph{When is a learned representation a World Model, i.e., a faithful map of the world's latent structure?}

\begin{figure}[h!]
    \centering
    \includegraphics[width=0.9\textwidth]{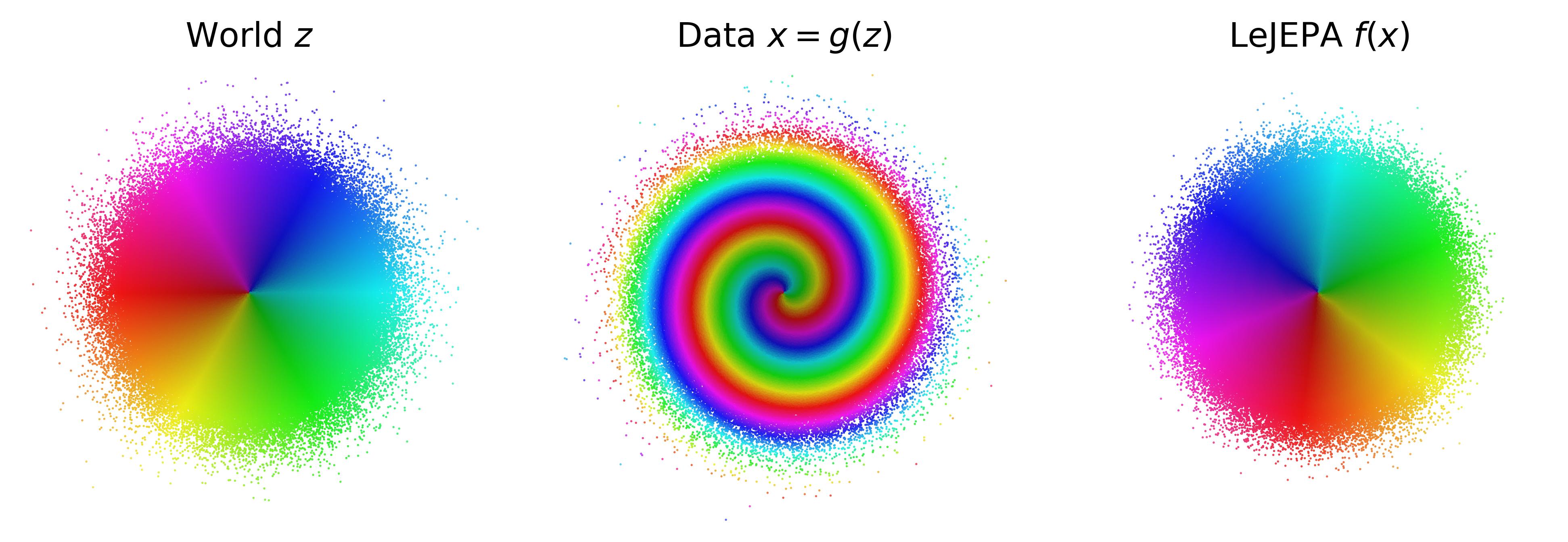}
    \vspace{-7pt}
    \caption{
        \textbf{LeJEPA learns the World Model.}
        \textbf{(left)} A world with independent Gaussian \textit{latent variables}.
        \textbf{(center)} An unknown nonlinear process scrambles them into the \textit{data} we observe.
        \textbf{(right)} LeJEPA \citep{balestriero2025lejepa} recovers the latent variables up to rotation. We prove this is the \textit{unique} optimum.
    }
    \label{fig:teaser}
    \vspace{-7pt}
\end{figure}

Our answer: \textit{When it linearly recovers the world's latent variables} (Fig.~\ref{fig:teaser}). A representation that entangles unrelated latent variables, scrambling an object's position with its color, or mixing velocity with texture, may score well on narrow tasks but will fail when the world changes \citep{scholkopf2021toward}. What we need is \emph{linear identifiability}, i.e., a mathematical guarantee that the learned representation recovers the underlying latent variables, up to simple symmetries \citep{hyvarinen2024identifiability}. Practitioners already test for this routinely. Every time a representation is evaluated via \textit{linear probing} \citep{alain2016understanding}, the implicit question is whether the model has learned a \textit{linear representation} of the latent variables \citep{rumelhart_model_1973, hinton1986learning, smolensky_tensor_1990, arora_linear_2018, park_linear_2024, klindt2025superposition}.
Without this, linear probes cannot recover latent variables exactly. Linear identifiability is thus a \textit{necessary}, albeit not \textit{sufficient} \citep{pacela2026stopprobingstartcoding}, condition for faithful linear probing.

Yet no identifiability results exist for JEPAs. This has remained out of reach because prior methods rely on implicit collapse prevention \citep{chen2021exploring, grill2020, caron2021emerging} with unspecified embedding distributions.
A recent proposal changes the picture: LeJEPA \citep{balestriero2025lejepa} prevents collapse by explicitly regularizing the embedding distribution toward an isotropic Gaussian via Sketched Isotropic Gaussian Regularization (SIGReg). 
Along with an alignment loss this enables stable end-to-end training from raw pixels \citep{maes2026leworldmodel}, but whether the resulting representation actually recovers the world's latent variables still remains open.

\begin{figure}[t]
    \centering
    \input{figures/dgp.tex}
    \caption{
        \textbf{LeJEPA Theory Illustration.} 
        \textbf{(left)} The world has clean latent structure (Gaussian, disentangled) with correlated positive pairs. 
        \textbf{(center)} An unknown nonlinear process produces the observations we actually see, scrambling the latent structure. 
        \textbf{(right)} LeJEPA trains a representation with two objectives: pull positive pairs together (\textit{attract}) and keep the embedding distribution Gaussian (SIGReg) to prevent collapse. We prove that the learned representation must be a rotation of the true latents, i.e., the representation is forced to learn the correct World Model (Thm.~\ref{thm:main}).
    }
    \label{fig:overview}
    \vspace{-10pt}
\end{figure}
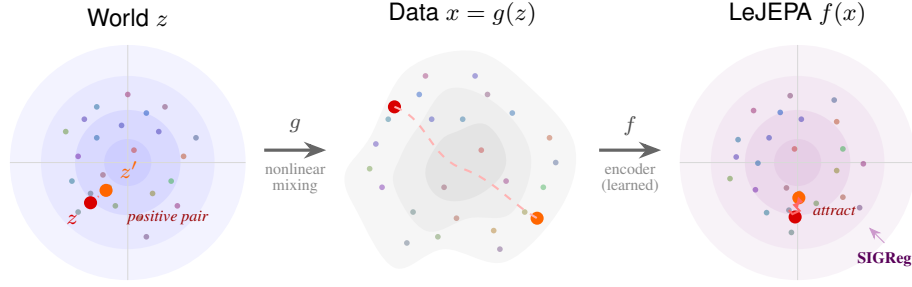

\paragraph{Contributions.}
We close this gap with the first identifiability result for JEPAs (Fig.~\ref{fig:overview}).
We consider a class of worlds with Gaussian latent variables and positive pairs from a stationary process with independent, additive-noise transitions, which are maximum-entropy choices \citep{jaynes1957information}. We prove that LeJEPA learns a linearly identifiable representation \textit{if and only if} the latent variables are \textit{Gaussian}. The forward direction rests on a spectral decomposition that shows how every degree of nonlinearity is penalized; the converse rules out every non-Gaussian alternative. We further prove that identifiability degrades gracefully when the objectives are only \textit{approximately} satisfied, and that linear identifiability suffices for \textit{optimal latent-space planning}. We validate empirically across 2D mixings, up to 4096-dim latents, distributional ablations, pixel-based robotic control, and the approximate bound, and probe beyond the assumptions with a misspecified latent dimension, learned predictors \citep{maes2026leworldmodel}, and natural images from ImageNet-X \citep{idrissi2023imagenetx} (Sec.~\ref{sec:experiments}).\footnote{Code, Lean proofs, and demo: \url{https://github.com/klindtlab/lejepa-identifiability}.}

\section{Related Work}
\label{sec:related}

\paragraph{Representation Learning.}
JEPAs \citep{lecun2022, assran2023ijepa, bardes2024vjepa, assran2025vjepa2} predict in representation space rather than pixel space, avoiding capacity wasted on irrelevant detail. More generally, SSL pulls positive views of the same content together while preventing collapse. Contrastive methods \citep{gutmann2010noise, chen2020simple, he2020momentum}, such as InfoNCE \citep[][]{oord2018representation}, use negatives explicitly; non-contrastive methods substitute stop-gradient teachers \citep{grill2020, chen2021exploring}, covariance regularization \citep[VICReg,][]{bardes2022, zbontar2021}, or self-distillation with feature clustering \citep{caron2021emerging, simeoni2025dinov3}. LeJEPA \citep{balestriero2025lejepa} adds an explicit Gaussianity regularizer (SIGReg); LeWorldModel \citep{maes2026leworldmodel} scales the recipe to action-conditioned control. InfoNCE, VICReg, and LeJEPA span a hierarchy of `Gaussianity' constraints, from implicit \citep{wang2020understanding, li2023rethinking, eftekhari2025importance, betser2026infonce} through second-moment to full; we test all three (Tab.~\ref{tab:scaling-comparison}, App.~\ref{app:gennorm}).

\paragraph{World Models.} The internal-model concept spans cognitive science, control theory, and modern ML. Cognitive scientists framed mental simulation as the substrate of reasoning, perception, and motor control \citep{craik1943,tolman1948,johnson_laird_1983,wolpert1995internal,gregory1980perceptions}, with free-energy \citep{friston2010free} and probabilistic-program \citep{lake2015human} formalizations of the brain as a generative model. Cybernetics extended this to engineered systems \citep{conant1970good,francis1976}, while classical control developed dynamics on a given state space \citep{bellman1957,kalman1960}. Neural world models progressed from recurrent controller-model pairs \citep{nguyen1990neural,schmidhuber1990making,sutton1990dyna} through latent dynamics from pixels \citep{watter2015embed,oh2015action,finn2016unsupervised,ha2018world,hafner2019planet}, value-equivalent prediction \citep{schrittwieser2020mastering}, generative \citep{hafner2025dreamerv3} and joint-embedding predictive \citep{lecun2022,maes2026leworldmodel} architectures, to large generative video simulators \citep{brooks2024sora,bruce2024genie}. The JEPA argument is that pixel-perfect prediction wastes capacity \citep{lecun2022}. We address the encoder side of this picture; classical control applies in the learned coordinates (Thm.~\ref{thm:planning}).

\paragraph{Identifiability.} Nonlinear ICA is unidentifiable without additional structure \citep{hyvarinen1999,locatello2019}. Identifiability becomes possible when the world supplies it: non-stationarity \citep{hyvarinen2016}, temporal dependence \citep{hyvarinen2017,klindt2021}, auxiliary variables \citep{khemakhem2020,hyvarinen_nonlinear_2019}, contrastive learning \citep{zimmermann2021}, augmentation \citep{vonkugelgen2021}, mechanism sparsity \citep{lachapelle2022}, interventions \citep{ahuja2023,buchholz2024}, or supervision \citep{reizinger2024cross,roeder2021linear}. These results typically constrain the representation to a smooth diffeomorphism; our Hermite approach works for arbitrary measurable maps. Most closely related is slow feature analysis (SFA) \citep{wiskott2002}, which JEPA objectives empirically recover \citep{sobal2022slow}; App.~\ref{app:prior-work} contrasts in depth with SFA theory \citep{sprekeler2014}. 
Similar spectral analysis in representation space has previously characterized other SSL objectives \citep{balestriero2022contrastivenoncontrastiveselfsupervisedlearning}, and our Thm.~\ref{thm:approx} connects to recent quantitative stability work \citep{buchholz2025robustness,nielsen2025does}. The common lesson is that identifiability is always a joint statement about the \textit{World} (`data generating process') and the \textit{Learner} (`learning objective', LeJEPA).

\section{The World and the Learner}\label{sec:setup}

What does it mean to \textit{Learn the World Model?}
Suppose the world has latent variables $z \in \mathbb{R}^N$,\footnote{To help readers, all notation is summarized in the Glossary (App.~\ref{glossary}).} for instance, position, velocity, color or lighting.
These degrees of freedom are called \emph{latent variables} / \textit{sources} in the ICA literature, or \emph{factors of variation} \citep{bengio2013representation} in representation learning.
We never observe $z$ directly.
Instead, an unknown process $g$ generates the data / observations $x = g(z)$.\footnote{More philosophically, we only ever observe the \textit{shadows} \((x)\) of \textit{reality} \((z)\), \textit{projected} \((g)\) onto the walls of \textit{Plato}'s cave.}
Think of $g$ as rendering a 3D scene into an image, or mapping physical states to sensor readings.
The process $g$ can be highly nonlinear, it scrambles the clean latent structure into complicated, entangled observations.
We train a representation $f$ that maps observations back to representations.
The ideal outcome is that $f$ undoes $g$: the composed map $h = f \circ g$ should recover the original latent variables $z$.
Of course, perfect recovery is too much to hope for, there are symmetries that cannot be resolved like the rotation invariance of a Gaussian.
We will show that $h$ must be a \emph{linear function} of the true latents: $h(z) = Qz$.
This is a necessary condition for linear probes to work, and it is what lets planning in the learned latent space cost the same as planning in the true latent space (Thm.~\ref{thm:planning}).

\subsection{The World}\label{sec:world-model}
The world specifies a joint distribution $p(z, z')$ over positive pairs $(g(z), g(z'))$.
In SSL, these are views of the same underlying content, e.g., two frames of a video, two augmentations of an image, two nearby time steps of a trajectory. 
Throughout, $z_i$ denotes the $i$-th coordinate of $z \in \mathbb{R}^N$, i.e., one scalar latent variable, and $z'$ the second view. 
A class of worlds is defined by three assumptions:

\begin{assumptionbox}[World]\label{assumptions}
\begin{enumerate}[label=(\roman*), leftmargin=32pt, itemsep=2pt]
    \item \textbf{Independence.}
    $p(z_i) \perp p(z_j)$ and transitions $p(z'_i \mid z_i) \perp p(z'_j \mid z_j)$ for all $i \neq j$.

    \item \textbf{Stationarity.}
    Both views share the same marginal: $p(z) = p(z')$.

    \item \textbf{Additive noise.} $z'_i = m_i(z_i) + \eta_i$ with $\eta_i$ independent of $z_i$.
\end{enumerate}
\end{assumptionbox}

Independent latent variables are the standard assumption shared by ICA and disentangled representation learning \citep{hyvarinen2024identifiability, bengio2013representation}.
Stationarity means the generative process does not change between views.
Additive noise is the simplest perturbation model: random fluctuations are added on top of a deterministic signal, as in sensor noise, Brownian motion, or data augmentation by jitter. It is also what continuous-time dynamics produce over short intervals: an It\^o diffusion $dz = b(z)\,dt + \sigma\,dW$ with constant volatility has, to first order in $\Delta t$, the transition $z' = z + b(z)\Delta t + \sigma\sqrt{\Delta t}\,\eta$, which is (iii) with $m(z) = z + b(z)\Delta t$, whatever the drift.
Together, these assumptions define a general class of World Models.
Our \textit{forward} result (Sec.~\ref{sec:thm}) will specialize this class by choosing a specific latent variable distribution (Gaussian).
Our \textit{converse} result (Sec.~\ref{sec:converse}) shows that this is the \textit{unique} choice yielding linear identifiability in this class of worlds.

\subsubsection{The Gaussian World}\label{sec:gaussian-wm}

We now make a specific distributional choice within the framework above.
We assume \textit{Gaussian latents}, i.e., $z \sim \mathcal{N}(0, I_N)$.
This is the maximum-entropy distribution for a given mean and covariance \citep{jaynes1957information}; it assumes as little structure as possible.
Moreover, task-relevant latents are often aggregates of many micro-variables, which, by the central limit theorem, tend toward Gaussianity.
Within this class we take the transition to be linear in the latent, $m_i(z_i) = \rho\, z_i$.
Stationarity then requires $z' \sim \mathcal{N}(0, I_N)$, and by Cram\'er's decomposition theorem the independent noise must itself be Gaussian, which yields the Ornstein--Uhlenbeck (OU) transition \citep{uhlenbeck1930, doob1942}:
\begin{equation}\label{eq:dgp}
    z' = \rho\, z + \sqrt{1-\rho^2}\;\eta, \qquad \eta \sim \mathcal{N}(0, I_N),\quad \eta \perp z,
\end{equation}
where $\rho \in (0,1)$ controls the correlation between views.
One can verify: $\mathbb{E}[z'] = 0$, $\mathrm{Var}(z') = \rho^2 I_N + (1-\rho^2)I_N = I_N$, and $\mathrm{Cov}(z,z') = \rho I_N$.
The Gaussian is the \textit{unique} distribution for which a linear channel with independent Gaussian noise preserves the marginal, a consequence of the Gaussian being a fixed point of convolution up to rescaling.
The independence assumption is satisfied because the components of $z$ are independent and the noise $\eta$ has diagonal covariance, so each $(z_i, z'_i)$ evolves independently.

\subsection{The Learner: LeJEPA}\label{sec:objective}

The representation is characterized by the composed map $h = f \circ g : \mathbb{R}^N \to \mathbb{R}^N$ with $g$ the unknown generative function of the world and $f$ our learned representation.
The two components of LeJEPA \citep{balestriero2025lejepa} training are an invariance loss pulling positive pairs together and a regularizer that shapes the embedding distribution to prevent collapse, i.e., $h$ (specifically, $f$ in $h=f \circ g$) is trained to
\begin{equation}
    \min_h \mathcal{L}(h) = \underbrace{\mathbb{E}\!\left[\|h(z') - h(z)\|^2\right]}_{\textbf{Alignment}} \qquad \text{s.t.} \qquad \underbrace{h(z) \sim \mathcal{N}(0, I_N)}_{\textbf{Gaussianity}}
\end{equation}
We model the situation where SIGReg has succeeded, i.e., $h(z)$ matches the target Gaussian; in practice this holds approximately (Sec.~\ref{sec:approx}).
LeJEPA has no predictor, since SIGReg alone prevents collapse, so this is the objective as used in practice \citep{balestriero2025lejepa}; adding a linear or MLP predictor \citep{maes2026leworldmodel} leaves identifiability unchanged empirically (Sec.~\ref{sec:beyond}).
We require nothing else about $h$ except measurability so expectations are defined.
The result thus applies to any neural network, with output dimension taken to equal the latent dimension $N$ throughout (a mismatched output dimension is studied in Sec.~\ref{sec:beyond}).
In addition, \textit{whitening} (i.e., $\mathrm{Cov}(h(z)) = I_N$) fixes $\mathbb{E}[\|h(z)\|^2] = \mathbb{E}[\|h(z')\|^2] = N$, so
\begin{equation}\label{eq:loss-expand}
    \mathcal{L}(h) = 2N - 2\sum_{i=1}^N \mathbb{E}\!\left[h_i(z')\, h_i(z)\right].
\end{equation}
Thus, \emph{minimizing distance is equivalent to maximizing correlation} between the two views. The question becomes: among all measure-preserving maps $h$ with $h(z) \sim \mathcal{N}(0, I_N)$, which achieves the highest correlation between positive pairs $h(z), h(z')$?

\section{Spectral Analysis of the World}\label{sec:spectral}

Before proving anything, let us build intuition for the mathematical tools behind our main results. 

\paragraph{Transition Operator.}
The world's transition from $z$ to $z'$ (\ref{assumptions}) acts on any scalar function $h_i$ through the \textit{transition operator} $(Th_i)(z) = \mathbb{E}[h_i(z') \mid z]$, the expected value of $h_i$ at the next view.
This is the (stochastic) \textit{Koopman operator} of the dynamics \citep{koopman1931hamiltonian, mezic2005spectral}: linear even when the dynamics are not, with eigenfunctions $T\varphi_k = \lambda_k \varphi_k$ and, for the diffusion-type worlds we consider, eigenvalues $1 = \lambda_0 > \lambda_1 \geq \lambda_2 \geq \cdots \geq 0$.
Eigenfunctions with larger eigenvalues are more correlated across positive pairs, i.e., the most predictable features of the latent variables; this spectral view underlies both our forward and converse results.

\paragraph{Gaussian World: Hermite Polynomials.}
For Gaussian worlds (Sec.~\ref{sec:gaussian-wm}), the eigenfunctions are known in closed form: they are the Hermite polynomials $\{\mathrm{He}_k\}_{k \geq 0}$, the natural orthogonal basis for functions of Gaussian variables, analogous to Fourier modes for periodic functions.
The eigenvalue of a degree-$d$ Hermite polynomial is exactly $\rho^d$, a consequence of Mehler's formula \citep{mehler1866}.
This means any function $h_i(z)$ with zero mean and unit variance can be decomposed into a linear part (degree~1), a quadratic part (degree~2), a cubic part (degree~3), and so on, with variance fractions $w_1, w_2, w_3, \ldots$ summing to~1.
The correlation across positive pairs decomposes:
\begin{equation}\label{eq:corr-decomp}
    \mathbb{E}\!\left[h_i(z')\, h_i(z)\right] = w_1 \cdot \rho + w_2 \cdot \rho^2 + w_3 \cdot \rho^3 + \cdots \;\leq\; \rho,
\end{equation}
with equality if and only if $w_1 = 1$, i.e., $h_i$ is linear.
In words: \textit{any nonlinear distortion of the representation strictly reduces the correlation between positive pairs}.
This is the key intuition.

\paragraph{General Case: Sturm-Liouville Theory.}
For a general latent variable distribution (not necessarily Gaussian) evolving under constant diffusion, the eigenfunctions of the transition operator are characterized by a classical Sturm--Liouville (SL) equation \citep{sprekeler2014}.
The first non-constant eigenfunction $\varphi_1$ is always monotonic, providing identifiability up to a monotonic transformation.
But \textit{linear} identifiability requires $\varphi_1$ to be affine, which places a strong constraint on the latent variable distribution.
This is the engine behind our converse result (Sec.~\ref{sec:converse}): we show that only the Gaussian satisfies this constraint.
Full details in App.~\ref{app:hermite-details} (Gau{\ss}/Hermite) and App.~\ref{app:prior-work} (SL connection).


\section{Theory}\label{sec:main}
We state four results that together characterize when LeJEPA learns the World Model.
Thm.~\ref{thm:main} shows that a Gaussian world is \textit{linearly identifiable} with LeJEPA.
Thm.~\ref{thm:converse} establishes that, within the class of worlds defined in Sec.~\ref{sec:world-model}, the Gaussian is the \textit{unique} distribution with this property.
Thm.~\ref{thm:approx} bounds the recovery error when objectives are not fully satisfied.
Thm.~\ref{thm:planning} shows that linear identifiability enables \textit{optimal planning} in latent space.
All proofs have been verified in Lean~4 theorem prover modulo standard background lemmas axiomatized from the literature (App.~\ref{app:lean}).

\subsection{Forward Direction: LeJEPA Learns the World Model}\label{sec:thm}

\begin{theorembox}[LeJEPA Linear Identifiability]
\thmlabel{thm:main}
Consider the Gaussian world (Sec.~\ref{sec:gaussian-wm}). Let $h: \mathbb{R}^N \to \mathbb{R}^N$ be any measurable map with $h(z) \sim \mathcal{N}(0, I_N)$. Then $\mathcal{L}(h) \geq 2(1-\rho)\,N$, with equality if and only if $h(z) = Qz$ for some orthogonal $Q \in O(N)$. At any such optimum, $h(z') \mid h(z) \sim \mathcal{N}(\rho\, h(z), (1-\rho^2)\,I_N)$.
\end{theorembox}

\paragraph{Proof Sketch.}
Minimizing $\mathcal{L}$ is equivalent to maximizing $\sum_i \mathbb{E}[h_i(z') h_i(z)]$~\eqref{eq:loss-expand}.
The spectral bound~\eqref{eq:corr-decomp} gives $\mathbb{E}[h_i(z')h_i(z)] \leq \rho$ for each component, with equality if and only if $h_i$ is linear.
At equality, $h(z) = Qz$ for a matrix $Q$ with unit-norm rows; Gaussianity forces $QQ^\top = I_N$, so $Q$ is orthogonal.
The transition then follows by direct substitution: $h(z') = \rho\,h(z) + \sqrt{1-\rho^2}\,Q\eta$, and since $Q$ is orthogonal, $Q\eta \sim \mathcal{N}(0, I_N)$ independently of $h(z)$.
(Proof in App.~\ref{app:proof-forward}.)

\paragraph{Interpretation.}
The representation has no choice but to learn the full World Model.
Any representation satisfying the two LeJEPA objectives must recover a rotation/reflection of the true latent variables \textit{and} the true transition dynamics.
The only remaining ambiguity is a global rotation, which is inherent to the isotropic Gaussian.
We provide an alternative proof via Dirichlet energy and the Mazur--Ulam theorem in App.~\ref{app:dirichlet}, operating in a more theoretical ($\rho \to 1$) regime.

\subsection{Converse Direction: The Gaussian is Unique}\label{sec:converse}
Thm.~\ref{thm:main} uses only that $h(z)$ has whitened covariance ($\mathrm{Cov}(h(z)) = I_N$, entailed by SIGReg, Fig.~\ref{fig:scatter}). Does the specific choice of Gaussian matter, or would any white distribution work?
The optimal representation extracts the \textit{slowest} features of the latent process \citep{sobal2022slow}. Under additive noise, Sturm--Liouville theory orders these as increasingly oscillatory eigenfunctions \citep{sprekeler2014}; the first is always monotonic, giving identifiability up to a monotonic transformation for \emph{any} latent distribution. For \emph{linear} identifiability this eigenfunction must be affine, which only the Gaussian satisfies.

\begin{theorembox}[Gaussian Uniqueness]
\thmlabel{thm:converse}
Consider any world satisfying Assumptions~\ref{assumptions}. Suppose every minimizer of \eqref{eq:loss-expand} with $\mathrm{Cov}(h(z)) = I_N$ is linear, $h(z) = Qz$. Then $z$ is Gaussian.
\end{theorembox}

\paragraph{Proof Sketch.} 
Demanding an affine eigenfunction forces the \emph{score function} $(\log p)'$ of the latent distribution to be linear.
If $\varphi(z_i) = a z_i + b$ is an eigenfunction, then $\varphi'$ is constant, and the eigenvalue equation collapses to a linear ODE for $(\log p)'$ in $z_i$.
Solving it yields $\log p(z_i) \propto -(z_i - \mu)^2$ (the sign fixed by normalizability), which is Gaussian.
The argument is per-component; independence of the latent variables lifts the conclusion to the joint distribution.
(Proof in App.~\ref{app:converse-proof}.)

\subsection{Approximate Identifiability}\label{sec:approx}

The preceding results assume exact optimality.
In practice, both objectives are only approximately satisfied: alignment reaches a value near but not equal to its minimum, and the regularizer enforces approximate whitening.
Identifiability degrades gracefully in both quantities:

\begin{theorembox}[Approximate Identifiability]
\thmlabel{thm:approx}
Consider the Gaussian world (\ref{sec:gaussian-wm}). Let $h: \mathbb{R}^N \to \mathbb{R}^N$ be measurable with $\mathbb{E}[h(z)] = 0$, satisfying:
\textit{Approximate alignment:} $\mathcal{L}(h) \leq 2(1-\rho)\,\mathrm{tr}(\mathrm{Cov}(h(z))) + \delta$;
\textit{Approximate whitening:} $\|\mathrm{Cov}(h(z)) - I_N\|_F \leq \varepsilon$.
Define $D = \delta/(2\rho(1-\rho))$. Then there exists $Q \in O(N)$ such that
\begin{equation}\label{eq:stability}
    \mathbb{E}\!\left[\|h(z) - Qz\|^2\right] \;\leq\; D \;+\; (\varepsilon + D)^2
\end{equation}
\end{theorembox}

\paragraph{Interpretation.}
The bound has two pieces: a nonlinearity term $D$ measuring how far $h$ is from linear, and a distortion term $(\varepsilon + D)^2$ measuring how far the linear part is from orthogonal. Both vanish in the exact case ($\delta = \varepsilon = 0$), recovering Thm.~\ref{thm:main}. In practice the first dominates, so recovery error scales as $\delta / (2\rho(1-\rho))$: alignment is hard, whitening essentially free. (Proof in App.~\ref{app:stability}.)

\subsection{Optimal Latent Planning}\label{sec:planning-equivalence}

Theorems~\ref{thm:main}--\ref{thm:approx} characterize what the encoder recovers. We now ask what orthogonal identifiability buys for planning, a key motivation of World Models.
In control, an action $a_t$ moves the latent from $z_{t-1}$ to $z_t \sim p(\cdot \mid z_{t-1}, a_t)$. A plan $a_{1:T}$ from $z_0$ costs $\mathbb{E}\big[\ell_T(z_T) + \sum_{t=1}^{T} \ell(z_{t-1}, a_t)\big]$, with stage cost $\ell$ and terminal cost $\ell_T$; let $V^*(z_0)$ denote the minimal cost and $a^*_{1:T}(z_0)$ the optimal plan.

\begin{theorembox}[Optimal Latent Planning]
\thmlabel{thm:planning}
Let $h(z) = Qz$ with $Q \in O(N)$, and let the costs be rotation-invariant:
\begin{equation}\label{eq:orth-inv-app}
    \ell(Rz, a) = \ell(z, a), \qquad \ell_T(Rz) = \ell_T(z), \qquad \forall\, R \in O(N).
\end{equation}
Then planning in the learned latent $\hat z = h(z)$, with transitions $p$ mapped through $h$ and the same costs, yields the same value and plan: $\hat V^*\big(h(z_0)\big) = V^*(z_0)$ and $\hat a^*_{1:T}\big(h(z_0)\big) = a^*_{1:T}(z_0)$.
\end{theorembox}



\paragraph{Interpretation.}
Linear identifiability turns the learned representation into a useful World Model for planning: trajectories planned in the learned latent are mathematically identical to trajectories planned in the true world, with the same actions and the same value.
(Proof in App.~\ref{app:planning}.)

\section{Experiments}\label{sec:experiments}

We validate each result: linear identifiability on Gaussian latent variables (Sec.~\ref{sec:exp-gaussian}, Thm.~\ref{thm:main}); 
converse on a latent-distribution sweep and RL-policy~\citep{maes2026leworldmodel} latents (Sec.~\ref{sec:exp-nongauss}, Thm.~\ref{thm:converse});
approximate bound across all runs (Sec.~\ref{sec:exp-bound}, Thm.~\ref{thm:approx}); and near-optimal latent planning~\citep{tassa2018dmcontrol} (Sec.~\ref{sec:planning-exp}, Thm.~\ref{thm:planning}). Sec.~\ref{sec:beyond} then tests what happens beyond the assumptions: a misspecified latent dimension, a learned predictor, and natural images from ImageNet-X \citep{idrissi2023imagenetx}.

\subsection{Forward: Linear Identifiability from Gaussian Latent Variables}
\label{sec:exp-gaussian}

\begin{figure}[t]
    \centering
    \includegraphics[width=\textwidth]{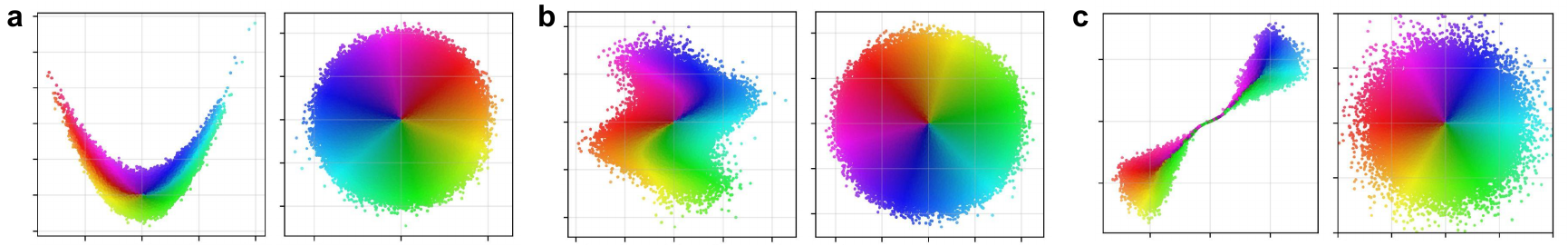}
    \caption{
        \textbf{2D Simulations.}
        Points colored by the polar angle and radius of the ground-truth latent variables $z \sim \mathcal{N}(0, I_2)$ (like Fig.~\ref{fig:teaser}).
        Observations (\textbf{a-c} left) $x = g(z)$ after nonlinear mixing: parabolic shear, sinusoidal shear, RealNVP coupling layer \citep{dinh2016density}.
        Learned embeddings (\textbf{a-c} right): LeJEPA recovers the isotropic Gaussian structure up to rotation, consistent with Thm.~\ref{thm:main}.}
    \label{fig:demo}
    \vspace{-5pt}
\end{figure}

We first verify Thm.~\ref{thm:main} in a controlled 2D setting.
We sample latents $z \sim \mathcal{N}(0, I_2)$ and apply four nonlinear mixing functions (Figures~\ref{fig:teaser}, \ref{fig:demo}):
a norm-dependent rotation $g(z) = R(\pi\|z\|_2)\,z$ producing a spiral \citep[c.f.][]{buchholz2025robustness},
a sinusoidal shear $g(z_1, z_2) = (z_1 + \sin(1.5\,z_2),\; z_2)$,
a parabolic shear $g(z_1, z_2) = (z_1,\; z_2 + z_1^2)$,
and a RealNVP-style coupling layer \citep{dinh2016density}.
All four are diffeomorphisms.
In addition, the spiral is measure-preserving, demonstrating that Gaussianity of the observations alone does not suffice for identifiability; alignment is essential.
We train a 4-layer MLP with the LeJEPA loss $\mathcal{L} = \lambda\,\mathcal{L}_{\mathrm{SIG}} + (1-\lambda)\,\mathcal{L}_{\mathrm{inv}}$, where positive pairs are generated via the OU transition in~\eqref{eq:dgp}.
Fig.~\ref{fig:demo} shows that the learned representation inverts each nonlinear mixing up to rotation, consistent with Thm.~\ref{thm:main}.
A grid search over $\lambda$ and $\rho$ (App.~\ref{app:grid}) shows: too much Gaussianity ($\lambda = 0.5$) collapses the representation, while the best recovery occurs at low $\lambda$ and high $\rho$.

\begin{table}[b]
\centering
\caption{\textbf{Scaling Comparison Across Regularizers} (mean $\pm$ std, 5 seeds). Three Gaussianity-enforcing objectives on a shared RealNVP mixing and matched encoder. SIGReg and VICReg maintain $R^2 > 0.999$ up to $N{=}2048$; InfoNCE matches at low $N$ but degrades at scale under fixed kernel width $\sigma{=}1$. At $N{=}4096$ all three fail at batch size 256; $^\dagger$raising the batch size to 1024 restores SIGReg (3 seeds). Per-method details in Tabs.~\ref{tab:scaling-sigreg}--\ref{tab:scaling-infonce}.}
\label{tab:scaling-comparison}
\begin{tabular}{r c ccc}
  & \textbf{Mixing} & \multicolumn{3}{c}{\textbf{Linear identifiability} $R^2(h \to z)$} \\
\cmidrule(lr){3-5}
$N$ & $R^2(x \to z)$ & SIGReg & VICReg & InfoNCE \\
 & {\scriptsize $\pm$std\,$\times 10^{-3}$} & {\scriptsize $\pm$std\,$\times 10^{-7}$} & {\scriptsize $\pm$std\,$\times 10^{-7}$} & {\scriptsize $\pm$std\,$\times 10^{-3}$} \\
\midrule
  2 & 0.781\tiny{$\pm$2.1} & 0.999998\tiny{$\pm$3.4} & 0.999996\tiny{$\pm$8.4} & 0.950961\tiny{$\pm$1.6} \\
  4 & 0.727\tiny{$\pm$24} & 0.999996\tiny{$\pm$12} & 0.999987\tiny{$\pm$54} & 0.910871\tiny{$\pm$8.2} \\
  8 & 0.728\tiny{$\pm$10} & 0.999993\tiny{$\pm$9.0} & 0.999988\tiny{$\pm$4.8} & 0.886818\tiny{$\pm$42} \\
  16 & 0.734\tiny{$\pm$6.3} & 0.999988\tiny{$\pm$4.9} & 0.999987\tiny{$\pm$4.6} & 0.999880\tiny{$\pm$0.01} \\
  32 & 0.737\tiny{$\pm$2.3} & 0.999981\tiny{$\pm$7.2} & 0.999981\tiny{$\pm$9.4} & 0.907809\tiny{$\pm$26} \\
  64 & 0.737\tiny{$\pm$1.5} & 0.999966\tiny{$\pm$7.4} & 0.999968\tiny{$\pm$8.1} & 0.648496\tiny{$\pm$3.1} \\
  128 & 0.739\tiny{$\pm$0.61} & 0.999938\tiny{$\pm$3.2} & 0.999942\tiny{$\pm$7.2} & 0.566955\tiny{$\pm$6.6} \\
  256 & 0.742\tiny{$\pm$0.49} & 0.999884\tiny{$\pm$7.9} & 0.999889\tiny{$\pm$7.2} & 0.696587\tiny{$\pm$0.49} \\
  512 & 0.749\tiny{$\pm$0.30} & 0.999775\tiny{$\pm$6.7} & 0.999785\tiny{$\pm$6.9} & 0.704393\tiny{$\pm$0.26} \\
  1024 & 0.763\tiny{$\pm$0.17} & 0.999561\tiny{$\pm$12} & 0.999582\tiny{$\pm$11} & 0.720241\tiny{$\pm$0.20} \\
  2048 & 0.768\tiny{$\pm$0.05} & 0.999043\tiny{$\pm$7.9} & 0.999094\tiny{$\pm$7.5} & 0.727140\tiny{$\pm$0.10} \\
  4096 & 0.768\tiny{$\pm$0.03} & 0.663\,$\to$\,0.999449$^\dagger$ & 0.652 & 0.727111\tiny{$\pm$0.02} \\
\bottomrule
\end{tabular}
\end{table}

\paragraph{Scaling to High Dimensions.}
We next sweep the latent dimension $N \in \{2^1,\ldots,2^{12}\}$ (beyond, e.g., DINOv3's $768$ embedding dimensions), using a RealNVP mixing and matched encoder~\citep{dinh2016density} so that any failure is attributable to optimization rather than encoder expressivity.
We test all three Gaussianity-enforcing classes on the same setup: SIGReg \citep{balestriero2025lejepa}, VICReg \citep{bardes2022}, and InfoNCE \citep{oord2018representation}. Table~\ref{tab:scaling-comparison}: batch-statistic estimators (SIGReg, VICReg) maintain $R^2 > 0.999$ up to $N = 2048$; InfoNCE degrades at scale under fixed kernel width. At $N = 4096$ the default batch of 256 becomes the bottleneck for both batch-statistic estimators, and a batch of 1024 restores SIGReg to $R^2 = 0.9994$ (App.~\ref{app:scaling}).
Thm.~\ref{thm:main} is thus supported across mixing functions, dimensions, and hyperparameters; where methods differ, the cause is estimation (kernel width, batch size), not identifiability.

\subsection{Converse: Non-Gaussian Latent Variables Break Linear Identifiability}
\label{sec:exp-nongauss}

\begin{table}[b]
\centering
\caption{\textbf{Linear Identifiability on RL Trajectories.} (mean $\pm$ std, 3 seeds).
\textbf{(left)} Gaussian OU pairs recover true latents with rising $R^2$.
\textbf{(right)} RL-policy trajectories~\citep{maes2026leworldmodel} ($\delta$: frame separation) show reduced identifiability because of anisotropy ($\rho_0 \neq \rho_1$) and non-Gaussian transitions (Figs.~\ref{fig:traj_distributions},\ref{fig:rho_vs_sigreg}).
}
\label{tab:reacher}
\resizebox{\textwidth}{!}{%
\begin{tabular}{r cc | r cc ccc}
 \multicolumn{3}{c}{\textbf{OU (Gaussian)}} & \multicolumn{6}{c}{\textbf{Trajectory (non-Gaussian)}} \\
\cmidrule(lr){1-3} \cmidrule(lr){4-9}
$\rho$ & $R^2(z \to h)$ & $R^2(h \to z)$ & $\delta$ & $\rho_0$ & $\rho_1$ & $R^2(z \to h)$ & $R^2(h \to z_0)$ & $R^2(h \to z_1)$ \\
\midrule
  0.30 & 0.67\tiny{$\pm$2e-02} & 0.67\tiny{$\pm$2e-02} & 1 & 1.000 & 0.999 & -0.39\tiny{$\pm$1e-01} & 0.71\tiny{$\pm$3e-02} & -0.03\tiny{$\pm$4e-02} \\
  0.50 & 0.86\tiny{$\pm$1e-02} & 0.86\tiny{$\pm$1e-02} & 2 & 0.999 & 0.996 & -0.47\tiny{$\pm$5e-02} & 0.73\tiny{$\pm$4e-03} & 0.01\tiny{$\pm$2e-03} \\
  0.70 & 0.93\tiny{$\pm$7e-03} & 0.93\tiny{$\pm$7e-03} & 4 & 0.997 & 0.991 & -0.05\tiny{$\pm$3e-01} & 0.51\tiny{$\pm$1e-01} & 0.43\tiny{$\pm$9e-02} \\
  0.80 & 0.94\tiny{$\pm$3e-04} & 0.94\tiny{$\pm$3e-04} & 8 & 0.992 & 0.982 & 0.50\tiny{$\pm$2e-02} & 0.80\tiny{$\pm$2e-03} & 0.78\tiny{$\pm$6e-03} \\
  0.90 & 0.95\tiny{$\pm$7e-04} & 0.95\tiny{$\pm$7e-04} & 16 & 0.981 & 0.963 & 0.44\tiny{$\pm$4e-02} & 0.63\tiny{$\pm$2e-02} & 0.87\tiny{$\pm$2e-02} \\
  0.95 & 0.95\tiny{$\pm$2e-04} & 0.95\tiny{$\pm$2e-04} & 32 & 0.959 & 0.928 & 0.45\tiny{$\pm$6e-02} & 0.62\tiny{$\pm$2e-02} & 0.81\tiny{$\pm$1e-02} \\
  0.99 & 0.95\tiny{$\pm$4e-04} & 0.95\tiny{$\pm$4e-04} & 64 & 0.915 & 0.863 & 0.44\tiny{$\pm$4e-02} & 0.55\tiny{$\pm$3e-02} & 0.77\tiny{$\pm$2e-02} \\
\bottomrule
\end{tabular}}
\end{table}

Next, we validate Thm.~\ref{thm:converse}'s predictions that non-Gaussian latents break linear identifiability. 

\begin{figure}[t]
    \centering
    \includegraphics[width=0.99\linewidth]{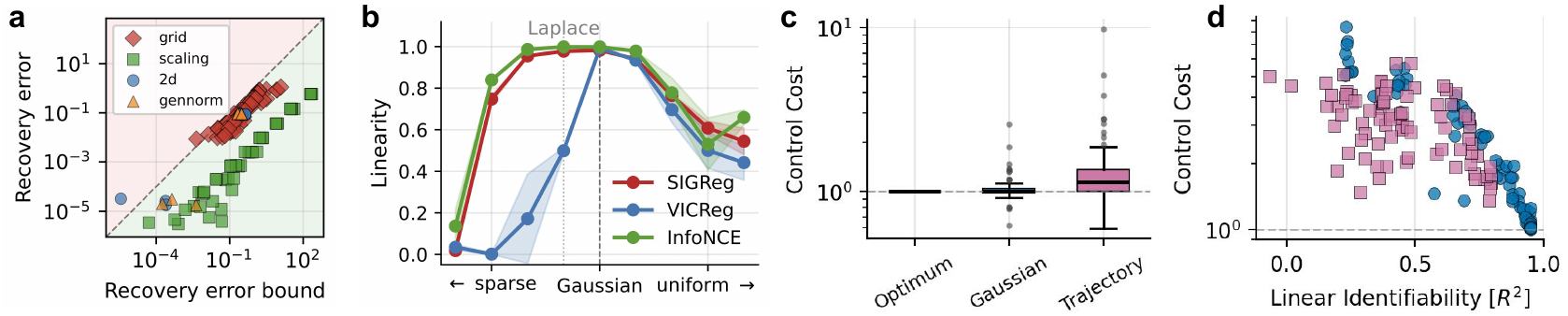}
    \caption{
    \textbf{Experimental Results.}
    \textbf{(a)} \textit{Bound Verification.} SIGReg runs across grid, 2D, scaling, and gennorm $\alpha{=}2$ lie below the diagonal, confirming Thm.~\ref{thm:approx}. Two near-zero outliers reflect finite-sample noise.
    \textbf{(b)} \textit{Gaussian Optimality.} Linear recovery, \(R^2(h \to z\)), peaks at Gaussian, illustrating Thm.~\ref{thm:converse}. SIGReg's Gaussianization of $h$ is more robust to non-Gaussian latent variable distributions than whitening.
    \textbf{(c)} Control cost over $K = 30$ random start-goal pairs (path length $\geq 1$, ideal $1$). The Gaussian encoder is statistically indistinguishable from the oracle; the Trajectory encoder is biased upward.
    \textbf{(d)} Control cost decreases with linear identifiability $R^2$, supporting Thm.~\ref{thm:planning}.
    }
    \label{fig:bound_unique_plan}
    \vspace{-5pt}
\end{figure}

\paragraph{Latent-Distribution Sweep.}
We sweep the latent variable through the generalized normal family with shape parameter $\alpha$ ($\alpha \to 0$ heavy-tailed, $\alpha=1$ Laplace, $\alpha=2$ Gaussian, $\alpha \to \infty$ uniform).
Linear recovery peaks at $\alpha=2$ across all three objectives (Fig.~\ref{fig:bound_unique_plan}b, App.~\ref{app:gennorm}), illustrating Thm.~\ref{thm:converse}; SIGReg and InfoNCE retain a wider plateau than VICReg for heavy-tailed latents.

\paragraph{Pixel-Based RL Trajectories.}
The DMC Reacher \citep{tassa2018dmcontrol} has two joints, giving a 2D latent state $z = (\theta_0, \theta_1)$ (Fig.~\ref{fig:reacher_annotated}).
We train a CNN encoder with LeJEPA (App.~\ref{app:reacher}) under two conditions sharing the same rendering pipeline but different distributions:
\textit{(i) OU:} Gaussian samples $z \sim \mathcal{N}(0, I_2)$ as before~\eqref{eq:dgp};
\textit{(ii) Trajectory:} joint-angle pairs $(z_t, z_{t+\delta})$ with $\delta$ frames separation from $10$k RL episodes~\citep{maes2026leworldmodel}.
Table~\ref{tab:reacher} (left) shows that OU pairs attain $R^2 = 0.95$ at $\rho = 0.99$, with the two joint dimensions linearly recovered.
In contrast, real trajectories break the Gaussian assumption (App.~\ref{app:reacher}, Figs.~\ref{fig:traj_distributions},\ref{fig:rho_vs_sigreg}).
Table~\ref{tab:reacher} (right) shows per-dimension $R^2$ is anisotropic, and total $R^2$ never exceeds $0.5$, consistent with Thm.~\ref{thm:converse}.
The trajectory condition violates several theory assumptions at once (non-Gaussian marginals, anisotropic $\rho_0 \neq \rho_1$, joint-limit wrapping).

\subsection{Approximation Bound and Loss Predict Identifiability}
\label{sec:exp-bound}

Thm.~\ref{thm:approx} bounds the recovery error by the whitening error $\varepsilon = \|\mathrm{Cov}(h(z)) - I\|_F$ and the alignment gap $\delta = \mathcal{L}(h) - 2(1-\rho)\,\mathrm{tr}(\mathrm{Cov}(h(z)))$.
For each run we compute $\varepsilon$, $\delta$, the bound $D + (\varepsilon + D)^2$, and the actual recovery error $\min_{Q \in O(N)} \mathbb{E}[\|h(z) - Qz\|^2]$ (Fig.~\ref{fig:bound_unique_plan}a).
The bound holds across grid search, 2D mixings, scaling, and the latent-distribution sweep, supporting Thm.~\ref{thm:approx} empirically.
As a practical corollary, training loss is a reliable proxy for identifiability (App.~\ref{app:loss}).

\subsection{Linear Identifiability Enables Latent-Space Planning}
\label{sec:planning-exp}

\begin{figure}[t]
    \centering
    \includegraphics[width=0.9\linewidth]{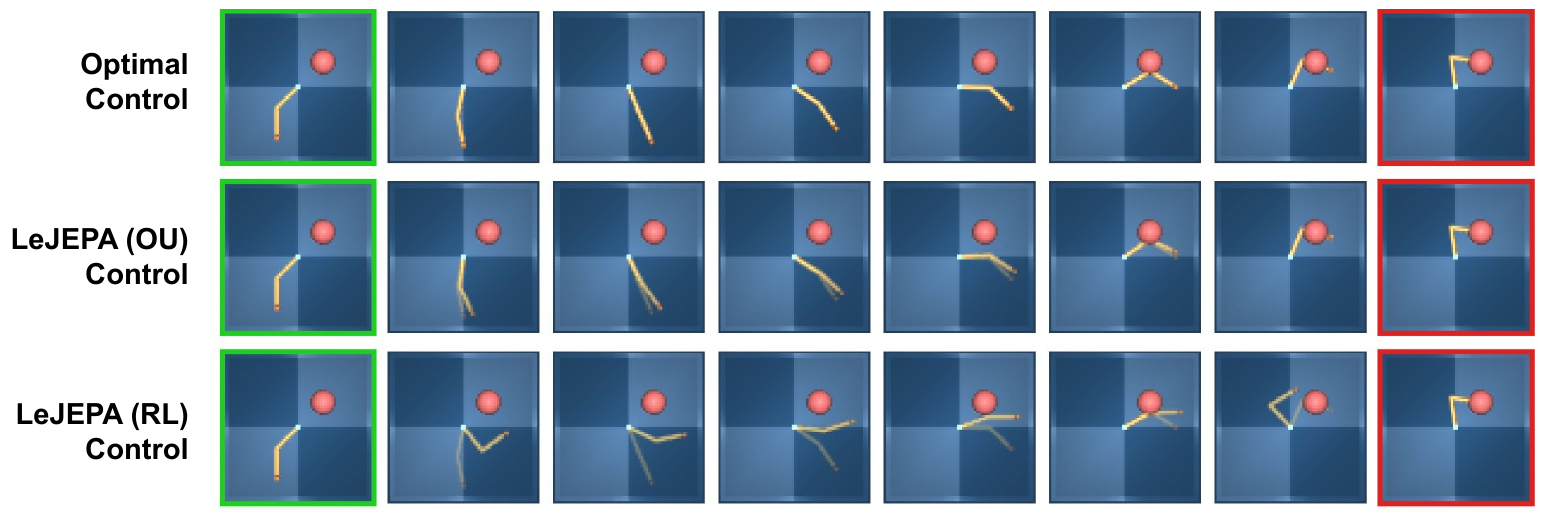}
    \vspace{-5pt}
    \caption{
        \textbf{Linear Identifiability Enables Latent-Space Planning.}
        Interpolation in each encoder's latent space between fixed \green{start} and \red{goal} frames, decoded by nearest-neighbor retrieval (Fig.~\ref{fig:planning_scatter}).
        \textbf{top)} Oracle (joint-space straight line) \citep{tassa2018dmcontrol}.
        \textbf{middle)} Gaussian encoder (Sec.~\ref{sec:gaussian-wm}: OU, $\rho = 0.99$) tracks the oracle (overlaid) closely. 
        \textbf{bottom)} RL trajectory \citep{maes2026leworldmodel} encoder (stride $\delta = 8$) deviates.
    }
    \label{fig:planning}
    \vspace{-5pt}
\end{figure}

Thm.~\ref{thm:planning} predicts that any planner using a rotation-invariant cost attains identical performance in the learned and true latent. We test a natural instance: goal-reaching, where the straight line $\hat z_0 \to \hat z^*$ is the cost-minimizing trajectory. For an encoder satisfying $h(z) \approx Qz$, this latent straight line decodes to a near-straight path in the true latent; without, the same plan induces a curved path (Fig.~\ref{fig:planning_scatter}).
Figs.~\ref{fig:bound_unique_plan}c,d;~\ref{fig:planning} confirm both. The Gaussian encoder's straight-latent plans decode to oracle-quality joint-space trajectories (Figs.~\ref{fig:planning},~\ref{fig:bound_unique_plan}c), while the Trajectory encoder inflates control cost. Across all models, control cost tracks linear identifiability $R^2$ monotonically (Fig.~\ref{fig:bound_unique_plan}d).
\textit{Thus, linear identifiability is the structural property that turns a faithful World Model into a useful planner.}

\subsection{Beyond the Assumptions: Latent Dimension, Natural Images, Predictors}
\label{sec:beyond}

\begin{wrapfigure}{r}{0.6\textwidth}
    \vspace{-12pt}
    \centering
    \includegraphics[width=\linewidth]{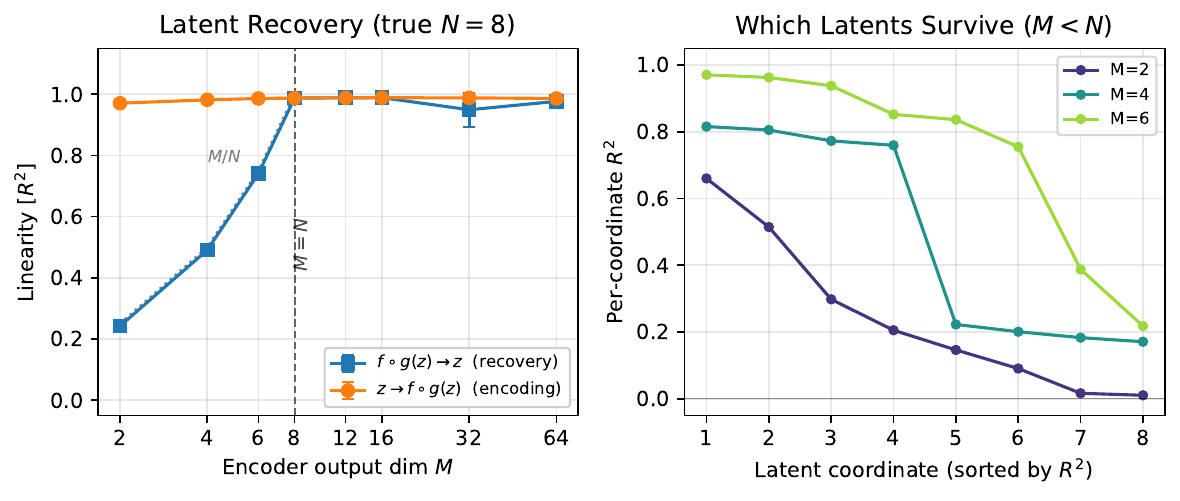}
    \vspace{-16pt}
    \caption{\textbf{Misspecified latent dimension} (true $N{=}8$, 5 seeds). \textbf{Left:} recovery tracks $M/N$ below $N$ and saturates above. \textbf{Right:} for $M < N$, $M$ latents are recovered well and the rest poorly.}
    \label{fig:dims}
    \vspace{-10pt}
\end{wrapfigure}
\paragraph{Unknown latent dimension.}
Our theorems assume the encoder output dimension $M$ equals the latent dimension $N$. What if it does not (Fig.~\ref{fig:dims})?
Below $N$, the encoder keeps an $M$-dimensional subspace of the latents intact rather than compressing all of them, so recovery tracks $M/N$. Above $N$, recovery saturates.
The true $N$ can be read off without ground truth from the participation ratio of the embedding covariance, $\mathrm{PR} = (\sum_i \lambda_i)^2 / \sum_i \lambda_i^2$: in both our settings, $\mathrm{PR} = M$ exactly up to $M{=}N$ and falls below $M$ beyond (App.~\ref{app:dims}).
Which subspace is kept, and when the system resorts to \textit{superposition} \citep{smolensky_tensor_1990, elhage_toy_2022, klindt2025superposition}, remains open.

\begin{wraptable}{r}{0.47\textwidth}
\vspace{-15pt}
\centering
\caption{\textbf{ImageNet-X Linear Probes.} Mean balanced accuracy over the nine factors with $>1000$ balanced samples (5 repetitions; full table in App.~\ref{app:imagenetx}).}
\label{tab:imagenetx}
\scriptsize
\setlength{\tabcolsep}{2.5pt}
\begin{tabular}{l ccc cc cc}
\toprule
 & LeJEPA & DINO & iBOT & v2 & v3 & init & pixels \\
\midrule
data & \multicolumn{3}{c}{IN-1k} & 142M & 1.7B & -- & -- \\
acc. & \textbf{.677} & .589 & .628 & .656 & .645 & .545 & .533 \\
\bottomrule
\end{tabular}
\end{wraptable}

\paragraph{Natural images.}
Without ground-truth latents, we test the observable consequence: annotated factors should be linearly decodable. Following \citet[Fig.~3]{reizinger2024cross}, we probe the ImageNet-X factors \citep{idrissi2023imagenetx} in ViT-B/16 encoders trained on ImageNet-1k with LeJEPA, DINO \citep{caron2021emerging}, and iBOT \citep{zhou2022ibot} under a shared codebase and schedule. LeJEPA leads (Tab.~\ref{tab:imagenetx}), also against DINOv2 \citep{oquab2024dinov2} and DINOv3 \citep{simeoni2025dinov3} trained on far more data, while meeting SIGReg only approximately, the regime of Thm.~\ref{thm:approx} (App.~\ref{app:imagenetx}).

\paragraph{Predictors.}
LeJEPA has no predictor, but action-conditioned World Models do \citep{maes2026leworldmodel}. Aligning $h(z')$ with a learned prediction $P(h(z))$ leaves recovery unchanged ($0.989$ without, $0.987$ linear, $0.988$ MLP predictor; $N{=}8$, 5 seeds), and the linear $P$ converges to the true transition operator ($\|P - \rho I\|_F = 0.011$). For the MLP this is empirical only, as an expressive predictor could absorb a nonlinear distortion of $h$ (App.~\ref{app:pred}).


\section{Limitations}\label{sec:limitations}

\paragraph{Are the latents Gaussian?}
The Gaussian is the maximum-entropy distribution for a given mean and covariance \citep{jaynes1957information}, making it the least-assumption prior. Whether real-world latents are Gaussian is unknowable from observations alone, but the same applies to the non-Gaussianity assumptions of classical ICA \citep{comon1994} (maybe a structuralist perspective \citep{joshi2026causality} is more agnostic). There is a scale-of-description argument in favor: individual micro-variables may be non-Gaussian, but task-relevant latent variables are often aggregates that tend toward Gaussianity by the central limit theorem.
These are arguments rather than evidence, but Thm.~\ref{thm:approx} does not need them, and we test non-Gaussian latents directly (Fig.~\ref{fig:bound_unique_plan}b). Dependent, non-Euclidean and other latents remain open questions.

\paragraph{Finite samples and optimization.} 
Our result is a population-level statement about the global optimum. Thm.~\ref{thm:approx} shows the guarantee degrades continuously with respect to alignment gap and covariance deviation, but does not address how these scale with sample size \citep{lyu2022finite} or training dynamics \citep{simon2023stepwise}. Empirically, recovery emerges abruptly after an initial plateau of a few hundred to a few thousand steps (App.~\ref{app:dynamics}). The few bound violations we observe (Fig.~\ref{fig:bound_unique_plan}a) are consistent with finite-sample estimation noise in $\varepsilon$ and $\delta$.


\section{Discussion}\label{sec:discussion}

\paragraph{Outlook.}
Linear identifiability addresses the \emph{state} side of a World Model; the action-conditioned transition $\hat p(\hat z' \mid \hat z, a)$ must still be learned. Lifting identifiability to that setting connects naturally to interventional causal representation learning, where actions act as interventions on the latent variables \citep{ahuja2023, buchholz2024}, and onward to causal graphs \citep{nam2026causal}. Actions may also relax the Gaussian assumption: action-conditioned World Models can prevent collapse through inverse-dynamics regularization instead of an isotropic Gaussian target \citep{ivashkov2026sensorimotor}, a promising route to identifiability for non-Gaussian latents that are subject to control. Our assumptions may not hold exactly in practice, but Thm.~\ref{thm:approx} shows the guarantee degrades gracefully.

\paragraph{Practical Implications.}
Two implications, one for data and one for objective. \textit{Data:} the Reacher result shows that the same physical system supports identifiability when sampled isotropically (OU) but not under a goal-directed policy, whose marginals collapse onto a low-entropy region of latent space (Tab.~\ref{tab:reacher}); for self-supervised pretraining, exploration approximating an isotropic random walk keeps the data in the regime our theory covers. \textit{Objective:} SIGReg, VICReg, and InfoNCE all yield linear identifiability when their assumptions hold, and our experiments suggest they fail in different ways: pair-based estimators are sensitive to kernel choice at scale (Tab.~\ref{tab:scaling-comparison}), moment-based estimators to non-Gaussian latents (App.~\ref{app:gennorm}). Which estimator is preferable in practice, and how the choice interacts with batch size, architecture, and data regime, is left to future work.

\paragraph{A Theoretical Foundation for World Models.}
Our theory inverts the classical ICA narrative: in \emph{linear} ICA the Gaussian is the one distribution where source separation fails \citep{comon1994}; in our nonlinear setting it is exactly what enables it. The payoff is in control: linear identifiability turns a learned representation into a usable \emph{state} for a control system, so any orthogonally-invariant cost transfers from the true world to the learned latent without modification, and simple planning (Thm.~\ref{thm:planning}) and linear probing come for free. \textit{This is what it means to provably learn a World Model.}

\section*{Acknowledgments}
We thank Aapo Hyv\"arinen, Patrik Reizinger, Attila Juhos, Heejeong (Hazel) Nam, Christian Intern\`o and Habon Issa for their feedback and inputs.
DK acknowledges the CSHL GPU cluster with assistance from the US National Institutes of Health Grant S10OD028632-01.

\bibliographystyle{unsrtnat}
\bibliography{references}

\newpage
\appendix
\input{appendix}

\end{document}

%% file: figures/dgp.tex
\begin{tikzpicture}[
    >=Stealth,
    font=\small,
    arrow/.style={->, thick, >=Stealth},
    spring/.style={decorate, decoration={zigzag, segment length=3.5pt, amplitude=1.2pt}},
    scale=0.82, every node/.style={transform shape=false},
]

\definecolor{ptA}{RGB}{50,100,200}
\definecolor{ptB}{RGB}{200,80,80}
\definecolor{ptC}{RGB}{80,180,100}
\definecolor{ptD}{RGB}{180,130,50}
\definecolor{ptE}{RGB}{130,60,180}

\begin{scope}[shift={(0,0)}]
    \node[font=\small\sffamily, anchor=south] at (0,2.05) {World $z$};

    \fill[blue!6, opacity=0.8] (0,0) ellipse (1.9 and 1.9);
    \fill[blue!10, opacity=0.8] (0,0) ellipse (1.4 and 1.4);
    \fill[blue!16, opacity=0.8] (0,0) ellipse (0.85 and 0.85);
    \fill[blue!22, opacity=0.6] (0,0) ellipse (0.38 and 0.38);

    \draw[black!15, thin] (-1.9,0) -- (1.9,0);
    \draw[black!15, thin] (0,-1.9) -- (0,1.9);

    \fill[ptA, opacity=0.6] (0.3,0.8) circle (1.3pt);
    \fill[ptA!80!ptC, opacity=0.6] (-0.5,0.4) circle (1.3pt);
    \fill[ptC, opacity=0.6] (0.7,-0.3) circle (1.3pt);
    \fill[ptC!70!ptB, opacity=0.6] (-0.2,-0.7) circle (1.3pt);
    \fill[ptB, opacity=0.6] (0.1,0.2) circle (1.3pt);
    \fill[ptE, opacity=0.6] (-0.8,0.1) circle (1.3pt);
    \fill[ptA!50!ptE, opacity=0.6] (0.5,0.5) circle (1.3pt);
    \fill[ptD, opacity=0.6] (-0.3,-0.4) circle (1.3pt);
    \fill[ptB!60!ptD, opacity=0.6] (0.9,0.1) circle (1.3pt);
    \fill[ptC!50!ptE, opacity=0.6] (-0.6,-0.5) circle (1.3pt);
    \fill[ptB!80!ptC, opacity=0.6] (0.2,-0.9) circle (1.3pt);
    \fill[ptA!60!ptE, opacity=0.6] (-0.1,0.6) circle (1.3pt);
    \fill[ptD!70!ptC, opacity=0.6] (0.4,-0.5) circle (1.3pt);
    \fill[ptE!60!ptA, opacity=0.6] (-0.7,0.7) circle (1.3pt);
    \fill[ptA!40!ptB, opacity=0.6] (0.6,0.9) circle (1.3pt);
    \fill[ptD!50!ptE, opacity=0.6] (-0.4,-0.2) circle (1.3pt);
    \fill[ptB!40!ptE, opacity=0.6] (0.8,-0.7) circle (1.3pt);
    \fill[ptC!40!ptA, opacity=0.6] (-0.9,-0.3) circle (1.3pt);
    \fill[ptE!50!ptB, opacity=0.6] (0.0,1.1) circle (1.3pt);
    \fill[ptA!70!ptC, opacity=0.6] (-0.5,0.9) circle (1.3pt);
    \fill[ptD!30!ptA, opacity=0.6] (1.1,0.4) circle (1.3pt);
    \fill[ptC!60!ptD, opacity=0.6] (-1.0,0.5) circle (1.3pt);
    \fill[ptB!50!ptA, opacity=0.6] (0.3,-1.2) circle (1.3pt);
    \fill[ptE!40!ptC, opacity=0.6] (-0.8,-0.8) circle (1.3pt);

    \fill[red!85!black] (-0.6,-0.65) circle (3pt);
    \fill[orange!80!red] (-0.35,-0.45) circle (3pt);
    \draw[red!50, thick, shorten <=3.5pt, shorten >=3.5pt] (-0.6,-0.65) -- (-0.35,-0.45);

    \node[font=\footnotesize, red!85!black, anchor=north east] at (-0.65,-0.7) {$z$};
    \node[font=\footnotesize, orange!80!red, anchor=south west] at (-0.28,-0.4) {$z'$};
    \node[font=\tiny, red!60!black, anchor=north west] at (-0.15,-0.62) {\textit{positive pair}};
\end{scope}

\draw[arrow, very thick, black!60] (2.2,0.2) -- (3.2,0.2);
\node[font=\footnotesize\bfseries, black!70] at (2.7,0.6) {$g$};
\node[font=\tiny, text=black!45, align=center] at (2.7,-0.25) {nonlinear\\[-1pt]mixing};

\begin{scope}[shift={(5.4,0)}]
    \node[font=\small\sffamily, anchor=south] at (0,2.05) {Data $x = g(z)$};

    \fill[black!4, opacity=0.9]
        plot[smooth cycle, tension=0.6] coordinates {
            (-1.5, 0.4) (-0.9, 1.5) (0.2, 1.8) (1.1, 1.2)
            (1.8, 0.3) (1.6, -0.7) (0.7, -1.6) (-0.3, -1.4)
            (-1.1, -1.7) (-1.7, -0.8) (-1.9, -0.2)
        };
    \fill[black!8, opacity=0.7]
        plot[smooth cycle, tension=0.65] coordinates {
            (-0.7, 0.6) (-0.2, 1.2) (0.6, 0.9)
            (1.1, 0.1) (0.8, -0.8) (0.1, -0.9)
            (-0.6, -1.1) (-1.1, -0.4) (-1.0, 0.1)
        };
    \fill[black!13, opacity=0.6]
        plot[smooth cycle, tension=0.7] coordinates {
            (-0.1, 0.5) (0.4, 0.6) (0.7, 0.0)
            (0.3, -0.5) (-0.3, -0.6) (-0.6, -0.2) (-0.4, 0.3)
        };

    \fill[ptA, opacity=0.5] (-0.7,1.1) circle (1.3pt);
    \fill[ptA!80!ptC, opacity=0.5] (0.8,0.8) circle (1.3pt);
    \fill[ptC, opacity=0.5] (1.3,-0.4) circle (1.3pt);
    \fill[ptC!70!ptB, opacity=0.5] (-0.4,-1.2) circle (1.3pt);
    \fill[ptB, opacity=0.5] (0.3,0.2) circle (1.3pt);
    \fill[ptE, opacity=0.5] (-1.3,-0.4) circle (1.3pt);
    \fill[ptA!50!ptE, opacity=0.5] (0.2,1.4) circle (1.3pt);
    \fill[ptD, opacity=0.5] (0.5,-1.1) circle (1.3pt);
    \fill[ptB!60!ptD, opacity=0.5] (1.4,0.4) circle (1.3pt);
    \fill[ptC!50!ptE, opacity=0.5] (-0.9,-1.3) circle (1.3pt);
    \fill[ptB!80!ptC, opacity=0.5] (-0.1,-0.9) circle (1.3pt);
    \fill[ptA!60!ptE, opacity=0.5] (-0.5,0.7) circle (1.3pt);
    \fill[ptD!70!ptC, opacity=0.5] (1.0,-0.8) circle (1.3pt);
    \fill[ptE!60!ptA, opacity=0.5] (-1.4,0.3) circle (1.3pt);
    \fill[ptA!40!ptB, opacity=0.5] (0.4,1.1) circle (1.3pt);
    \fill[ptD!50!ptE, opacity=0.5] (-0.3,-0.4) circle (1.3pt);
    \fill[ptB!40!ptE, opacity=0.5] (0.7,-0.3) circle (1.3pt);
    \fill[ptC!40!ptA, opacity=0.5] (-1.2,0.7) circle (1.3pt);
    \fill[ptE!50!ptB, opacity=0.5] (-0.6,1.4) circle (1.3pt);
    \fill[ptA!70!ptC, opacity=0.5] (0.0,0.7) circle (1.3pt);
    \fill[ptD!30!ptA, opacity=0.5] (1.5,0.0) circle (1.3pt);
    \fill[ptC!60!ptD, opacity=0.5] (-1.5,-0.1) circle (1.3pt);
    \fill[ptB!50!ptA, opacity=0.5] (0.6,-1.4) circle (1.3pt);
    \fill[ptE!40!ptC, opacity=0.5] (-0.8,-0.8) circle (1.3pt);

    \fill[red!85!black] (-1.1,0.9) circle (3pt);
    \fill[orange!80!red] (1.2,-0.9) circle (3pt);

    \draw[red!30, dashed, thick] (-1.1,0.9) to[out=-20, in=160] (0.0,-0.1) to[out=-20, in=150] (1.2,-0.9);
\end{scope}

\draw[arrow, very thick, black!60] (7.6,0.2) -- (8.6,0.2);
\node[font=\footnotesize\bfseries, black!70] at (8.1,0.6) {$f$};
\node[font=\tiny, text=black!45, align=center] at (8.1,-0.25) {encoder\\[-1pt](learned)};

\begin{scope}[shift={(10.8,0)}]
    \node[font=\small\sffamily, anchor=south] at (0,2.05) {LeJEPA $f(x)$};

    \fill[violet!6, opacity=0.8] (0,0) ellipse (1.9 and 1.9);
    \fill[violet!10, opacity=0.8] (0,0) ellipse (1.4 and 1.4);
    \fill[violet!16, opacity=0.8] (0,0) ellipse (0.85 and 0.85);
    \fill[violet!22, opacity=0.6] (0,0) ellipse (0.38 and 0.38);

    \draw[black!15, thin] (-1.9,0) -- (1.9,0);
    \draw[black!15, thin] (0,-1.9) -- (0,1.9);

    \fill[ptA, opacity=0.6] (-0.28,0.81) circle (1.3pt);
    \fill[ptA!80!ptC, opacity=0.6] (-0.64,-0.01) circle (1.3pt);
    \fill[ptC, opacity=0.6] (0.73,0.22) circle (1.3pt);
    \fill[ptC!70!ptB, opacity=0.6] (0.30,-0.66) circle (1.3pt);
    \fill[ptB, opacity=0.6] (-0.05,0.22) circle (1.3pt);
    \fill[ptE, opacity=0.6] (-0.68,-0.44) circle (1.3pt);
    \fill[ptA!50!ptE, opacity=0.6] (0.06,0.70) circle (1.3pt);
    \fill[ptD, opacity=0.6] (0.03,-0.50) circle (1.3pt);
    \fill[ptB!60!ptD, opacity=0.6] (0.63,0.66) circle (1.3pt);
    \fill[ptC!50!ptE, opacity=0.6] (-0.14,-0.77) circle (1.3pt);
    \fill[ptB!80!ptC, opacity=0.6] (0.73,-0.56) circle (1.3pt);
    \fill[ptA!60!ptE, opacity=0.6] (-0.46,0.40) circle (1.3pt);
    \fill[ptD!70!ptC, opacity=0.6] (0.63,-0.13) circle (1.3pt);
    \fill[ptE!60!ptA, opacity=0.6] (-0.99,0.09) circle (1.3pt);
    \fill[ptA!40!ptB, opacity=0.6] (-0.12,1.08) circle (1.3pt);
    \fill[ptD!50!ptE, opacity=0.6] (-0.18,-0.41) circle (1.3pt);
    \fill[ptB!40!ptE, opacity=0.6] (1.06,-0.02) circle (1.3pt);
    \fill[ptC!40!ptA, opacity=0.6] (-0.50,-0.81) circle (1.3pt);
    \fill[ptE!50!ptB, opacity=0.6] (-0.71,0.84) circle (1.3pt);
    \fill[ptA!70!ptC, opacity=0.6] (-0.96,0.37) circle (1.3pt);
    \fill[ptD!30!ptA, opacity=0.6] (0.59,1.01) circle (1.3pt);
    \fill[ptC!60!ptD, opacity=0.6] (-1.09,-0.26) circle (1.3pt);
    \fill[ptB!50!ptA, opacity=0.6] (1.00,-0.73) circle (1.3pt);
    \fill[ptE!40!ptC, opacity=0.6] (-0.10,-1.13) circle (1.3pt);

    \fill[red!85!black] (-0.04,-0.88) circle (3pt);
    \fill[orange!80!red] (0.02,-0.57) circle (3pt);
    \draw[red!60, thick, spring, shorten <=3.5pt, shorten >=3.5pt] (-0.04,-0.88) -- (0.02,-0.57);
    \node[font=\tiny, red!60!black, anchor=west] at (0.1,-0.78) {\textit{attract}};

    \node[font=\tiny\bfseries, violet!70!black, anchor=north] at (1.4,-1.3) {SIGReg};
    \draw[violet!40, thin, ->] (1.4,-1.3) -- (1.15,-1.05);
\end{scope}

\end{tikzpicture}

%% file: appendix.tex
 
\section*{Appendix Overview}

\begin{center}
\renewcommand{\arraystretch}{1.15}
\begin{tabular}{lll}
\toprule
& \textbf{Contents} & \textbf{Page} \\
\midrule
Appendix A & Proof of Thm.~\ref{thm:main} (linear identifiability), incl.\ Hermite background & \pageref{app:proof-forward} \\
Appendix B & Proof of Thm.~\ref{thm:converse} (Gaussian uniqueness) & \pageref{app:converse-proof} \\
Appendix C & Proof of Thm.~\ref{thm:approx} (approximate identifiability) & \pageref{app:stability} \\
Appendix D & Proof and discussion of Thm.~\ref{thm:planning} (optimal latent planning) & \pageref{app:planning} \\
Appendix E & Alternative proof via Dirichlet energy & \pageref{app:dirichlet} \\
Appendix F & Prior work: connection to Slow Feature Analysis & \pageref{app:prior-work} \\
Appendix G & Lean~4 formal verification & \pageref{app:lean} \\
Appendix H & Experimental details and additional results & \pageref{app:experiment} \\
\bottomrule
\end{tabular}
\end{center}

{\label{glossary}
\begin{center}
\renewcommand{\arraystretch}{1.15}
\begin{tabular}{cl}
\toprule
\textbf{Symbol} & \textbf{Description} \\
\midrule
\multicolumn{2}{l}{\textit{World Model}} \\
$z, z' \in \mathbb{R}^N$ & True latent variables and positive pair (second view) \\
$z_i$ & $i$-th coordinate of $z$, one scalar latent variable \\
$N$ & Latent dimension \\
$M$ & Encoder output dimension ($M = N$ in theorems; App.~\ref{app:dims}) \\
$m_i$ & Deterministic part of the additive-noise transition, $z'_i = m_i(z_i) + \eta_i$ \\
$\rho \in (0,1)$ & Autocorrelation of the Ornstein--Uhlenbeck transition, $m_i(z_i) = \rho\, z_i$ \\
$\eta \sim \mathcal{N}(0, I_N)$ & Independent transition noise \\
$K > 0$ & Diffusion coefficient (Sturm--Liouville setting) \\
\midrule
\multicolumn{2}{l}{\textit{Maps}} \\
$g$ & Unknown nonlinear mixing (generative) function; $x = g(z)$ \\
$f$ & Learned encoder; $y = f(x)$ \\
$h = f \circ g$ & Composed map from latent space to representation space \\
$J_h(z)$ & Jacobian matrix of $h$ at $z$ \\
\midrule
\multicolumn{2}{l}{\textit{Linear algebra}} \\
$O(N)$ & Orthogonal group (matrices with $QQ^\top = I_N$) \\
$Q \in O(N)$ & Orthogonal recovery matrix (exact identifiability) \\
\midrule
\multicolumn{2}{l}{\textit{Losses and objectives}} \\
$\mathcal{L}(h)$ & Alignment loss: $\mathbb{E}[\|h(z') - h(z)\|^2]$ \\
$\mathcal{L}_{\mathrm{inv}}$ & Empirical invariance (alignment) loss \\
$\mathcal{L}_{\mathrm{SIG}}$ & SIGReg Gaussianity regularizer \\
$\lambda$ & Regularization weight balancing $\mathcal{L}_{\mathrm{SIG}}$ and $\mathcal{L}_{\mathrm{inv}}$ \\
\midrule
\multicolumn{2}{l}{\textit{Hermite polynomials and spectral decomposition}} \\
$\mathrm{He}_k(x)$ & Probabilist's Hermite polynomial of degree $k$ \\
$\hat{h}_k(x)$ & Normalized Hermite polynomial: $\mathrm{He}_k(x)/\sqrt{k!}$ \\
$\alpha \in \mathbb{N}^N$ & Multi-index $(\alpha_1, \ldots, \alpha_N)$; $|\alpha| = \alpha_1 + \cdots + \alpha_N$ is the total degree \\
$H_\alpha(z)$ & Multivariate Hermite basis function: $\prod_j \hat{h}_{\alpha_j}(z_j)$ \\
$c_\alpha^{(i)}$ & Hermite coefficient of $h_i$ at multi-index $\alpha$ \\
$w_d^{(i)}$ & Variance fraction of $h_i$ at degree $d$: $\sum_{|\alpha|=d} (c_\alpha^{(i)})^2$ \\
$T$ & Transition operator: $(T\varphi)(z) = \mathbb{E}[\varphi(z') \mid z]$ \\
\midrule
\multicolumn{2}{l}{\textit{Approximate identifiability}} \\
$\delta \geq 0$ & Alignment gap (excess loss above the linear optimum) \\
$\varepsilon \geq 0$ & Covariance deviation: $\|\mathrm{Cov}(h(z)) - I_N\|_F$ \\
$D$ & Normalized alignment gap: $\delta / (2\rho(1-\rho))$ \\
\bottomrule
\end{tabular}
\end{center}
}

\section{Proof of Theorem~\ref{thm:main} (Forward Direction)}\label{app:proof-forward}

\begin{theoremrestate}[Theorem~\ref{thm:main} (LeJEPA Linear Identifiability)]
Consider the Gaussian world (Sec.~\ref{sec:gaussian-wm}). Let $h: \mathbb{R}^N \to \mathbb{R}^N$ be any measurable map with $h(z) \sim \mathcal{N}(0, I_N)$. Then $\mathcal{L}(h) \geq 2(1-\rho)\,N$, with equality if and only if $h(z) = Qz$ for some orthogonal $Q \in O(N)$. At any such optimum, $h(z') \mid h(z) \sim \mathcal{N}(\rho\, h(z), (1-\rho^2)\,I_N)$.
\end{theoremrestate}

\subsection{Intuition: Why Should This Force Linearity?}

Before the formal proof, let us build intuition for why linearity is forced.
A linear orthogonal map $h(z) = Qz$ clearly preserves the Gaussian and achieves correlation $\rho$ per component, giving $\mathcal{L} = 2(1-\rho)N$.
\textit{Could a nonlinear map do better?}

Here is a suggestive information-theoretic argument.
The variables $z$ and $z'$ are jointly Gaussian with correlation $\rho$, so $I(z; z') = -\frac{N}{2}\ln(1-\rho^2)$.
For Gaussians, mutual information depends only on the correlation.
If a nonlinear $h$ achieved higher correlation while keeping Gaussian marginals, and if the joint $(h(z), h(z'))$ were also Gaussian, then $I(h(z); h(z')) > I(z; z')$, violating the data processing inequality.
So nonlinear maps cannot beat linear ones.

This argument has the right conclusion but two gaps: (1)~a nonlinear $h$ can produce a non-Gaussian \textit{joint} even when both marginals are Gaussian, so the mutual information formula does not apply; (2)~even if nonlinear maps cannot beat linear ones, perhaps they can \textit{tie}.
To close both gaps, we need to understand precisely how nonlinearity interacts with the transition between positive pairs.
The right tool turns out to be the Hermite polynomials, the natural orthogonal basis for functions of Gaussian variables.
Just as Fourier analysis decomposes a signal into frequencies, the Hermite expansion decomposes any function into a linear part, a quadratic part, a cubic part, and so on.
The key payoff is that the transition between positive pairs acts on each degree separately, and attenuates higher degrees more.
This lets us compute the correlation of any function in closed form, closing both gaps simultaneously.

\subsection{Hermite Polynomial Background}\label{app:hermite-details}

The proof rests on the Hermite polynomials, the natural orthogonal basis for functions of Gaussian random variables.
This subsection collects the definitions and properties used throughout the paper.

\paragraph{Definitions and orthogonality.}
The probabilist's Hermite polynomials $\mathrm{He}_k(x)$ are defined by the Rodrigues formula:
\begin{equation}
    \mathrm{He}_k(x) = (-1)^k\, e^{x^2/2}\, \frac{d^k}{dx^k}\, e^{-x^2/2}, \qquad k = 0, 1, 2, \ldots
\end{equation}
The first several are:
\begin{equation}
    \mathrm{He}_0(x) = 1, \quad \mathrm{He}_1(x) = x, \quad \mathrm{He}_2(x) = x^2 - 1, \quad \mathrm{He}_3(x) = x^3 - 3x
\end{equation}
Each $\mathrm{He}_k$ is a degree-$k$ polynomial.
Note that $\mathrm{He}_2(x) = x^2 - 1$ includes a correction term that makes it orthogonal to the constant and linear terms under the Gaussian measure.
More generally, these polynomials satisfy the orthogonality relation under the standard Gaussian measure $\gamma$ with density $p(x) = (2\pi)^{-1/2}e^{-x^2/2}$:
\begin{equation}
    \mathbb{E}_{x \sim \gamma}\!\left[\mathrm{He}_j(x)\,\mathrm{He}_k(x)\right] = k!\,\delta_{jk}.
\end{equation}
Normalizing by $\hat{h}_k(x) := \mathrm{He}_k(x)/\sqrt{k!}$ gives an orthonormal basis of $L^2(\mathbb{R}, \gamma)$: $\mathbb{E}[\hat{h}_j(x)\,\hat{h}_k(x)] = \delta_{jk}$.
The completeness of this basis means that any $f \in L^2(\mathbb{R}, \gamma)$ can be uniquely expanded as $f(x) = \sum_{k=0}^\infty a_k \hat{h}_k(x)$ with $\sum_k a_k^2 = \mathbb{E}[f(x)^2] < \infty$ (Parseval's identity).

\paragraph{Multivariate extension.}
For $z = (z_1, \ldots, z_N) \in \mathbb{R}^N$ with independent standard Gaussian components, the multivariate Hermite basis is constructed as tensor products.
For a multi-index $\alpha = (\alpha_1, \ldots, \alpha_N) \in \mathbb{N}^N$ with total degree $|\alpha| = \alpha_1 + \cdots + \alpha_N$, define:
\begin{equation}
    H_\alpha(z) = \prod_{j=1}^N \hat{h}_{\alpha_j}(z_j).
\end{equation}
The family $\{H_\alpha\}_{\alpha \in \mathbb{N}^N}$ is an orthonormal basis of $L^2(\mathbb{R}^N, \gamma_N)$, where $\gamma_N$ is the standard Gaussian measure on $\mathbb{R}^N$.
The degree $|\alpha|$ captures the ``order of nonlinearity'':
\begin{itemize}
    \item Degree 0: $H_0(z) = 1$. The constant function.
    \item Degree 1: $H_{e_j}(z) = z_j$ for $j = 1, \ldots, N$. The $N$ linear coordinate functions.
    \item Degree 2: terms like $z_iz_j$ (for $i \neq j$) and $(z_j^2 - 1)/\sqrt{2}$. Quadratic nonlinearities.
    \item Degree $d$: captures $d$-th order nonlinear structure. 
\end{itemize}

\paragraph{Hermite expansion of measure-preserving maps.}
Any component $h_i: \mathbb{R}^N \to \mathbb{R}$ of a map satisfying $h(z) \sim \mathcal{N}(0, I_N)$ has $\mathbb{E}[h_i(z)] = 0$ and $\mathbb{E}[h_i(z)^2] = 1$.
By completeness, it expands as:
\begin{equation}\label{eq:app-hermite-expand}
    h_i(z) = \sum_{|\alpha| \geq 1} c_\alpha^{(i)}\, H_\alpha(z), \qquad \sum_{|\alpha| \geq 1} (c_\alpha^{(i)})^2 = 1.
\end{equation}
The constant term ($|\alpha| = 0$) vanishes because $\mathbb{E}[h_i] = 0$.
The normalization $\sum (c_\alpha^{(i)})^2 = 1$ follows from Parseval's identity and $\mathbb{E}[h_i^2] = 1$.
Grouping by total degree defines the \emph{variance at degree $d$}:
\begin{equation}
    w_d^{(i)} = \sum_{|\alpha|=d} (c_\alpha^{(i)})^2, \qquad \sum_{d \geq 1} w_d^{(i)} = 1.
\end{equation}
If $h_i$ is a linear function of $z$, then $w_1^{(i)} = 1$ and $w_d^{(i)} = 0$ for all $d \geq 2$.
If $h_i$ has any nonlinear component, then $w_d^{(i)} > 0$ for some $d \geq 2$.

\subsection{Proof}


\paragraph{Setup and notation.}
Since $z \sim \mathcal{N}(0, I_N)$, the multivariate Hermite polynomials $\{H_\alpha\}_{\alpha \in \mathbb{N}^N}$ form an orthonormal basis of $L^2(\mathbb{R}^N, \gamma_N)$ (Appendix~\ref{app:hermite-details}).
Each component $h_i$ has $\mathbb{E}[h_i(z)] = 0$ and $\mathbb{E}[h_i(z)^2] = 1$ by the Gaussianity constraint, and admits the expansion
\begin{equation}\label{eq:app-hermite-expand-pf}
    h_i(z) = \sum_{|\alpha| \geq 1} c_\alpha^{(i)}\, H_\alpha(z), \qquad \sum_{|\alpha| \geq 1} (c_\alpha^{(i)})^2 = 1.
\end{equation}
Grouping by total degree $d = |\alpha|$, let $w_d^{(i)} = \sum_{|\alpha|=d} (c_\alpha^{(i)})^2$ be the fraction of variance at degree $d$.
Then $w_1^{(i)} + w_2^{(i)} + w_3^{(i)} + \cdots = 1$.
If $h_i$ is linear, then $w_1^{(i)} = 1$; any nonlinearity spills variance into $d \geq 2$.
By equation~\eqref{eq:loss-expand}, minimizing $\mathcal{L}$ is equivalent to maximizing $\sum_i \mathbb{E}[h_i(z') h_i(z)]$.
The proof establishes that each term satisfies $\mathbb{E}[h_i(z') h_i(z)] \leq \rho$, with equality if and only if $h_i$ is linear.
The argument has three main steps.
First, we need to know what the transition $z \to z'$ does to each Hermite component individually.
Second, we need the cross-correlation structure between all components across the two views: does the degree-2 part of $h_i(z')$ correlate with the degree-1 part of $h_i(z)$?
Third, we combine these to bound the total correlation any function can achieve.

\paragraph{Step 1: The transition attenuates each Hermite degree by $\rho^d$.}
We show that $\mathbb{E}_\eta[\mathrm{He}_k(z')] = \rho^k\,\mathrm{He}_k(z)$ for $z' = \rho z + \sqrt{1-\rho^2}\,\eta$ with $\eta \sim \mathcal{N}(0,1)$ independent of $z \sim \mathcal{N}(0,1)$.

We use the exponential generating function for the Hermite polynomials:
\begin{equation}\label{eq:app-genfun-pf}
    \sum_{k=0}^\infty \frac{t^k}{k!}\,\mathrm{He}_k(x) = \exp\!\left(tx - \frac{t^2}{2}\right).
\end{equation}
This identity encodes all Hermite polynomials into a single exponential, allowing us to compute the effect of noise on all degrees simultaneously.

Evaluate~\eqref{eq:app-genfun-pf} at $x = z' = \rho z + \sqrt{1-\rho^2}\,\eta$:
\begin{equation}
    \sum_{k=0}^\infty \frac{t^k}{k!}\,\mathrm{He}_k(z') = \exp\!\left(t(\rho z + \sqrt{1-\rho^2}\,\eta) - \frac{t^2}{2}\right).
\end{equation}
Since $\eta$ is independent of $z$, the right side factors:
\begin{equation}
    \exp\!\left(t\rho z - \frac{t^2}{2}\right) \cdot \exp\!\left(t\sqrt{1-\rho^2}\,\eta\right).
\end{equation}
The first factor depends only on $z$ and comes out of $\mathbb{E}_\eta$.
For the second factor, we use the Gaussian moment generating function: if $\eta \sim \mathcal{N}(0,1)$, then $\mathbb{E}[e^{s\eta}] = e^{s^2/2}$ for any $s$.
Setting $s = t\sqrt{1-\rho^2}$:
\begin{equation}
    \mathbb{E}_\eta\!\left[\exp\!\left(t\sqrt{1-\rho^2}\,\eta\right)\right] = \exp\!\left(\frac{t^2(1-\rho^2)}{2}\right).
\end{equation}
Multiplying the two pieces, the exponents combine:
\begin{equation}
    t\rho z - \frac{t^2}{2} + \frac{t^2(1-\rho^2)}{2} = t\rho z - \frac{t^2}{2} + \frac{t^2}{2} - \frac{t^2\rho^2}{2} = t\rho z - \frac{(t\rho)^2}{2}.
\end{equation}
So we obtain:
\begin{equation}\label{eq:app-after-noise-pf}
    \sum_{k=0}^\infty \frac{t^k}{k!}\,\mathbb{E}_\eta\!\left[\mathrm{He}_k(z')\right] = \exp\!\left((\rho t) z - \frac{(\rho t)^2}{2}\right).
\end{equation}
The right side of~\eqref{eq:app-after-noise-pf} has exactly the form of the generating function~\eqref{eq:app-genfun-pf} with $t$ replaced by $\rho t$:
\begin{equation}
    \exp\!\left((\rho t) z - \frac{(\rho t)^2}{2}\right) = \sum_{k=0}^\infty \frac{(\rho t)^k}{k!}\,\mathrm{He}_k(z) = \sum_{k=0}^\infty \frac{t^k}{k!}\,\rho^k\,\mathrm{He}_k(z).
\end{equation}
Matching coefficients of $t^k/k!$:
\begin{equation}\label{eq:contraction-result}
    \mathbb{E}_\eta\!\left[\mathrm{He}_k(z')\right] = \rho^k\,\mathrm{He}_k(z).
\end{equation}
Since $0 < \rho < 1$, we have $\rho^2 < \rho$, $\rho^3 < \rho^2$, and so on.
\textit{The linear component ($k=1$) survives the transition best; every higher-degree component is strictly more attenuated.}

This contraction is specific to Gaussians.
The proof succeeds because the Gaussian MGF ($e^{s^2/2}$) partially cancels the $e^{-t^2/2}$ in the Hermite generating function, leaving $e^{-t^2\rho^2/2}$, which is the generating function again with $t \to \rho t$.
This replacement is where the $\rho^k$ contraction originates.
For non-Gaussian distributions, the transition operator still has a spectral decomposition with ordered eigenvalues (see Appendix~\ref{app:prior-work}), but the eigenfunctions are no longer Hermite polynomials and the eigenvalues no longer decay geometrically. In particular, the first eigenfunction is monotonic but generally nonlinear, which is why monotonic (not linear) identifiability holds in the general case.

\paragraph{Step 2: Different Hermite components do not interact across views.}
Step~1 tells us what happens to each Hermite component in isolation.
But to compute the correlation $\mathbb{E}[h_i(z') h_i(z)]$ for an arbitrary function $h_i$, we also need to know whether different components interact across views: does the degree-2 part of $h_i(z')$ correlate with the degree-1 part of $h_i(z)$?

The answer is no.
We show that $\mathbb{E}[H_\alpha(z')\, H_\beta(z)] = \rho^{|\alpha|}\,\delta_{\alpha\beta}$ (Mehler's formula).
The $\delta_{\alpha\beta}$ means that cross-terms vanish: the degree-1 part of $h(z')$ only correlates with the degree-1 part of $h(z)$, the degree-2 part only with degree-2, and so on.
Each degree-$d$ component contributes correlation $\rho^d$.
The Hermite basis \textit{diagonalizes} the transition.

In the univariate case, using the law of iterated expectations:
\begin{equation}
    \mathbb{E}\!\left[\hat{h}_k(z')\,\hat{h}_j(z)\right] = \mathbb{E}_z\!\left[\mathbb{E}_\eta\!\left[\hat{h}_k(z')\right]\,\hat{h}_j(z)\right].
\end{equation}
By the contraction property~\eqref{eq:contraction-result}, $\mathbb{E}_\eta[\hat{h}_k(z')] = \rho^k\,\hat{h}_k(z)$.
Substituting:
\begin{equation}
    \mathbb{E}\!\left[\hat{h}_k(z')\,\hat{h}_j(z)\right] = \rho^k\,\mathbb{E}_z\!\left[\hat{h}_k(z)\,\hat{h}_j(z)\right] = \rho^k\,\delta_{jk},
\end{equation}
where the last step uses orthonormality of the $\hat{h}_k$ under $\gamma$.

In the multivariate case, since $z$ has independent components and the noise in the Ornstein--Uhlenbeck channel acts independently on each coordinate, the multivariate Hermite products factorize:
\begin{equation}\label{eq:mehler-result}
    \mathbb{E}\!\left[H_\alpha(z')\, H_\beta(z)\right] = \prod_{j=1}^N \mathbb{E}\!\left[\hat{h}_{\alpha_j}(z_j')\,\hat{h}_{\beta_j}(z_j)\right] = \prod_{j=1}^N \rho^{\alpha_j}\,\delta_{\alpha_j \beta_j} = \rho^{|\alpha|}\,\delta_{\alpha\beta}.
\end{equation}

\paragraph{Step 3: Nonlinearity strictly reduces correlation.}
Now we can compute the total correlation of \textit{any} function through the transition.
Since the Hermite basis diagonalizes the problem, the answer is simply a weighted sum over degrees, and because $\rho^d$ decreases with $d$, putting variance on higher degrees always costs correlation.

Expanding $h_i$ in the Hermite basis~\eqref{eq:app-hermite-expand-pf} and applying Mehler's formula~\eqref{eq:mehler-result}:
\begin{equation}
    \mathbb{E}\!\left[h_i(z')\, h_i(z)\right] = \sum_{|\alpha| \geq 1} \sum_{|\beta| \geq 1} c_\alpha^{(i)} c_\beta^{(i)} \,\mathbb{E}\!\left[H_\alpha(z')\, H_\beta(z)\right] = \sum_{|\alpha| \geq 1} (c_\alpha^{(i)})^2\,\rho^{|\alpha|}.
\end{equation}
Note that all cross-terms vanish ($\delta_{\alpha\beta} = 0$ for $\alpha \neq \beta$), and each surviving term contributes its squared coefficient times $\rho^{|\alpha|}$.

Grouping by degree $d = |\alpha|$ with $w_d = \sum_{|\alpha|=d} (c_\alpha^{(i)})^2$:
\begin{equation}
    \mathbb{E}\!\left[h_i(z')\, h_i(z)\right] = w_1 \cdot \rho + w_2 \cdot \rho^2 + w_3 \cdot \rho^3 + \cdots
\end{equation}
with $w_d \geq 0$ and $w_1 + w_2 + w_3 + \cdots = 1$.
This is a weighted average of $\rho, \rho^2, \rho^3, \ldots$ with weights summing to 1.
Since $0 < \rho < 1$, each $\rho^d \leq \rho$ with strict inequality for $d \geq 2$:
\begin{equation}
    w_1 \cdot \rho + w_2 \cdot \rho^2 + w_3 \cdot \rho^3 + \cdots \leq w_1 \cdot \rho + w_2 \cdot \rho + w_3 \cdot \rho + \cdots = \rho \cdot (w_1 + w_2 + \cdots) = \rho.
\end{equation}

Equality requires $w_d \cdot \rho^d = w_d \cdot \rho$ for every $d$, which means $w_d = 0$ for all $d \geq 2$ (since $\rho^d < \rho$ for $d \geq 2$).
So all the variance is at degree 1: $w_1 = 1$.
The degree-1 Hermite basis consists of the coordinate functions $\{z_1, \ldots, z_N\}$, so:
\begin{equation}
    h_i(z) = \sum_{j=1}^N c_{e_j}^{(i)}\, z_j = \langle q_i, z\rangle
\end{equation}
for some vector $q_i$ with $\|q_i\|^2 = w_1 = 1$.
That is, $h_i$ must be a linear function.

\textit{No nonlinear map can match the correlation of a linear map.}
This closes both gaps from the information-theoretic argument.
That argument failed because it assumed the joint distribution of $(h(z), h(z'))$ is Gaussian, which need not hold for nonlinear $h$.
The Hermite analysis avoids this entirely: Mehler's formula computes $\mathbb{E}[h_i(z') h_i(z)]$ exactly for any $h_i \in L^2(\gamma_N)$, using only the joint Gaussianity of $(z, z')$, not of $(h_i(z), h_i(z'))$.
The prior argument also left open the possibility that a nonlinear map could match the correlation of a linear one.
The strict inequality above rules this out: any nonlinearity ($w_d > 0$ for some $d \geq 2$) strictly reduces correlation.

\paragraph{Step 4: Assembly.}
Summing the correlation bound over all $N$ components:
\begin{equation}
    \sum_{i=1}^N \mathbb{E}[h_i(z') h_i(z)] \leq \rho N,
\end{equation}
which by~\eqref{eq:loss-expand} gives $\mathcal{L}(h) \geq 2(1-\rho)N$.

At equality, every $h_i$ is linear: $h_i(z) = \langle q_i, z\rangle$ with $\|q_i\| = 1$.
So $h(z) = Qz$ for some matrix $Q$ with unit-norm rows.
The Gaussianity constraint requires $\mathrm{Cov}(h(z)) = Q Q^\top = I_N$, so $Q$ is orthogonal.
Conversely, any orthogonal $Q$ achieves $\sum_i \mathbb{E}[\langle Q_i, z'\rangle \langle Q_i, z\rangle] = \rho N$, giving equality.

\paragraph{Step 5: Recovery of the transition.}
At the optimum, $h(z') = Qz' = \rho\,Qz + \sqrt{1-\rho^2}\,Q\eta = \rho\,h(z) + \sqrt{1-\rho^2}\,Q\eta$.
Since $Q$ is orthogonal and $\eta \sim \mathcal{N}(0, I_N)$, we have $Q\eta \sim \mathcal{N}(0, I_N)$ independently of $h(z)$.
Therefore $h(z') \mid h(z) \sim \mathcal{N}(\rho\,h(z), (1-\rho^2)I_N)$, which is the same conditional as $z' \mid z$.
The encoder recovers the latent variables and their dynamics up to a global rotation. \qed

\newpage
\section{Proof of Theorem~\ref{thm:converse} (Gaussian Uniqueness)}\label{app:converse-proof}

\begin{theoremrestate}[Theorem~\ref{thm:converse} (Gaussian Uniqueness)]
Consider any world satisfying Assumptions~\ref{assumptions}. Suppose every minimizer of \eqref{eq:loss-expand} with $\mathrm{Cov}(h(z)) = I_N$ is linear, $h(z) = Qz$. Then $z$ is Gaussian.
\end{theoremrestate}

Since the latent variables are independent, the transition operator separates, and it suffices to prove the result for a single component $z_i$ whose stationary distribution has full support on $\mathbb{R}$.

\subsection{Background}

The forward result (Thm.~\ref{thm:main}) shows that the optimal encoder extracts the first few eigenfunctions of the transition operator $T$ (Section~\ref{sec:spectral}): these are the functions of the latent variables with the highest temporal autocorrelation.
If the first (slowest) non-constant eigenfunction is linear, the encoder recovers the latent variable itself; if it is nonlinear, the encoder recovers a nonlinear transformation of the latent variable, and linear identifiability fails.
The question is therefore: for which latent variable distributions is the first eigenfunction linear?

The proof uses the Sturm--Liouville characterization of these eigenfunctions \citep{sprekeler2014}. Assumption (iii) ($\eta_i$ independent of $z_i$) implies the noise variance does not depend on the state; in continuous time, this is the constant-diffusion regime in which Sturm--Liouville machinery applies. The transition operator $T$ has an infinitesimal generator $\mathcal{D}$ that shares the same eigenfunctions; $\mathcal{D}$ is a second-order differential operator whose structure makes the eigenvalue problem tractable (see Appendix~\ref{app:prior-work} for the full connection to Slow Feature Analysis).
For a scalar latent variable evolving under constant diffusion $K > 0$ with stationary density $p(z_i) > 0$, the generator, written in self-adjoint form, acts on test functions $\varphi$ as
\begin{equation}\label{eq:sfa-operator}
    \mathcal{D}\varphi = \frac{1}{p(z_i)}\frac{d}{dz_i}
    \Big[K\,p(z_i)\,\varphi'(z_i)\Big].
\end{equation}
The eigenfunctions $\varphi_k$ are properties of the latent process, not of the encoder $h$; the encoder inherits its structure from them.

When the score function diverges ($(\log p)'(z_i) \to \mp\infty$ as $z_i \to \pm\infty$), the operator has a purely discrete spectrum $0 < \lambda_1 < \lambda_2 < \cdots$ with eigenfunctions $\varphi_1, \varphi_2, \ldots$ forming a complete orthogonal basis of $L^2(\mathbb{R}, p\,dz)$.
The eigenvalues are ordered by increasing ``speed'': $\varphi_1$ is the slowest non-constant feature, $\varphi_2$ the next slowest, and so on.
By classical Sturm--Liouville oscillation theory, $\varphi_k$ has exactly $k-1$ interior zeros.
In particular, $\varphi_1$ is always monotonic, so one always gets identifiability up to a monotonic transformation.
Linear identifiability requires $\varphi_1$ to be affine, and since an affine function is monotonic (zero interior zeros), only $\varphi_1$ can be affine: no higher eigenfunction can be.
(For heavy-tailed densities such as Laplace, $\mathcal{D}$ develops continuous spectrum and the discrete picture does not apply; this does not affect our result, since assuming an affine eigenfunction exists forces $p$ to be Gaussian, which has discrete spectrum.)
For the full connection between the Sturm--Liouville framework and Slow Feature Analysis, see Appendix~\ref{app:prior-work}.

\subsection{Proof}

The key observation is that the operator $\mathcal{D}$ encodes the score function $(\log p)'$ of the latent variable distribution.
Demanding that the first eigenfunction is affine forces the score to be linear in $z_i$, and a linear score uniquely characterizes the Gaussian.

Suppose $\varphi(z_i) = a z_i + b$ is an eigenfunction of $\mathcal{D}$ with eigenvalue $\lambda > 0$.
Since $\varphi' = a$ is constant, the operator simplifies:
\begin{equation}
    \mathcal{D}[a z_i + b] = a\,K(\log p)'(z_i).
\end{equation}
The eigenfunction equation $\mathcal{D}\varphi = -\lambda\varphi$ requires $a\,K(\log p)'(z_i) = -\lambda(a z_i + b)$, giving
\begin{equation}
    K(\log p)'(z_i) = -\lambda z_i - \lambda b/a.
\end{equation}
Integrating both sides:
\begin{equation}
    \log p(z_i) = -\frac{\lambda}{2K}
    \left(z_i + \frac{b}{a}\right)^2 + C,
\end{equation}
which is a Gaussian density with mean $-b/a$.
No other density satisfies this equation.
Under the zero-mean convention $b = 0$, and $a$ is fixed by unit-variance normalization. \qed

\paragraph{Remark: the OU transition is a modeling choice.}
Gaussian marginals and Assumptions~\ref{assumptions} do not by themselves force the OU transition~\eqref{eq:dgp}. Any map $m$ that preserves $\mathcal{N}(0,1)$ yields a stationary transition, e.g., $m(z) = z$ for $|z| > 1$ and $m(z) = -z$ otherwise, with $\eta = 0$; composing such a map with the OU channel, $z' = \rho\, m(z) + \sqrt{1-\rho^2}\,\eta$, gives examples with $\eta \neq 0$ (we thank Siddharth Ancha for pointing this out). Thm.~\ref{thm:main} relies on the OU form through its Hermite spectrum, and such transitions have different eigenfunctions. The OU form is, however, forced in the continuous-time setting that motivates Assumption (iii) (Sec.~\ref{sec:world-model}): a diffusion $dz = b(z)\,dt + \sigma\,dW$ with constant volatility has stationary density $p(z) \propto \exp\big(\tfrac{2}{\sigma^2}\int^z b(u)\,du\big)$, which is Gaussian if and only if the drift $b$ is linear, i.e., if and only if the process is OU. The counterexamples above are discrete-time maps that no such diffusion produces over a short interval.

\section{Proof of Theorem~\ref{thm:approx} (Approximate Identifiability)}\label{app:stability}

\begin{theoremrestate}[Theorem~\ref{thm:approx} (Approximate Identifiability)]
Consider the Gaussian world (\ref{sec:gaussian-wm}). Let $h: \mathbb{R}^N \to \mathbb{R}^N$ be measurable with $\mathbb{E}[h(z)] = 0$, satisfying:
\textit{Approximate alignment:} $\mathcal{L}(h) \leq 2(1-\rho)\,\mathrm{tr}(\mathrm{Cov}(h(z))) + \delta$;
\textit{Approximate whitening:} $\|\mathrm{Cov}(h(z)) - I_N\|_F \leq \varepsilon$.
Define $D = \delta/(2\rho(1-\rho))$. Then there exists $Q \in O(N)$ such that
\begin{equation}
    \mathbb{E}\!\left[\|h(z) - Qz\|^2\right] \;\leq\; D \;+\; (\varepsilon + D)^2
\end{equation}
\end{theoremrestate}

We prove the stability bound in three steps: (1) the alignment gap controls
nonlinear energy, (2) the whitening constraint forces the linear part
near orthogonality, and (3) orthogonality of Hermite degrees gives a
clean Pythagorean decomposition.

\paragraph{Setup and notation.}
Since $z \sim \mathcal{N}(0, I_N)$, each zero-mean component $h_i$ of the encoder admits a Hermite expansion (Appendix~\ref{app:hermite-details}):
\begin{equation}
    h_i(z) \;=\; \sum_{|\alpha| \geq 1} c_\alpha^{(i)}\, H_\alpha(z),
\end{equation}
where the constant term vanishes by the zero-mean assumption.
We split this into a linear part and a nonlinear residual.

The linear part is $Mz$, where $M \in \mathbb{R}^{N \times N}$ is the matrix of correlations between encoder outputs and latent variables:
\begin{equation}
    M_{ij} \;=\; \mathbb{E}\bigl[h_i(z)\, z_j\bigr].
\end{equation}
This extracts exactly the degree-1 Hermite coefficients, since all higher-degree terms are orthogonal to the coordinate functions $z_j$ under the Gaussian measure.

The nonlinear residual $\nu(z) = h(z) - Mz$ contains everything at Hermite degree $\geq 2$.
By construction, $\mathbb{E}[\nu(z)\, z^\top] = 0$: the residual is uncorrelated with the latent variables.
We define the total nonlinear energy as
\begin{equation}
    W_{\mathrm{nl}} \;=\; \mathbb{E}\bigl[\|\nu(z)\|^2\bigr].
\end{equation}

Since the linear and nonlinear parts live on orthogonal Hermite subspaces, the total variance decomposes without cross terms:
\begin{equation}\label{eq:parseval-split}
    \mathrm{tr}\bigl(\mathrm{Cov}(h(z))\bigr)
    \;=\; \|M\|_F^2 \;+\; W_{\mathrm{nl}}.
\end{equation}
This Pythagorean decomposition is the key identity underlying Step~3 below.

\subsection{Step 1: Alignment gap controls nonlinear energy}

Recall that the alignment loss decomposes as
\begin{equation}
    \mathcal{L}(h) \;=\; 2N \;-\; 2\sum_{i=1}^N
    \mathbb{E}\!\left[h_i(z')\, h_i(z)\right],
\end{equation}
where $(z, z')$ is a pair from the OU process with autocorrelation $\rho$.
Mehler's formula (equation~\eqref{eq:mehler-result} in Appendix~\ref{app:proof-forward}) states that Hermite polynomials
diagonalize the OU conditional expectation:
\begin{equation}
    \mathbb{E}\bigl[H_\alpha(z')\,\big|\,z\bigr]
    \;=\; \rho^{|\alpha|}\, H_\alpha(z),
\end{equation}
so each Hermite degree $|\alpha|$ is an eigenspace with eigenvalue
$\rho^{|\alpha|}$. Substituting the Hermite expansion of $h$ into the
cross-correlation $\mathbb{E}[h_i(z')\,h_i(z)]$ and applying this
eigenstructure yields an exact decomposition over degrees:
\begin{equation}\label{eq:loss-hermite}
    \mathcal{L}(h) \;=\; 2\sum_{i=1}^N \sum_{|\alpha| \geq 1}
    (1 - \rho^{|\alpha|})\,(c_\alpha^{(i)})^2.
\end{equation}
The weight $1 - \rho^{|\alpha|}$ increases with degree: higher-order
components are penalized more heavily because they decorrelate faster
under the OU transition.

We now split the sum by degree. At degree~1 the weight is exactly
$1 - \rho$. At degree $|\alpha| \geq 2$, the bound
$\rho^{|\alpha|} \leq \rho^2$ (valid for $0 < \rho < 1$) gives
$1 - \rho^{|\alpha|} \geq 1 - \rho^2 = (1-\rho)(1+\rho)$, so
\begin{equation}
    \mathcal{L}(h) \;\geq\; 2(1-\rho)\,\|M\|_F^2
    \;+\; 2(1-\rho)(1+\rho)\, W_{\mathrm{nl}}.
\end{equation}
Substituting the Pythagorean decomposition
$\|M\|_F^2 = \mathrm{tr}(\mathrm{Cov}(h(z))) - W_{\mathrm{nl}}$
from~\eqref{eq:parseval-split} and collecting terms:
\begin{align}
    \mathcal{L}(h)
    &\;\geq\; 2(1-\rho)\bigl[\mathrm{tr}(\mathrm{Cov}(h(z)))
        - W_{\mathrm{nl}}\bigr]
        + 2(1-\rho)(1+\rho)\, W_{\mathrm{nl}} \notag\\
    &\;=\; 2(1-\rho)\,\mathrm{tr}(\mathrm{Cov}(h(z)))
        + 2(1-\rho)\bigl[(1+\rho) - 1\bigr]\, W_{\mathrm{nl}} \notag\\
    &\;=\; 2(1-\rho)\,\mathrm{tr}(\mathrm{Cov}(h(z)))
        + 2\rho(1-\rho)\, W_{\mathrm{nl}}.
\end{align}

To extract a bound on $W_{\mathrm{nl}}$, we need an upper bound on
$\mathcal{L}(h)$. An encoder achieving exact alignment would satisfy
$\mathcal{L}(h) = 2(1-\rho)\,\mathrm{tr}(\mathrm{Cov}(h(z)))$
(attained when $h$ is linear). We assume the encoder is
$\delta$-approximately aligned, meaning
\begin{equation}
    \mathcal{L}(h) \;\leq\;
    2(1-\rho)\,\mathrm{tr}(\mathrm{Cov}(h(z))) + \delta.
\end{equation}
Combining the upper and lower bounds and rearranging:
\begin{equation}\label{eq:wnl-bound}
    \boxed{W_{\mathrm{nl}} \;\leq\; \frac{\delta}{2\rho(1-\rho)}.}
\end{equation}
The alignment gap $\delta$ directly controls how much variance the
encoder can place on nonlinear Hermite components. The denominator
$2\rho(1-\rho)$ is the spectral gap between degree-1 and degree-2
eigenvalues, and governs the sensitivity of this bound.

\subsection{Step 2: Whitening constraint forces $M$ near orthogonal}
 
The covariance of $h(z)$ decomposes by Hermite orthogonality as
\begin{equation}
    \mathrm{Cov}(h(z)) = MM^\top + N,
\end{equation}
where $N_{ij} = \sum_{|\alpha| \geq 2} c_\alpha^{(i)}\, c_\alpha^{(j)}$ is the
contribution from nonlinear terms. Since $N$ is a Gram matrix (positive
semidefinite) with $\mathrm{tr}(N) = W_{\mathrm{nl}}$, and for any PSD matrix
$\|N\|_F \leq \mathrm{tr}(N)$ (all eigenvalues are non-negative, so
$\sum \lambda_i^2 \leq (\sum \lambda_i)^2$), we have
$\|N\|_F \leq W_{\mathrm{nl}}$.
 
The approximate whitening assumption gives:
\begin{equation}
    \|MM^\top - I_N\|_F
    = \|\mathrm{Cov}(h(z)) - I_N - N\|_F
    \leq \varepsilon + W_{\mathrm{nl}}.
\end{equation}
Let $M = QP$ be the polar decomposition with $Q \in O(N)$ and $P$ symmetric
positive semidefinite with singular values $\sigma_1, \ldots, \sigma_N$.
Then $MM^\top = QP^2Q^\top$, so
$\|P^2 - I_N\|_F = \|MM^\top - I_N\|_F$.
Since $P$ is PSD, each $\sigma_i \geq 0$, and
$|\sigma_i - 1| \leq |\sigma_i + 1||\sigma_i - 1| = |\sigma_i^2 - 1|$.
Therefore:
\begin{equation}\label{eq:M-Q-bound}
    \|M - Q\|_F = \|P - I_N\|_F \leq \|P^2 - I_N\|_F
    \leq \varepsilon + W_{\mathrm{nl}}.
\end{equation}
 
\subsection{Step 3: Pythagorean decomposition}
 
Write
\begin{equation}
    h(z) - Qz = \underbrace{(M - Q)z}_{\text{degree 1}} + \underbrace{\nu(z)}_{\text{degree} \geq 2}.
\end{equation}
Since $(M-Q)z$ lives entirely in Hermite degree~1 and $\nu(z)$ lives entirely
in degree $\geq 2$, orthogonality of Hermite degrees under $\gamma_N$ gives:
\begin{equation}
    \mathbb{E}[\|h(z) - Qz\|^2]
    = \mathbb{E}[\|(M-Q)z\|^2] + \mathbb{E}[\|\nu(z)\|^2].
\end{equation}
The cross term $\mathbb{E}[\nu(z)^\top (M-Q)z] = 0$ vanishes exactly.
For the first term, since $z \sim \mathcal{N}(0, I_N)$:
\begin{equation}
    \mathbb{E}[\|(M-Q)z\|^2]
    = \mathrm{tr}\!\left((M-Q)\,\mathbb{E}[zz^\top]\,(M-Q)^\top\right)
    = \|M - Q\|_F^2.
\end{equation}
Combining with~\eqref{eq:wnl-bound} and~\eqref{eq:M-Q-bound}:
\begin{equation}
    \mathbb{E}[\|h(z) - Qz\|^2]
    = \|M - Q\|_F^2 + W_{\mathrm{nl}}
    \leq (\varepsilon + W_{\mathrm{nl}})^2 + W_{\mathrm{nl}}
    \leq \left(\varepsilon + \frac{\delta}{2\rho(1-\rho)}\right)^{\!2}
    + \frac{\delta}{2\rho(1-\rho)}.
\end{equation}
This completes the proof. \qed

\paragraph{Role of the Gaussian on the latent side vs.\ the output side.}
The proof uses the Gaussian assumption on $z$ in two essential ways:
the Hermite basis and Mehler's formula (which provide the spectral
decomposition~\eqref{eq:loss-hermite}), and the geometric eigenvalue
decay $\rho^d$ (which creates the spectral gap exploited in
Step~1). By contrast, the proof uses \emph{no} distributional
assumption on $h(z)$ beyond zero mean and approximate whitening.
In particular, the full distributional Gaussianity condition
$h(z) \sim \mathcal{N}(0, I_N)$ of Thm.~\ref{thm:main}
can be relaxed to $\mathbb{E}[h(z)] = 0$ and
$\|\mathrm{Cov}(h(z)) - I_N\|_F \leq \varepsilon$ without
any change to the bound.

\section{Proof and Discussion of Theorem~\ref{thm:planning}}\label{app:planning}

\begin{theoremrestate}[Theorem~\ref{thm:planning} (Optimal Latent Planning)]
Let $h(z) = Qz$ with $Q \in O(N)$, write $\hat z := h(z)$, and consider any finite-horizon optimal control problem whose stage and terminal costs $\ell, \ell_T$ are $O(N)$-invariant in the state argument:
\begin{equation}
    \ell(Rz, a) = \ell(z, a), \qquad \ell_T(Rz) = \ell_T(z), \qquad \forall\, R \in O(N).
\end{equation}
Let $\hat p(\cdot \mid \hat z_{t-1}, a_t)$ denote the pushforward of $p$ under $h$, and let $\hat V^*, \hat a^*_{1:T}$ denote the value function and optimal plan of the $\hat z$-space problem defined by replacing $p$ with $\hat p$ and leaving costs unchanged. Then $\hat V^*\!\big(h(z_0)\big) = V^*(z_0)$ and $\hat a^*_{1:T}\!\big(h(z_0)\big) = a^*_{1:T}(z_0)$.
\end{theoremrestate}

\begin{proof}
Fix any action sequence $a_{1:T} \in \mathcal{A}^T$ and initial state $z_0$, and let $\hat z_0 = h(z_0) = Q z_0$. By construction of the pushforward, the joint law of $(\hat z_0, \hat z_1, \ldots, \hat z_T)$ under $\hat p$ equals the joint law of $(Q z_0, Q z_1, \ldots, Q z_T)$ under $p$. Applying~\eqref{eq:orth-inv-app} with $R = Q$,
\begin{equation}
    \ell(\hat z_t, a_{t+1}) \;=\; \ell(Q z_t, a_{t+1}) \;=\; \ell(z_t, a_{t+1}) \quad \text{a.s.}, \qquad \ell_T(\hat z_T) \;=\; \ell_T(z_T) \quad \text{a.s.}
\end{equation}
Summing and taking expectations,
\begin{equation}
    \mathbb{E}_{\hat p}\!\left[\sum_{t=1}^{T}\ell(\hat z_{t-1}, a_t) + \ell_T(\hat z_T) \,\Big|\, \hat z_0\right]
    \;=\; \mathbb{E}_{p}\!\left[\sum_{t=1}^{T}\ell(z_{t-1}, a_t) + \ell_T(z_T) \,\Big|\, z_0\right].
\end{equation}
Since this holds for every $a_{1:T}$, the minima coincide, giving $\hat V^*(\hat z_0) = V^*(z_0)$, and any minimizer of one side is a minimizer of the other.
\end{proof}

\subsection{Worked Applications}\label{app:planning-instances}

The invariance condition~\eqref{eq:orth-inv-app} covers a substantial part of control applications. We spell out three.

\paragraph{Goal-reaching.}
The terminal cost $\ell_T(z) = \|z - z^*\|^2$ is $O(N)$-invariant when the state and goal rotate together: $\|Rz - Rz^*\|^2 = \|z - z^*\|^2$. A planner choosing a goal $\hat z^* \in \mathbb{R}^N$ in the learned latent and minimizing $\|\hat z_T - \hat z^*\|^2$ solves the true goal-reaching problem for the goal $z^* = Q^\top \hat z^*$. The learned goal need never be mapped back to the true latent for this to work; the planner's actions are correct as a sequence, regardless of the coordinate system they are computed in.

\paragraph{Linear-quadratic regulation.}
Take linear dynamics $z_{t+1} = A z_t + B a_t + \varepsilon_t$ with $\varepsilon_t \sim \mathcal{N}(0, \Sigma)$, and quadratic costs $\ell(z,a) = z^\top W z + a^\top R a$, $\ell_T(z) = z^\top W_T z$, with $W, W_T \succeq 0$ and $R \succ 0$.
The pushforward dynamics in $\hat z$-coordinates are
\begin{equation}
    \hat z_{t+1} = (Q A Q^\top)\,\hat z_t + (Q B)\,a_t + Q\varepsilon_t, \qquad Q\varepsilon_t \sim \mathcal{N}(0, Q\Sigma Q^\top),
\end{equation}
so the LQR problem in $\hat z$-space has system matrices $(\hat A, \hat B, \hat\Sigma) = (QAQ^\top, QB, Q\Sigma Q^\top)$ and cost matrices $(\hat W, \hat W_T, R) = (QWQ^\top, QW_T Q^\top, R)$. The discrete algebraic Riccati equation
\begin{equation}
    P = A^\top P A - A^\top P B (R + B^\top P B)^{-1} B^\top P A + W
\end{equation}
is covariant under $(A, B, W, P) \mapsto (QAQ^\top, QB, QWQ^\top, QPQ^\top)$: if $P$ solves the true-latent equation, then $\hat P = QPQ^\top$ solves the $\hat z$-space equation. The optimal feedback gain $K = (R + B^\top P B)^{-1} B^\top P A$ transforms as $\hat K = KQ^\top$, and the feedback action is
\begin{equation}
    \hat a_t = -\hat K \hat z_t = -(K Q^\top)(Q z_t) = -K z_t,
\end{equation}
which is the true LQR action at every state. The value function is likewise $\hat V^*(\hat z) = \hat z^\top \hat P \hat z = z^\top P z = V^*(z)$. LQR is therefore a special case of Thm.~\ref{thm:planning} in which the invariance is inherited from the quadratic structure and the covariance of all system matrices under the rotation.

\paragraph{General rotation-invariant objectives.}
Any cost depending on the state only through quantities preserved by rotations satisfies~\eqref{eq:orth-inv-app}. Examples: norms $\|z\|$ and radial functions $f(\|z\|)$; inner products $\langle z, z' \rangle$ between pairs of states; quadratic forms $z^\top W z$ with $W$ specified in the learned coordinates (which silently corresponds to $Q^\top W Q$ in the true latent); Gaussian belief updates, whose covariance updates are covariant under rotation; and linear value functions $w^\top z$ with $w$ specified in the learned coordinates. In each case, an engineer who writes down the cost using only these operations has unwittingly written down a cost in the equivalence class of~\eqref{eq:orth-inv-app}, and Thm.~\ref{thm:planning} certifies that planning is unaffected by the ambiguity.

\subsection{Scope: encoder versus action-conditioned dynamics}\label{app:planning-scope}

Thm.~\ref{thm:planning} concerns the encoder alone. The action-conditioned transition $\hat p(\hat z' \mid \hat z, a)$ must still be learned from data \citep{maes2026leworldmodel}; our theorems do not prove that it is. What they do guarantee is that the encoder does not corrupt the learning problem for the transition: for $h(z) = Qz$, the pushforward of the true transition lies in the same function class as the original (linear transitions remain linear, Gaussian transitions remain Gaussian, and any class closed under orthogonal reparametrization of the state is equally expressive in both coordinate systems). The residue of the orthogonal ambiguity is a consistent reparametrization across the encoder, the transition model, and the cost: so long as these three components are trained or specified to be mutually consistent in the learned coordinates, planning in those coordinates is planning in the true ones.

The natural next theoretical step is to prove that the action-conditioned transition itself is identifiable. This is a genuinely different problem: the encoder identifiability proved here relies on stationarity plus Gaussianity of the latent distribution, whereas transition identifiability would trade these for an excitation condition on the action sequence (the actions must explore all latent directions). Persistent excitation is the classical condition in subspace identification \citep{ahuja2023, buchholz2024}, and extending LeJEPA-style identifiability to the controlled setting is the subject of ongoing work.

\subsection{Connection to classical optimal control}\label{app:planning-classical}

The covariance of LQR under orthogonal reparametrization has been exploited for decades in control theory: the Kalman filter, the Riccati equation, and Lyapunov stability analysis all transform covariantly under orthogonal changes of basis, and this is why principal-component and whitening transforms are used as preprocessing steps in classical state estimation. What is new here is not the covariance itself but its role as a \emph{consequence of an identifiability guarantee}. In classical control, the state is assumed given; the engineer chooses a coordinate system. In the LeJEPA setting, the coordinate system is chosen by the learning procedure, and the theorem is that the choice is made consistently enough, orthogonal rather than arbitrary linear (let alone arbitrary nonlinear), that classical control tools continue to apply without modification. Latent-space planning with a learned world model has a long heritage \citep{brysonho1975, nguyen1990neural, ha2018world}; Thm.~\ref{thm:planning} provides a representation-learning foundation for the classical pipeline.

\section{Alternative Proof via Dirichlet Energy}\label{app:dirichlet}

We provide a second, independent proof of identifiability using a different noise model and different mathematical machinery. This proof requires stronger regularity ($C^1$ diffeomorphism and infinitesimal noise limit) but offers complementary geometric insight and makes explicit contact with the local isometry approach used in prior nonlinear ICA  \citep{buchholz2025robustness}.

\subsection{Setup}

Consider the alternative noise model $z' = z + \sqrt{\epsilon}\,\eta$ with $\eta \sim \mathcal{N}(0, I_N)$, define the normalized loss:
\begin{equation}
    \mathcal{L}(h) = \lim_{\epsilon \to 0} \frac{1}{\epsilon}\, \mathbb{E}\!\left[\|h(z + \sqrt{\epsilon}\eta) - h(z)\|^2\right].
\end{equation}
We require $h: \mathbb{R}^N \to \mathbb{R}^N$ to be a $C^1$ diffeomorphism with $h(z) \sim \mathcal{N}(0, I_N)$ when $z \sim \mathcal{N}(0, I_N)$.

\subsection{Proof}

\begin{theorembox}[Identifiability via Dirichlet energy]\label{thm:dirichlet}
Any $C^1$ diffeomorphism $h$ that preserves the standard Gaussian measure and minimizes $\mathcal{L}(h)$ must satisfy $h(z) = Qz$ for some orthogonal matrix $Q \in O(N)$.
\end{theorembox}

\paragraph{Step 1: Reduction to Dirichlet energy.}
By the first-order Taylor expansion $h(z + \sqrt{\epsilon}\eta) - h(z) = \sqrt{\epsilon}\, J_h(z)\,\eta + o(\sqrt{\epsilon})$, where $J_h(z)$ is the Jacobian of $h$ at $z$. Squaring, dividing by $\epsilon$, taking expectations, and using $\mathbb{E}_\eta[\|A\eta\|^2] = \|A\|_F^2$ for any matrix $A$:
\begin{equation}
    \mathcal{L}(h) = \mathbb{E}_{z \sim \gamma_N}\!\left[\|J_h(z)\|_F^2\right],
\end{equation}
which is the Dirichlet energy of $h$ with respect to the Gaussian measure $\gamma_N$.

The exchange of limit and expectation is justified by dominated convergence: the $C^1$ regularity of $h$ combined with the Gaussian decay of $\gamma_N$ provides the necessary integrability. The full computation, including careful treatment of the remainder term, is given below.

Starting from the Taylor expansion:
\begin{equation}
    h(z + \sqrt{\epsilon}\eta) - h(z) = \sqrt{\epsilon}\, J_h(z)\,\eta + R(z, \sqrt{\epsilon}\eta),
\end{equation}
where $\|R(z,\sqrt{\epsilon}\eta)\| = o(\sqrt{\epsilon})$. Expanding the squared norm:
\begin{equation}
    \|h(z{+}\sqrt{\epsilon}\eta) - h(z)\|^2 = \epsilon\,\|J_h(z)\eta\|^2 + 2\sqrt{\epsilon}\,(J_h(z)\eta)^\top R + \|R\|^2.
\end{equation}
Dividing by $\epsilon$ and taking $\epsilon \to 0$, the last two terms vanish, yielding:
\begin{equation}
    \frac{1}{\epsilon}\,\mathbb{E}_{z,\eta}\!\left[\|h(z{+}\sqrt{\epsilon}\eta) - h(z)\|^2\right] \;\xrightarrow{\epsilon \to 0}\; \mathbb{E}_{z,\eta}\!\left[\|J_h(z)\eta\|^2\right].
\end{equation}
Using $\mathbb{E}_\eta[\eta^\top M \eta] = \mathrm{tr}(M)$ for $M = J_h(z)^\top J_h(z)$:
\begin{equation}
    \mathbb{E}_\eta\!\left[\|J_h(z)\eta\|^2\right] = \mathrm{tr}(J_h(z)^\top J_h(z)) = \|J_h(z)\|_F^2.
\end{equation}

\paragraph{Step 2: The expected log-determinant is zero.}
By the change of variables formula for densities, $p(h(z))\,|\det J_h(z)| = p(z)$ for all $z$. Substituting the Gaussian density $p(z) = (2\pi)^{-N/2}\exp(-\|z\|^2/2)$ and noting that the normalizing constants cancel:
\begin{equation}
    \ln|\det J_h(z)| = \tfrac{1}{2}\|h(z)\|^2 - \tfrac{1}{2}\|z\|^2.
\end{equation}
Taking expectations and using $\mathbb{E}[\|h(z)\|^2] = \mathbb{E}[\|z\|^2] = N$ (both follow from $h(z) \sim \mathcal{N}(0,I_N)$):
\begin{equation}
    \mathbb{E}\!\left[\ln|\det J_h(z)|\right] = 0.
\end{equation}

\paragraph{Step 3: Lower bound via AM-GM and Jensen.}
Let $\sigma_1(z), \ldots, \sigma_N(z)$ be the singular values of $J_h(z)$. Then $\|J_h(z)\|_F^2 = \sum_i \sigma_i(z)^2$ and $|\det J_h(z)| = \prod_i \sigma_i(z)$. By AM-GM applied to $\sigma_1^2, \ldots, \sigma_N^2$:
\begin{equation}
    \frac{1}{N}\sum_{i=1}^N \sigma_i(z)^2 \geq \left(\prod_{i=1}^N \sigma_i(z)^2\right)^{1/N} = |\det J_h(z)|^{2/N},
\end{equation}
so $\|J_h(z)\|_F^2 \geq N\,|\det J_h(z)|^{2/N}$ pointwise. Taking expectations:
\begin{equation}
    \mathcal{L}(h) \geq N\,\mathbb{E}\!\left[|\det J_h(z)|^{2/N}\right].
\end{equation}
Writing $|\det J_h(z)|^{2/N} = \exp\!\left(\tfrac{2}{N}\ln|\det J_h(z)|\right)$ and applying Jensen's inequality to the strictly convex function $x \mapsto e^{(2/N)x}$:
\begin{equation}
    \mathbb{E}\!\left[|\det J_h(z)|^{2/N}\right] \geq \exp\!\left(\tfrac{2}{N}\,\mathbb{E}\!\left[\ln|\det J_h(z)|\right]\right) = \exp(0) = 1.
\end{equation}
Therefore $\mathcal{L}(h) \geq N$. Any orthogonal map $h(z) = Qz$ achieves this bound since $\|Q\|_F^2 = N$.

\paragraph{Step 4: Equality forces orthogonal Jacobian.}
At the minimum $\mathcal{L}(h) = N$, equality must hold in both Jensen and AM-GM. Strict convexity of the exponential forces $\ln|\det J_h(z)|$ to be constant a.e., and since its expectation is 0, we get $|\det J_h(z)| = 1$ a.e. Equality in AM-GM forces all squared singular values to be equal: $\sigma_i(z)^2 = \cdots = \sigma_N(z)^2$. Combined with $\prod_i \sigma_i = 1$, this gives $\sigma_i(z) = 1$ for all $i$. Thus:
\begin{equation}
    J_h(z)^\top J_h(z) = I_N \qquad \text{for a.e.\ } z.
\end{equation}
By $C^1$ continuity, this holds everywhere.

\paragraph{Step 5: Orthogonal Jacobian implies linearity (Mazur--Ulam).}
Since $J_h(z)^\top J_h(z) = I_N$ everywhere, $h$ preserves the norm of tangent vectors: $\|J_h(z) v\| = \|v\|$ for all $v$. For any $z_1, z_2 \in \mathbb{R}^N$, integrating along the straight-line path $\gamma(t) = z_1 + t(z_2 - z_1)$:
\begin{equation}
    h(z_2) - h(z_1) = \int_0^1 J_h(\gamma(t))(z_2 - z_1)\,dt.
\end{equation}
Taking norms and applying the triangle inequality:
\begin{equation}
    \|h(z_2) - h(z_1)\| \leq \int_0^1 \|J_h(\gamma(t))(z_2-z_1)\|\,dt = \int_0^1 \|z_2-z_1\|\,dt = \|z_2-z_1\|.
\end{equation}
Since $h$ is a diffeomorphism, $h^{-1}$ exists and is $C^1$ with $J_{h^{-1}}(h(z)) = J_h(z)^{-1} = J_h(z)^\top$, which is also orthogonal ($J_h J_h^\top = I_N$ for square matrices with $J_h^\top J_h = I_N$). Applying the same argument to $h^{-1}$ yields $\|z_2 - z_1\| \leq \|h(z_2) - h(z_1)\|$. Together:
\begin{equation}
    \|h(z_1) - h(z_2)\| = \|z_1 - z_2\| \qquad \text{for all } z_1, z_2 \in \mathbb{R}^N,
\end{equation}
so $h$ is a surjective isometry of $\mathbb{R}^N$. By the Mazur--Ulam theorem \citep{mazur1932}, every surjective isometry of a real normed space is affine: $h(z) = Qz + b$ with $Q$ orthogonal and $b \in \mathbb{R}^N$. Measure preservation forces $0 = \mathbb{E}[h(z)] = Q\mathbb{E}[z] + b = b$, giving $h(z) = Qz$. \qed

\paragraph{Generalization.}
No step in the proof above uses the Gaussian assumption on~$p$;
the noise~$\eta$ need only satisfy $\mathbb{E}[\eta\eta^\top] = I_N$
(Step~1), and Steps~2--5 hold for any density preserved by~$h$.
The identity map always achieves $\mathcal{L} = N$, so the
lower bound is tight regardless of~$p$.
The conclusion is that any $C^1$ measure-preserving
diffeomorphism minimizing~$\mathcal{L}$ must be an isometry
$h(z) = Qz + b$ with $p(Qz + b) = p(z)$.

\subsection{Comparison of the Two Proofs}

The Hermite proof (Thm.~\ref{thm:main}) and the Dirichlet energy proof (Thm.~\ref{thm:dirichlet}) establish the same conclusion through different mechanisms:

\begin{center}
\renewcommand{\arraystretch}{1.3}
\begin{tabular}{lll}
\hline
& \textbf{Hermite proof} & \textbf{Dirichlet energy proof} \\
\hline
Regularity & Measurable & $C^1$ diffeomorphism \\
Noise level & Any $\rho \in (0,1)$ & Infinitesimal ($\rho \to 1$) \\
Key tool & Mehler's formula & AM-GM + Jensen + Mazur--Ulam \\
Mechanism & Low-pass filter on degrees & Jacobian rigidity \\
Connection to & Harmonic analysis & Differential geometry \\
\hline
\end{tabular}
\end{center}

The Hermite proof is strictly more general (weaker assumptions, finite noise), but the Dirichlet energy proof provides geometric intuition that complements the spectral perspective: the Jacobian must be an isometry at every point, so the map cannot compress or stretch any direction anywhere.

\section{Prior Work: Connection to Slow Feature Analysis}
\label{app:prior-work}

Our identifiability result has a deep connection to the theory of Slow Feature Analysis (SFA) developed by \citet{wiskott2002} and extended to nonlinear blind source separation by \citet{sprekeler2014}. \citet{sobal2022slow} analyzed the tendency of JEPAs to focus on slow features. We spell out this connection in detail, both because it clarifies the scope of our contribution and because the SFA perspective suggests natural generalizations.

\paragraph{SFA and its optimization problem.}
SFA seeks scalar output functions $g_j(x)$ whose output signals $y_j(t) = g_j(x(t))$ vary as slowly as possible in time, as measured by the $\Delta$-value $\Delta(y_j) = \langle \dot{y}_j^2 \rangle_t$, subject to the constraints $\langle y_j \rangle = 0$ (zero mean), $\langle y_j^2 \rangle = 1$ (unit variance), and $\langle y_i y_j \rangle = 0$ for $i < j$ (decorrelation) \citep{wiskott2002}.
For discrete-time data, the $\Delta$-value is proportional to $\langle y^2 \rangle - \langle y(t) y(t+1) \rangle$, so that for unit-variance outputs, minimizing $\Delta$ is equivalent to maximizing the temporal correlation $\langle y(t) y(t+1) \rangle$.
The SFA constraints are therefore precisely whitening, and the SFA objective is precisely the alignment loss used in our setting.

\paragraph{Nonlinear identifiability via SFA.}
\citet{sprekeler2014} analyzed SFA in the setting where the input data $x(t) = F(s(t))$ are generated from statistically independent latent variables $s(t)$ via an unknown invertible nonlinear mixing function $F$.
They showed that if the function space $\mathcal{F}$ accessible to SFA is unrestricted (a Sobolev space), then the set of achievable output signals is independent of $F$: for every $g \in \mathcal{F}$, the composition $g \circ F$ is also in $\mathcal{F}$, so the mixing function drops out of the analysis entirely.
This is the same role played by our assumption that the encoder is expressive enough to invert the mixing function.

\paragraph{Eigenfunction structure and Sturm--Liouville theory.}
Building on \citet{franzius2007}, \citet{sprekeler2014} showed that the optimal SFA functions satisfy an eigenvalue equation $\mathcal{D} g = \lambda g$ for a second-order partial differential operator $\mathcal{D}$.
When the latent variables are statistically independent, the operator separates as $\mathcal{D} = \sum_\alpha \mathcal{D}_\alpha$, where each $\mathcal{D}_\alpha$ depends on a single latent variable $s_\alpha$ only.
By separation of variables, the eigenfunctions of the full operator are products $g_{\mathbf{i}}(s) = \prod_\alpha g_{\alpha i_\alpha}(s_\alpha)$ of the eigenfunctions of the individual operators, with eigenvalues $\lambda_{\mathbf{i}} = \sum_\alpha \lambda_{\alpha i_\alpha}$.
Each one-dimensional eigenvalue problem is of Sturm--Liouville type, and standard results guarantee that the eigenfunctions $g_{\alpha i}$ are oscillatory with exactly $i$ zeros.
In particular, the first non-constant eigenfunction $g_{\alpha 1}$ is monotonic and therefore invertible, yielding identifiability of each latent variable up to a monotonic transformation for \emph{any} latent variable distribution.

\paragraph{The Gaussian specialization and Hermite polynomials.}
When the latent variables are reversible Gaussian processes (equivalently, Ornstein--Uhlenbeck processes), the conditional variance of the derivative given the current value is constant ($K_\alpha(s_\alpha) = K_\alpha$), and the latent variable density is Gaussian.
The Sturm--Liouville equation then reduces to Hermite's differential equation, with eigenfunctions given by the Hermite polynomials $H_i$ and eigenvalues $\lambda_{\alpha i} = i K_\alpha$.
The first non-constant eigenfunction is simply $g_{\alpha 1}(s_\alpha) = s_\alpha$: the identity.
This is the same Hermite eigenstructure that appears in our proof via Mehler's formula.
Indeed, Mehler's formula is the spectral expansion of the Green's function (transition kernel) of the operator $\mathcal{D}_\alpha$ in the Gaussian case, so the two proof routes, Sturm--Liouville eigenvalue analysis and direct Mehler expansion, access the same underlying mathematical structure from different entry points.

\paragraph{Sequential versus simultaneous extraction.}
A key difference between the approach of \citet{sprekeler2014} and ours lies in how the $d$ output dimensions are determined.
Their xSFA algorithm extracts latent variables \emph{sequentially}: it identifies the slowest latent variable, projects out all its nonlinear transformations, and repeats.
This greedy procedure, combined with the assumption that the latent variables have distinct autocorrelation rates ($K_\alpha$ all different), yields identifiability up to \emph{permutation}, a stronger result than our orthogonal identifiability.
However, the sequential approach is fragile in practice: it requires representing the nonlinear transformations of already-extracted latent variables (approximated by high-degree polynomial expansions), accumulates estimation errors across iterations, and degrades rapidly as the number of latent variables increases \citep[Section~5]{sprekeler2014}.

Our approach instead optimizes all $d$ output dimensions \emph{simultaneously} via a joint loss (alignment plus whitening or SIGReg).
Because whitening is invariant under orthogonal transformations, the simultaneous approach cannot resolve individual latent variable axes, yielding identifiability up to orthogonal rotation rather than permutation.
This is the natural and complete identifiability class for the simultaneous setting.

\paragraph{The role of isotropic transitions.}
The distinction between sequential and simultaneous extraction is intimately connected to the structure of the latent transitions.
When the autocorrelation parameters $\rho_\alpha$ differ across dimensions, the eigenvalue $\lambda_{\alpha i} = i K_\alpha$ of the $i$-th Hermite component of latent variable $\alpha$ depends on $\alpha$.
For the simultaneous approach to correctly identify the subspace of first harmonics (degree-1 Hermite components), all first harmonics must have smaller eigenvalues than all higher-order terms.
This requires $\max_\alpha K_\alpha < 2 \min_\beta K_\beta$, or equivalently, the fastest latent variable must not be more than twice as fast as the slowest.
When $\rho_\alpha$ are all equal (isotropic transitions), this condition is trivially satisfied, and the first $d$ eigenfunctions are exactly the $d$ first harmonics, yielding clean orthogonal identifiability.

We verified empirically that anisotropic transitions break the simultaneous approach: with distinct $\rho_\alpha$, the encoder recovers the second Hermite polynomial of the slow latent variable in place of the first Hermite polynomial of the fast latent variable, exactly as the eigenvalue interleaving predicts. See also the anisotropy in Tab.~\ref{tab:reacher}.
This confirms that isotropic transitions are not merely a simplifying assumption but a necessary condition for simultaneous (parallel) identifiability.
The anisotropic case requires either sequential extraction \`a la \citet{sprekeler2014} or a modified objective such as the AnInfoNCE loss of \citet{rusak2024}, which reweights the contrastive objective to account for per-dimension variance differences, at the cost of downstream accuracy.

\paragraph{Beyond Gaussian latents.}
The Sturm--Liouville perspective of \citet{sprekeler2014} suggests a generalization of our result that does not require Gaussian latents.
If all latent variables are drawn i.i.d.\ from the \emph{same} (arbitrary) distribution with the same transition structure, then the eigenvalue spectra of the individual operators $\mathcal{D}_\alpha$ are identical: $\lambda_{\alpha i} = \lambda_i$ for all $\alpha$.
All first harmonics share the same eigenvalue $\lambda_1$, all second harmonics share $\lambda_2 > \lambda_1$ (the gap is guaranteed by Sturm--Liouville theory), and the top-$d$ eigenspace of the simultaneous problem is exactly the span of the first harmonics.
The simultaneous approach then yields identifiability up to orthogonal rotation, not in the original latent variable space but in the space of first harmonics.
Since the first harmonics are monotonic functions of the latent variables, this amounts to identifiability up to an orthogonal rotation composed with a shared monotonic nonlinearity applied elementwise.
Only in the Gaussian case does the first harmonic equal the identity, collapsing this to linear orthogonal identifiability.
A rigorous treatment of this non-Gaussian generalization is an interesting direction for future work.

\paragraph{Relation to Mehler's formula and maximal correlation.}
The Hermite expansion and temporal correlation bounds used in our proof are closely related to classical results in probability theory.
Specifically, the correlation bound (equation~\eqref{eq:corr-decomp}) is a consequence of the classical Hirschfeld--Gebelein--R\'enyi maximal correlation for jointly Gaussian variables \citep{gebelein1941, renyi1959, lancaster1958}.
Our contribution is not the bound itself but its application to identifiability via the measure-preservation constraint.
As discussed above, Mehler's formula, the key technical tool in our proof, is equivalent to the spectral expansion of the transition kernel associated with the Sturm--Liouville (SL) operator of \citet{sprekeler2014} in the Gaussian case.
The SL route is more general (handles arbitrary latent variable distributions) but does not directly yield the quantitative bound (Thm.~\ref{thm:approx}) that Mehler's formula provides.

\paragraph{Practical differences: learned features versus fixed kernels.}
While the theoretical optimization problems coincide, the practical implementations differ substantially.
\citet{sprekeler2014} use fixed polynomial kernel expansions (degree 7 for two latent variables) followed by a linear eigenvalue problem, a two-stage procedure in which the feature space is chosen a priori and never adapted.
In contrast, the LeJEPA/SIGReg pipeline learns the encoder end-to-end via gradient descent on a neural network, jointly optimizing the feature representation and the alignment plus regularization objectives.
This learned-feature approach avoids the numerical instabilities of high-degree polynomial expansions noted by \citet{sprekeler2014}, scales to high-dimensional data, and is the standard paradigm in modern self-supervised learning.
Our identifiability result thus provides theoretical grounding for an architecture and training procedure that is already in widespread use. 

\paragraph{Connections to diffusion maps and spectral geometry.}
\citet{sprekeler2011} showed that SFA is closely related to Laplacian eigenmaps and diffusion maps, with the operator $\mathcal{D}$ playing the role of a Laplace--Beltrami operator weighted by the temporal structure of the data.
The eigenfunctions of $\mathcal{D}$ are the same objects studied in spectral geometry and manifold learning (see also \citep{balestriero2022contrastivenoncontrastiveselfsupervisedlearning}).
\citet{singer2008} exploited a similar connection, using data-driven diffusion maps with a carefully chosen local metric to achieve nonlinear latent variable separation.
This spectral-geometric perspective suggests that the identifiability phenomenon we study, encoders converging to specific linear transformations of the latents, may be an instance of a broader principle: that temporal or geometric structure on the data manifold, combined with appropriate regularization, constrains learned representations to align with the intrinsic coordinates of the generative process.

\paragraph{Summary of the relationship.}
Table~\ref{tab:comparison-sfa} summarizes the key similarities and differences.
The mathematical core, Hermite eigenfunctions under Gaussian measure with temporal correlations yielding identifiability, is shared.
Our contributions are (a)~the simultaneous formulation with its clean orthogonal identifiability class, (b)~the converse result identifying the Gaussian as the \emph{unique} latent distribution yielding linear identifiability (Thm.~\ref{thm:converse}), (c)~the quantitative approximate identifiability bound under whitening alone (Thm.~\ref{thm:approx}), (d)~the observation that isotropic transitions are necessary for the simultaneous approach, and (e)~the planning equivalence (Thm.~\ref{thm:planning}) connecting linear identifiability to optimal latent-space planning.

\begin{table}[t]
\centering
\small
\begin{tabular}{lcc}
\toprule
& \textbf{Sprekeler et al.\ (2014)} & \textbf{This work} \\
\midrule
Identifiability class & Permutation & Orthogonal \\
Latent variable distributions & Any independent & Gaussian (or i.i.d.\ factorial) \\
Transition structure & Distinct rates required & Isotropic required \\
Extraction & Sequential (greedy) & Simultaneous \\
Function space & Fixed polynomial kernel & Learned (neural network) \\
Proof technique & Sturm--Liouville / PDE & Mehler's formula / Parseval \\
Approximate bound & None & $D + (\varepsilon + D)^2$ \\
Practical algorithm & xSFA (fragile, $\leq 6$ latent variables) & LeJEPA / SIGReg (scalable) \\
\bottomrule
\end{tabular}
\vspace{5pt}
\caption{Comparison of the SFA identifiability theory of \citet{sprekeler2014} and the present work.}
\label{tab:comparison-sfa}
\end{table}

\section{Lean Verification}\label{app:lean}

We formally verify all five theoretical results, namely the Hermite proof of Thm.~\ref{thm:main}, the Gaussian uniqueness result (Thm.~\ref{thm:converse}), the Dirichlet energy proof (Appendix~\ref{app:dirichlet}), the approximate identifiability bound (Thm.~\ref{thm:approx}), and the planning equivalence theorem (Thm.~\ref{thm:planning}), in the Lean~4 theorem prover using the Mathlib mathematical library. The formalization compiles with zero \texttt{sorry} obligations: every logical step from axiomatized premises to stated conclusions is machine-checked.

\paragraph{Scope and methodology.}
Lean~4 verification requires every inference step to be justified by a previously established lemma or axiom. When a standard mathematical result exists in Mathlib (e.g., \texttt{Finset.sum\_lt\_sum} for strict finite-sum monotonicity), we invoke it directly. When a result is standard but not yet available in Mathlib, either because the mathematical objects have not been formalized (Hermite polynomials) or because connecting available results to our specific statement forms would require substantial software engineering (Mazur--Ulam, AM-GM with uniform weights), we introduce it as a Lean \texttt{axiom} with a reference to its mathematical source. The complete reasoning chains \emph{between} these axioms are fully verified.
Table~\ref{tab:lean-summary} gives the complete inventory.
The formalization is organized into five files corresponding to the five results.

\begin{table}[h]
\centering
\small
\renewcommand{\arraystretch}{1.15}
\begin{tabular}{llc}
\toprule
\textbf{File / Component} & \textbf{Lean name} & \textbf{Status} \\
\midrule
\multicolumn{3}{l}{\textit{Hermite.lean: Thm.~\ref{thm:main} via Hermite polynomials}} \\
Hermite basis completeness & \texttt{mehler\_summability} & axiomatized \\
Contraction lemma ($\rho^d$ decay) & (folded into above) & axiomatized \\
Mehler's formula & \texttt{correlation\_eq\_spectral\_sum} & axiomatized \\
$\rho^d \leq \rho$ for $d \geq 1$ & \texttt{pow\_le\_self\_of\_pos\_lt\_one} & \textsc{verified} \\
$\rho^d < \rho$ for $d \geq 2$ (strict) & \texttt{pow\_lt\_self\_of\_ge\_two} & \textsc{verified} \\
Pointwise bound $w_d \rho^d \leq w_d \rho$ & \texttt{spectral\_term\_le} & \textsc{verified} \\
Correlation bound $\leq \rho$~\eqref{eq:corr-decomp} & \texttt{correlation\_le\_rho} & \textsc{verified} \\
Equality $\Leftrightarrow$ $w_1 = 1$ & \texttt{equality\_forces\_degree\_one} & \textsc{verified} \\
Degree-1 $\Rightarrow$ linearity & \texttt{linear\_of\_degree\_one} & axiomatized \\
Gaussianity + linear $\Rightarrow$ orthogonal & \texttt{orthogonal\_of\_gaussian\_linear} & axiomatized \\
Loss lower bound $\mathcal{L} \geq 2(1-\rho)N$ & \texttt{loss\_lower\_bound} & \textsc{verified} \\
Theorem assembly $h = Qz$ & \texttt{hermite\_identifiability} & \textsc{verified} \\
\midrule
\multicolumn{3}{l}{\textit{Uniqueness.lean: Thm.~\ref{thm:converse} (Gaussian uniqueness)}} \\
Affine eigenfunction $\Rightarrow$ affine score & \texttt{score\_affine\_of\_eigenfunction} & \textsc{verified} \\
Affine score $\Rightarrow$ Gaussian density & \texttt{gaussian\_of\_affine\_score} & axiomatized \\
Only-if direction assembly & \texttt{gaussian\_of\_affine\_eigenfunction} & \textsc{verified} \\
Gaussian $\Rightarrow$ Hermite eigenfunctions & \texttt{hermite\_first\_eigenfunction\_of\_gaussian} & axiomatized \\
Full biconditional & \texttt{gaussian\_uniqueness} & \textsc{verified} \\
Zero-mean specialization & \texttt{score\_pure\_linear\_zero\_mean} & \textsc{verified} \\
\midrule
\multicolumn{3}{l}{\textit{Dirichlet.lean: Appendix~\ref{app:dirichlet} via Dirichlet energy}} \\
AM-GM inequality & \texttt{amgm\_sum\_ge\_prod\_pow} & axiomatized \\
Jensen's inequality (exp) & \texttt{exp\_mean\_ge\_mean\_exp} & axiomatized \\
Mazur--Ulam theorem & \texttt{mazur\_ulam} & axiomatized \\
Norm-preserving CLM $\Rightarrow$ isometry & \texttt{clm\_isometry\_of\_norm\_preserving} & \textsc{verified} \\
Orthogonal Jacobian $\Rightarrow$ 1-Lipschitz & \texttt{lipschitz\_of\_orthogonal\_jacobian} & \textsc{verified} \\
Bilipschitz $\Rightarrow$ global isometry & \texttt{isometry\_of\_bilipschitz} & \textsc{verified} \\
$h(0) = 0 \Rightarrow$ linear isometry & (part of assembly) & \textsc{verified} \\
Theorem assembly $h = Qz$ & \texttt{dirichlet\_identifiability} & \textsc{verified} \\
\midrule
\multicolumn{3}{l}{\textit{Approx.lean: Thm.~\ref{thm:approx} (approximate bound)}} \\
Spectral gap positivity & \texttt{spectral\_gap\_pos} & \textsc{verified} \\
$W_{\mathrm{nl}} \leq D$ from alignment gap & \texttt{nonlinear\_energy\_le\_D} & \textsc{verified} \\
Polar decomposition bound & \texttt{polar\_bound\_axiom} & axiomatized \\
Pythagorean decomposition & \texttt{pythagorean\_axiom} & axiomatized \\
$\|M - Q\|_F^2$ bound & \texttt{linear\_deviation\_sq\_bound} & \textsc{verified} \\
Monotonicity of $({\varepsilon} + t)^2 + t$ & \texttt{bound\_monotone} & \textsc{verified} \\
Full bound $D + (\varepsilon + D)^2$ & \texttt{approximate\_identifiability} & \textsc{verified} \\
Exact recovery ($\delta = \varepsilon = 0$) & \texttt{exact\_recovery\_special\_case} & \textsc{verified} \\
\midrule
\multicolumn{3}{l}{\textit{Planning.lean: Thm.~\ref{thm:planning} (planning equivalence)}} \\
Orthogonal invariance definition        & \texttt{IsOrthogonalInvariant}   & \textsc{verified} \\
Stage pushforward                        & \texttt{stage\_pushforward}      & axiomatized \\
Terminal pushforward                     & \texttt{terminal\_pushforward}   & axiomatized \\
Per-step stage cost equivalence          & \texttt{stage\_cost\_equiv}      & \textsc{verified} \\
Terminal cost equivalence                & \texttt{terminal\_cost\_equiv}   & \textsc{verified} \\
Total cost equivalence (main step)       & \texttt{planning\_equivalence}   & \textsc{verified} \\
Minimizer equivalence                    & \texttt{minimizer\_equivalence}  & \textsc{verified} \\
Value equivalence                        & \texttt{value\_equivalence}      & \textsc{verified} \\
\bottomrule
\end{tabular}
\vspace{1pt}
\caption{Lean~4 verification status for each proof component. \textsc{verified}: machine-checked with zero \texttt{sorry}. \textsc{axiomatized}: stated as a Lean \texttt{axiom}; standard result not yet in Mathlib or requiring nontrivial API software engineering.}
\label{tab:lean-summary}
\end{table}

\subsection{What Is Axiomatized and Why}

The axiomatized results fall into five categories:

\paragraph{1.\ Hermite polynomial infrastructure.}
Mathlib does not yet define Hermite polynomials or their properties under the Gaussian measure. The contraction property~\eqref{eq:contraction-result}, Mehler's formula~\eqref{eq:mehler-result}, and the correspondence between degree-1 concentration and linearity are axiomatized for this reason. These are classical results in probability theory \citep{mehler1866, lancaster1958}; their eventual formalization in Mathlib would allow these axioms to be discharged.

\paragraph{2.\ Sturm--Liouville spectral theory.}
The Gaussian uniqueness proof (Thm.~\ref{thm:converse}) axiomatizes two results: that an affine score function uniquely determines a Gaussian density (standard exponential family characterization), and that a Gaussian density yields Hermite eigenfunctions with the identity as the first non-constant eigenfunction. The core algebraic step between these, namely extracting the affine score from the Sturm--Liouville eigenfunction equation via field arithmetic (including the sign argument that the score slope $-\lambda_1/K$ is strictly negative), is fully verified.

\paragraph{3.\ Standard analysis results requiring API software engineering.}
The AM-GM inequality with uniform weights, Jensen's inequality for $\exp$, and the Mazur--Ulam theorem are all available in Mathlib in closely related forms, but connecting them to our specific statement signatures requires nontrivial type-class and measure-theory software engineering. We axiomatize the exact statements we need and reference their Mathlib counterparts in the source code.

\paragraph{4.\ Measure-theoretic and matrix-analytic facts.}
The polar decomposition bound ($\|M - Q\|_F \leq \varepsilon + W_{\mathrm{nl}}$) and the Pythagorean decomposition ($\mathbb{E}[\|h(z) - Qz\|^2] = \|M - Q\|_F^2 + W_{\mathrm{nl}}$) combine matrix analysis with integration under the Gaussian measure. These are stated as axioms with their mathematical content documented in the source.

\paragraph{5.\ Trajectory pushforward for expected costs.}
The planning equivalence theorem (Thm.~\ref{thm:planning}) axiomatizes the pushforward relation between the learned-latent and true-latent expected per-step costs: for any test function $c$, the expected value of $c(\hat z_t, a_t)$ under the pushforward dynamics starting from $Q z_0$ equals the expected value of $c(Q z_t, a_t)$ under the original dynamics starting from $z_0$. This is the measure-theoretic content of ``the joint law of $(\hat z_0, \ldots, \hat z_T)$ under the pushforward equals the joint law of $(Q z_0, \ldots, Q z_T)$ under the original''; stating it at the level of expected values rather than joint distributions sidesteps the need to formalize stochastic dynamics in Lean while preserving the algebraic content of the theorem's proof.

\paragraph{What is verified.}
All reasoning chains \emph{between} the axiomatized premises are fully machine-checked: the correlation bound and its strict equality characterization (the mathematical core of the Hermite proof), the algebraic chain from the SL eigenfunction equation to the affine score function and the biconditional assembly (the Gaussian uniqueness proof), the geometric chain from orthogonal Jacobian to linear isometry (the core of the Dirichlet proof), the spectral gap argument controlling nonlinear energy, the monotonicity and algebraic assembly of the approximate bound, and the recovery of the exact theorem as a special case. The inverse function theorem is incorporated as a structural assumption on the diffeomorphism rather than invoked as a separate lemma.

\subsection{Build Information}
The project compiles against Lean~4 v4.28.0 with Mathlib v4.28.0 (8{,}032 build targets, zero errors, zero \texttt{sorry} obligations) and is available at \url{https://github.com/klindtlab/lejepa-identifiability}.

\section{Experimental Details and Additional Results}\label{app:experiment}

\subsection{Nonlinear Mixing Functions}
The four mixing functions $g: \mathbb{R}^2 \to \mathbb{R}^2$ used in Figs.~\ref{fig:teaser},\ref{fig:demo} are:
\begin{enumerate}
    \item \textbf{Spiral:} $g(z) = R(\pi\|z\|_2)\, z$, where $R(\theta) = \bigl(\begin{smallmatrix} \cos\theta & -\sin\theta \\ \sin\theta & \cos\theta \end{smallmatrix}\bigr)$. This norm-dependent rotation is a measure-preserving diffeomorphism of $\mathcal{N}(0, I_2)$: at each point, $g$ acts as an orthogonal rotation ($|\det J_g| = 1$), so it preserves the Gaussian density despite being highly nonlinear.
    \item \textbf{Sinusoidal shear:} $g(z_1, z_2) = (z_1 + \sin(1.5\,z_2),\; z_2)$.
    \item \textbf{Parabolic shear:} $g(z_1, z_2) = (z_1,\; z_2 + z_1^2)$.
    \item \textbf{RealNVP coupling layer:} a single affine coupling block \citep{dinh2016density} of the form $g(z_1, z_2) = (z_1,\; z_2 \odot \exp(s(z_1)) + t(z_1))$, where $s, t$ are small MLPs with random initialization. This is the same family used as the matched encoder in the scaling experiment (App.~\ref{app:scaling}); here we use it in 2D as an additional nonlinear diffeomorphism.
\end{enumerate}

The first (spiral) map applies a rotation whose angle $\pi\|z\|_2$ depends on the distance from the origin. Points at different radii are rotated by different angles, producing a characteristic spiral structure in observation space (Figure~\ref{fig:demo}, center). The map is a measure-preserving diffeomorphism of $\mathcal{N}(0, I_2)$: at each point, $g$ acts as an orthogonal rotation ($|\det J_g| = 1$ everywhere), so it preserves the Gaussian density. This is an interesting case because it entails SIGReg is already fulfilled in input space. However, $g$ is highly nonlinear; it entangles the radial and angular components of the latent space.

\subsection{Network Architecture and Training}

The encoder $f$ is a 4-layer MLP with hidden dimension 256 and GELU activations. The architecture is:
\begin{equation}
    \mathbb{R}^2 \xrightarrow{\text{Linear}(2,256)} \xrightarrow{\text{GELU}} \xrightarrow{\text{Linear}(256,256)} \xrightarrow{\text{GELU}} \xrightarrow{\text{Linear}(256,256)} \xrightarrow{\text{GELU}} \xrightarrow{\text{Linear}(256,2)} \mathbb{R}^2.
\end{equation}

\subsection{Training Procedure}

The loss function combines two terms with a balancing coefficient $\lambda \in [0, 1]$:
\begin{equation}\label{eq:loss-combined}
    \mathcal{L} = \lambda\,\mathcal{L}_{\mathrm{SIG}} + (1 - \lambda)\,\mathcal{L}_{\mathrm{inv}}.
\end{equation}
\begin{enumerate}
    \item \textbf{Invariance loss:} $\mathcal{L}_{\text{inv}} = \frac{1}{B}\sum_{i=1}^B \|f(x_i) - f(x_i')\|^2$, where $(x_i, x_i')$ are positive pairs generated by sampling $z_i \sim \mathcal{N}(0, I_2)$, generating $z_i' = \rho z_i + \sqrt{1-\rho^2}\eta_i$ via the Ornstein--Uhlenbeck channel, and mixing: $x_i = g(z_i)$, $x_i' = g(z_i')$.
    \item \textbf{SIGReg regularization:} The SIGReg regularizer of \citet{balestriero2025lejepa} encourages the empirical distribution of embeddings to match the standard Gaussian by penalizing deviations of the empirical sliced characteristic function from the Gaussian target.
\end{enumerate}

\subsection{Hyperparameters}

All experiments share the same training infrastructure and learning rate schedule: constant learning rate for the first half of training, followed by cosine decay to zero. Data is generated online (fresh samples each step, infinite data regime). 

\begin{center}
\renewcommand{\arraystretch}{1.2}
\begin{tabular}{lll}
\hline
\textbf{Hyperparameter} & \textbf{2D / Gennorm / Grid} & \textbf{Scaling} \\
\hline
Optimizer & AdamW & AdamW \\
Learning rate & $3 \times 10^{-3}$ & $3 \times 10^{-3}$ \\
LR schedule & warmup (10k) + cosine (10k) & warmup (10k) + cosine (10k) \\
Weight decay & 0 & 0 \\
Batch size & 256 & 256 \\
Training steps & 20{,}000 & 20{,}000 \\
Data regime & online (infinite) & online (infinite) \\
Eval points & 10{,}000 (fixed) & 10{,}000 (fixed) \\
Views per sample & 2 & 2 \\
$\rho$ & 0.95 (2D/Gennorm), swept (Grid) & 0.95 \\
$\lambda$ & $10^{-3}$ (2D/Gennorm), swept (Grid) & $10^{-6}$ \\
Encoder & MLP (spiral/banana/sinusoid) & matched inverse-NVP \citep{dinh2016density} \\
 & matched NVP (nvp mixing) & \\
Hidden dimension & 256 (MLP) & --- \\
MLP layers & 4 & --- \\
NVP coupling layers & 8 (2D/Gennorm) & 4 \\
Activation & GELU (MLP) & tanh (NVP) \\
\hline
\end{tabular}
\end{center}

\subsection{Evaluation Metrics}

After training, we evaluate the composed map $h = f \circ g$ on a fixed evaluation set of 10{,}000 points. All metrics are computed on GPU via PyTorch for efficiency at large $N$. We report:

\begin{enumerate}
    \item \textbf{Linear $R^2$ (bidirectional):} We fit linear maps $\hat{z} = Az + b$ and $\hat{h} = Bh + c$ via ordinary least squares and report $R^2(z \to h)$ and $R^2(h \to z)$. Values close to 1 indicate that $h$ is well-approximated by a linear function of $z$.

    \item \textbf{Orthogonality error:} For the fitted linear map $\hat Q$, we report $\|\hat Q^\top \hat Q - I_N\|_F / \sqrt{N}$. A value near 0 indicates that $\hat Q$ is orthogonal, consistent with Thm.~\ref{thm:main}.

    \item \textbf{Approximate bound quantities:} The covariance deviation $\varepsilon = \|\mathrm{Cov}(h(z)) - I_N\|_F$, the alignment gap $\delta = \mathcal{L}(h) - 2(1-\rho)\,\mathrm{tr}(\mathrm{Cov}(h(z)))$ (clamped to $\geq 0$), and the Orthogonal recovery error $\min_{Q \in O(N)} \mathbb{E}[\|h(z) - Qz\|^2]$ solved via SVD.
\end{enumerate}

\subsection{Grid Search}\label{app:grid}

We perform a 2D grid search over the regularization weight $\lambda$ and the OU correlation $\rho$:
\begin{itemize}
    \item $\lambda \in \{10^{-6},\; 10^{-5},\; 10^{-4},\; 10^{-3},\; 5 \times 10^{-3},\; 10^{-2},\; 5 \times 10^{-2},\; 10^{-1},\; 5 \times 10^{-1}\}$
    \item $\rho \in \{0.3,\; 0.5,\; 0.7,\; 0.8,\; 0.9,\; 0.95,\; 0.99\}$
\end{itemize}
For each of the $9 \times 7 = 63$ settings, we train 3 independent seeds, for a total of 189 runs.
The extended $\lambda$ range (down to $10^{-6}$) covers the regime where SIGReg has negligible weight, showing that Gaussianity regularization below $\lambda = 10^{-4}$ is insufficient for identifiability regardless of $\rho$.

\begin{figure}[t]
    \centering
    \includegraphics[width=.7\textwidth]{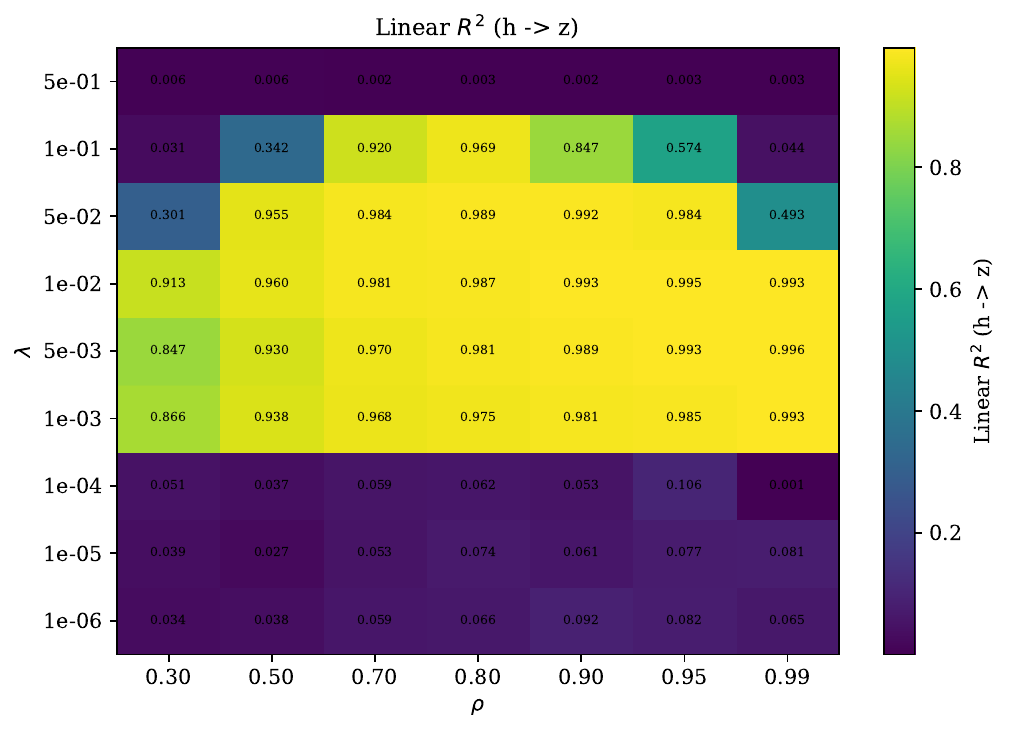}
    \includegraphics[width=.7\textwidth]{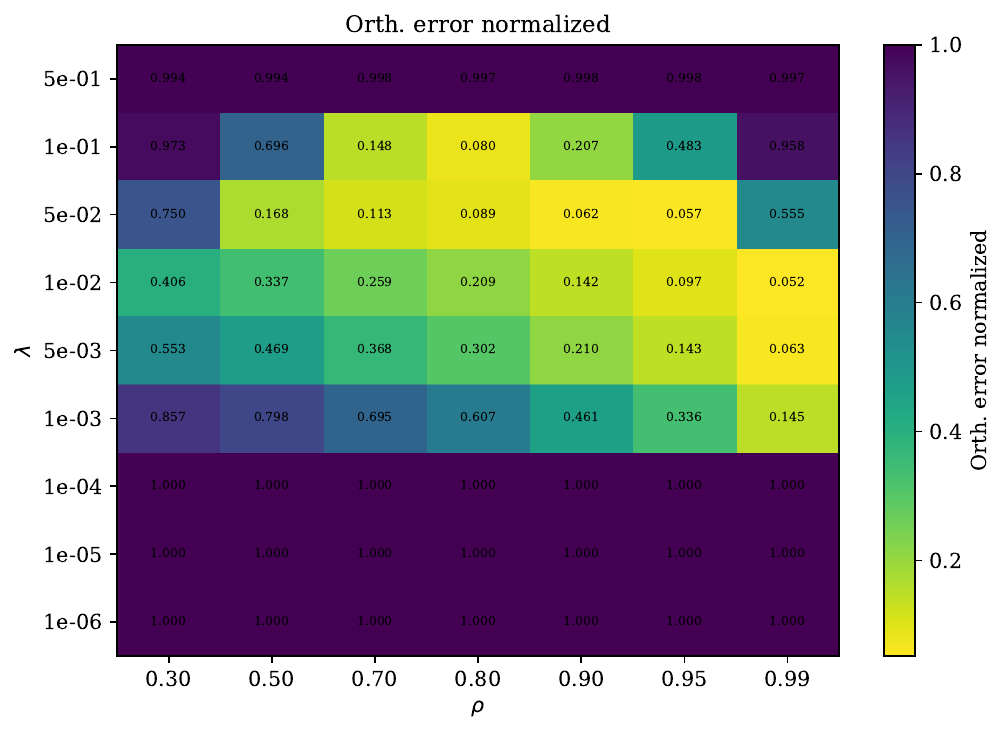}
    \caption{Grid search over the regularization weight $\lambda$ and OU correlation $\rho$ (5 seeds each, mean $\pm$ std). \textbf{Left:} Linear $R^2$ between true and learned latents ($\uparrow$ better). \textbf{Right:} Normalized orthogonality error of the fitted linear map ($\downarrow$ better). Identifiability requires both objectives to be active: too much Gaussianity ($\lambda = 0.5$) collapses the representation entirely ($R^2 \approx 0$), while even weak regularization ($\lambda \leq 10^{-2}$) suffices when the alignment signal is strong. At very high correlation ($\rho = 0.99$), performance degrades for moderate $\lambda$ because near-identical positive pairs make the invariance loss trivially small, allowing SIGReg to dominate. The best identifiability ($R^2 > 0.96$, orthogonality error $< 0.2$) occurs at $\lambda \in [10^{-3}, 10^{-2}]$ with $\rho \in [0.8, 0.95]$.}
    \label{fig:grid}
\end{figure}

Figure~\ref{fig:grid} shows the results. Three regimes are visible. First, when Gaussianity regularization is too strong ($\lambda = 0.5$), the encoder collapses to a non-informative Gaussian representation with $R^2 \approx 0$, regardless of $\rho$: the regularizer overwhelms the alignment signal. Second, at low $\lambda$ ($\leq 10^{-2}$), identifiability improves monotonically with $\rho$, consistent with the theoretical prediction that the correlation bound tightens as $\rho \to 1$. The best performance ($R^2 > 0.97$, orthogonality error $\approx 0.15$) is achieved at $\lambda \in \{10^{-3}, 5 \times 10^{-3}\}$ with $\rho \in \{0.9, 0.95\}$. Third, at intermediate $\lambda$ ($5 \times 10^{-2}$ to $10^{-1}$), there is a non-monotonic interaction: performance peaks at moderate $\rho$ and degrades at $\rho = 0.99$. This occurs because very high correlation makes the invariance loss trivially small, shifting the loss balance toward Gaussianity and pushing the system into the collapse regime.

\subsection{Latent Variable Distribution Sweep Across Mixings}
\label{app:gennorm}

\begin{figure}[h]
    \centering
    \includegraphics[width=\textwidth]{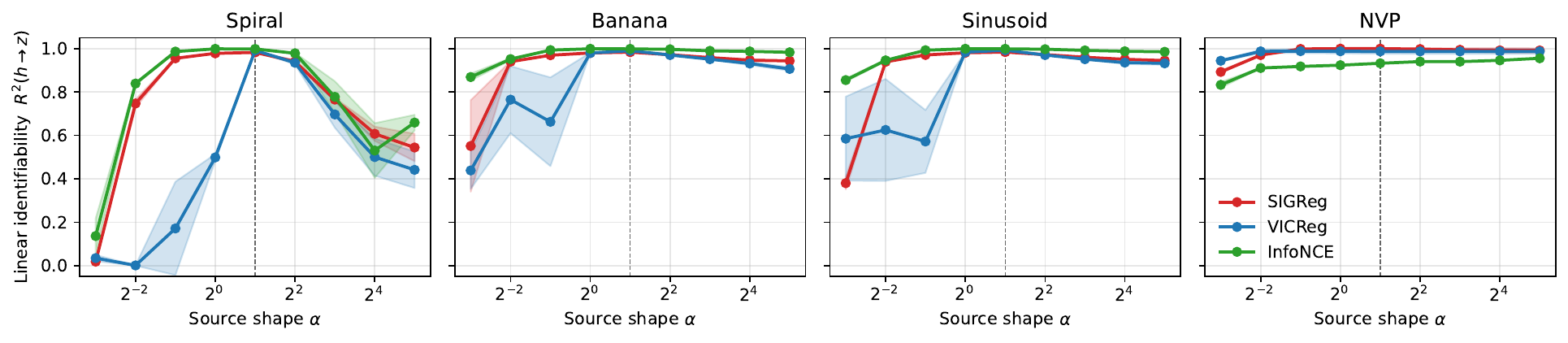}
    \caption{
        \textbf{Linear identifiability across the gennorm family for all four 2D mixings.}
        $R^2(h \to z)$ on a fixed evaluation grid as a function of latent shape $\alpha$ ($\alpha{=}1$ Laplace, $\alpha{=}2$ Gaussian). All three methods peak at $\alpha = 2$ as predicted by Thm.~\ref{thm:converse}; SIGReg and InfoNCE, which constrain $h$ beyond second moments, retain a wider plateau than VICReg for heavy-tailed latents. Mean $\pm$ std, 3 seeds.
        \label{fig:gennorm_full}
    }
\end{figure}

\begin{figure}[h]
    \centering
    \includegraphics[width=\textwidth]{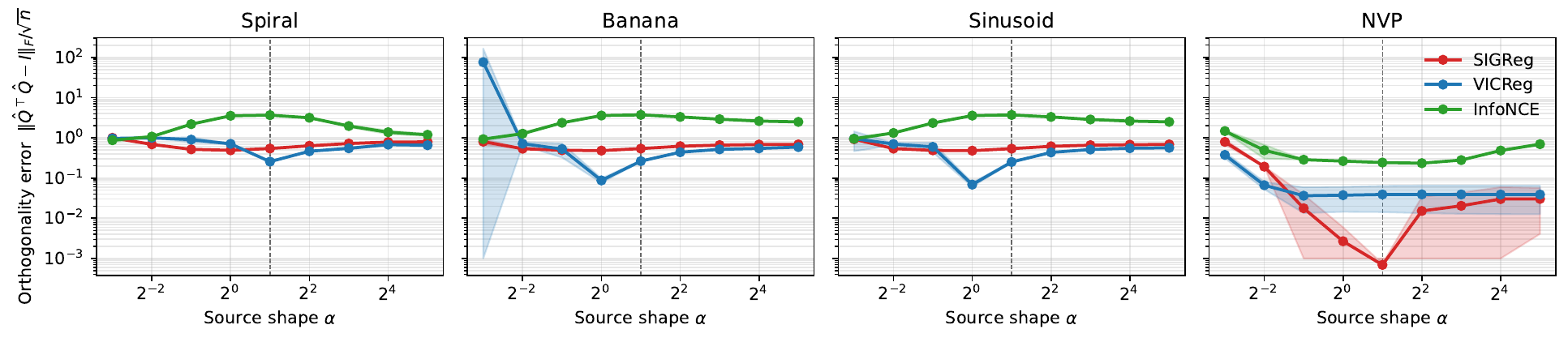}
    \caption{
        \textbf{Orthogonality error across the gennorm family for all four 2D mixings.}
        $\|\hat Q^\top \hat Q - I\|_F / \sqrt{N}$ on the same fixed grid (log scale). VICReg and SIGReg dip near $\alpha=2$ where their constraints align with the latent distribution; InfoNCE remains roughly flat, reflecting weaker control over the linear map under fixed kernel width. Mean $\pm$ std over 3 seeds.
        \label{fig:gennorm_orth}
    }
\end{figure}

The main-text gennorm figure (Fig.~\ref{fig:bound_unique_plan}b) shows linear recovery on the spiral mixing across latent distribution shapes $\alpha \in \{2^{-3}, \ldots, 2^{5}\}$ in the generalized normal family ($\alpha=2$ Gaussian, $\alpha=1$ Laplace, $\alpha \to \infty$ uniform). We extend this sweep to all four 2D mixings used in our paper (Figs.~\ref{fig:gennorm_full},~\ref{fig:gennorm_orth}).

Two patterns are robust across mixings. First, both methods peak at $\alpha=2$, confirming Thm.~\ref{thm:converse} empirically: linear identifiability is maximal precisely at the Gaussian latents. Second, SIGReg's plateau is consistently wider than whitening's. Whitening collapses for $\alpha < 2$ (heavier-than-Gaussian tails), whereas SIGReg retains high recovery across a broad range of latent distribution shapes. This confirms that SIGReg's Gaussianization of $h$ is doing more than enforcing covariance: it absorbs latent-side non-Gaussianity that whitening cannot see.

The contrast between methods is most pronounced for the spiral mixing, the only one of the four that is approximately measure-preserving and therefore unconstrained by second-order statistics alone. For the banana, sinusoid, and NVP mixings, the mixing structure itself partially constrains the recovery, narrowing the gap between methods. The orthogonality error (Fig.~\ref{fig:gennorm_orth}) tells the same story from the geometric side: whitening achieves a minimum at $\alpha=2$ where the second-order constraint exactly matches the latents, while SIGReg's basin is broader.

\subsection{Approximate Bound Decomposition}\label{app:bound}

\begin{figure}[h]
    \centering
    \includegraphics[width=\textwidth]{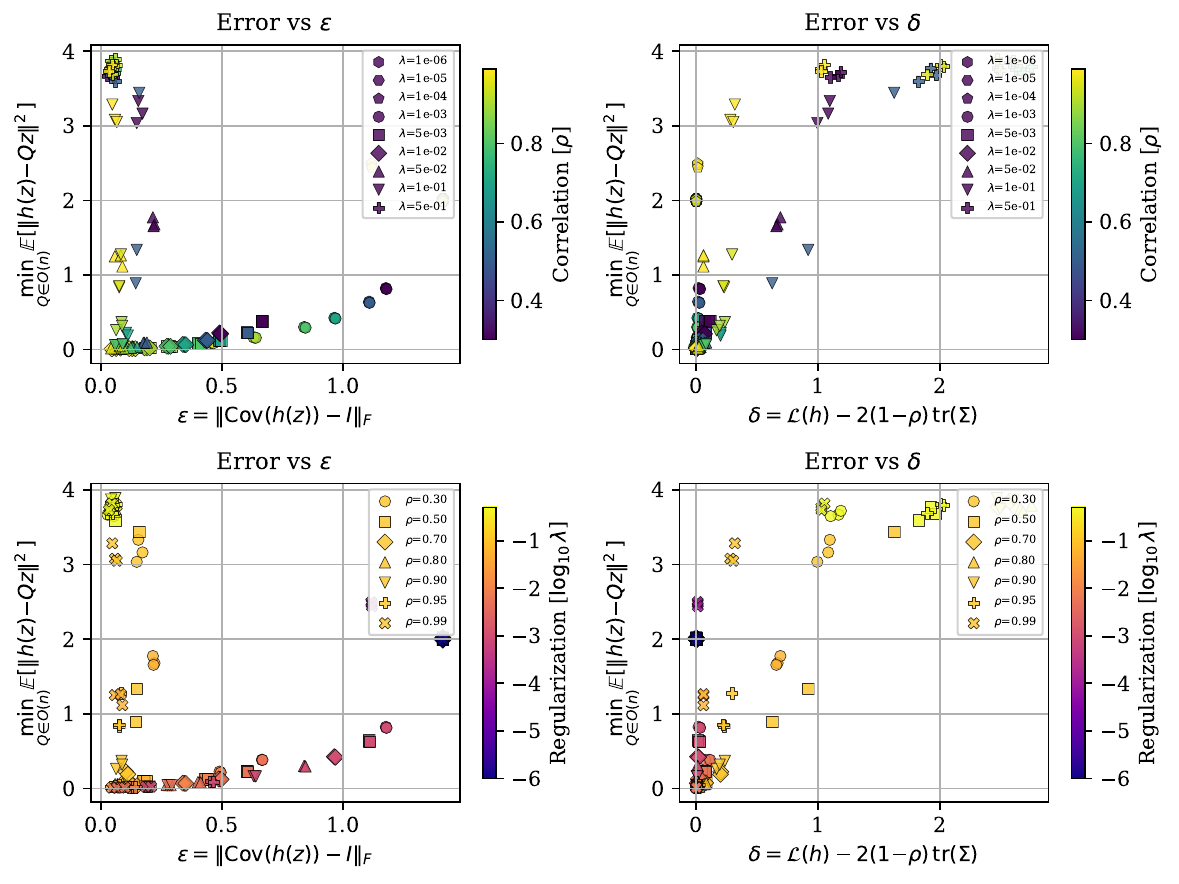}
    \caption{Decomposition of the recovery error into its two sources. \textbf{Left column:} Error vs whitening deviation $\varepsilon$. \textbf{Right column:} Error vs alignment gap $\delta$. Top and bottom rows use different marker/color encodings for $\lambda$ and $\rho$. The alignment gap $\delta$ is a much stronger predictor of recovery error than the whitening error $\varepsilon$, confirming that approximate whitening is the easy part; the binding constraint is alignment quality.}
    \label{fig:bound-decomp}
\end{figure}

Figure~\ref{fig:bound-decomp} decomposes the recovery error into its two sources: the whitening error $\varepsilon$ and the alignment gap $\delta$. The two quantities have distinct roles. Large $\varepsilon$ alone does not predict poor recovery: the $\lambda = 5 \times 10^{-1}$ runs (crosses) achieve near-zero $\varepsilon$ because SIGReg dominates, yet their recovery error is maximal because the alignment signal is absent ($\delta$ is large). Conversely, the alignment gap $\delta$ is a much stronger predictor of actual recovery error, consistent with the interpretation that the $D$ term in the bound captures the nonlinear energy that alignment fails to suppress. This asymmetry confirms the practical takeaway from Thm.~\ref{thm:approx}: approximate whitening is easy to achieve in practice, so the binding constraint on identifiability is the quality of the alignment objective.

\subsection{Loss Predictivity}\label{app:loss}

\begin{figure}[t]
    \centering
    \includegraphics[width=0.6\linewidth]{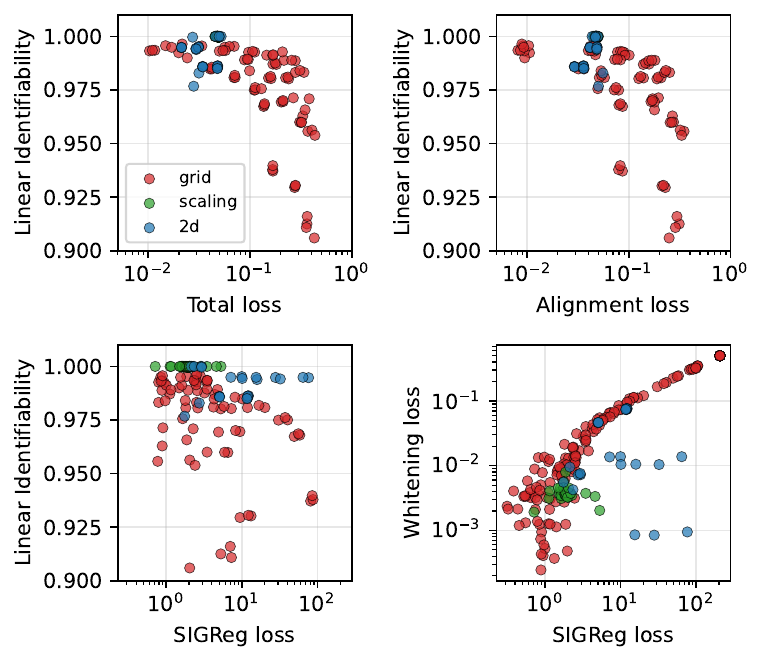}
    \caption{
    \textbf{Losses predict linear identifiability.}
    Each point is one trained encoder from one of our three experiments (2D illustrations, grid search, scaling), zoomed to the converged regime ($R^2 > 0.9$).
    \textbf{Top left:} Total loss (alignment + $\lambda$ SIGReg) correlates with linear $R^2(h \to z)$.
    \textbf{Top right:} Alignment loss alone is predictive of identifiability quality.
    \textbf{Bottom left:} SIGReg loss vs.\ $R^2$.
    \textbf{Bottom right:} SIGReg and whitening losses are correlated.
    }
    \label{fig:scatter}
\end{figure}

Figure~\ref{fig:scatter} pools all trained encoders across our four experiments to examine whether the training losses are predictive of linear identifiability.
We restrict the view to converged runs ($R^2 > 0.9$) to focus on the regime where the theory applies.
The alignment loss is the strongest single predictor of $R^2$, consistent with its direct connection to the alignment gap $\delta$ in Thm.~\ref{thm:approx}.

\paragraph{SIGReg does not vanish at the optimum.}
SIGReg is a goodness-of-fit statistic of Epps--Pulley type, multiplied by the batch size \citep{balestriero2025lejepa}. Under the null, i.e., for embeddings that are exactly $\mathcal{N}(0, I)$, the bias of the estimator vanishes with the batch size while the multiplication holds its variance at order one, so the statistic converges to a constant of order one rather than to zero. On true isotropic Gaussian samples at our batch size and embedding dimension it reads $1.04$; in identified runs it settles at $1.8$--$2.0$ for all $N \geq 16$, independently of the dimension up to $N{=}4096$ (Tabs.~\ref{tab:scaling-sigreg},~\ref{tab:scaling-4096}). A value of order one therefore indicates that the target is met up to sampling noise, not a failure to optimize, and the remaining gap is what Thm.~\ref{thm:approx} accounts for through $\varepsilon$.

\subsection{Scaling Experiment}\label{app:scaling}

\begin{table}[t]
\centering
\caption{\textbf{Scaling Experiment (SIGReg)} (mean $\pm$ std, 5 seeds). The RealNVP mixing is consistently nonlinear across dimensions ($R^2(x \to z) < 1$). The learned model nonetheless recovers the true latents at all dimensions. Training losses are stable; orthogonality error grows gradually with $N$. $^\dagger$Batch size 1024, 3 seeds (App.~\ref{app:scaling}, Tab.~\ref{tab:scaling-4096}).}
\label{tab:scaling-sigreg}
\resizebox{\textwidth}{!}{%
\begin{tabular}{r cc cc c cc}
\toprule
 \multicolumn{1}{c}{\textbf{Latents}} & \multicolumn{2}{c}{\textbf{Mixing difficulty}} & \multicolumn{2}{c}{\textbf{Linear identifiability}} & \multicolumn{1}{c}{\textbf{Orthogonality}} & \multicolumn{2}{c}{\textbf{SIGReg losses}} \\
\cmidrule(lr){1-1} \cmidrule(lr){2-3} \cmidrule(lr){4-5} \cmidrule(lr){6-6} \cmidrule(lr){7-8}
$N$ & $R^2(z \to x)$ & $R^2(x \to z)$ & $R^2(z \to h)$ & $R^2(h \to z)$ & $\|\hat Q^\top \hat Q - I\|_F / \sqrt{N}$ & Align & SIGReg \\
 & {\scriptsize $\pm$std\,$\times 10^{-3}$} & {\scriptsize $\pm$std\,$\times 10^{-3}$} & {\scriptsize $\pm$std\,$\times 10^{-7}$} & {\scriptsize $\pm$std\,$\times 10^{-7}$} & {\scriptsize $\pm$std\,$\times 10^{-5}$} & {\scriptsize $\pm$std\,$\times 10^{-4}$} & {\scriptsize $\pm$std} \\
\midrule
  $2^{1}$ & 0.871\tiny{$\pm$1.4} & 0.782\tiny{$\pm$2.8} & 0.99999\tiny{$\pm$3.4} & 0.99999\tiny{$\pm$3.4} & 0.000\tiny{$\pm$24} & 0.0472\tiny{$\pm$15} & 2.54\tiny{$\pm$1.4} \\
  $2^{2}$ & 0.865\tiny{$\pm$16} & 0.724\tiny{$\pm$24} & 0.99999\tiny{$\pm$12} & 0.99999\tiny{$\pm$12} & 0.001\tiny{$\pm$12} & 0.0475\tiny{$\pm$10} & 2.09\tiny{$\pm$1.3} \\
  $2^{3}$ & 0.860\tiny{$\pm$4.2} & 0.729\tiny{$\pm$10} & 0.99999\tiny{$\pm$9.0} & 0.99999\tiny{$\pm$9.0} & 0.001\tiny{$\pm$14} & 0.0473\tiny{$\pm$13} & 2.06\tiny{$\pm$0.78} \\
  $2^{4}$ & 0.858\tiny{$\pm$6.9} & 0.733\tiny{$\pm$6.0} & 0.99998\tiny{$\pm$4.9} & 0.99998\tiny{$\pm$4.9} & 0.002\tiny{$\pm$18} & 0.0476\tiny{$\pm$3.6} & 1.76\tiny{$\pm$0.18} \\
  $2^{5}$ & 0.851\tiny{$\pm$3.0} & 0.738\tiny{$\pm$1.6} & 0.99998\tiny{$\pm$7.2} & 0.99998\tiny{$\pm$7.2} & 0.003\tiny{$\pm$7.6} & 0.0479\tiny{$\pm$5.0} & 1.93\tiny{$\pm$0.17} \\
  $2^{6}$ & 0.854\tiny{$\pm$1.5} & 0.737\tiny{$\pm$1.4} & 0.99996\tiny{$\pm$7.6} & 0.99996\tiny{$\pm$7.4} & 0.005\tiny{$\pm$10} & 0.0484\tiny{$\pm$3.9} & 2.03\tiny{$\pm$0.14} \\
  $2^{7}$ & 0.853\tiny{$\pm$0.62} & 0.739\tiny{$\pm$0.50} & 0.99993\tiny{$\pm$3.1} & 0.99993\tiny{$\pm$3.2} & 0.007\tiny{$\pm$7.4} & 0.0486\tiny{$\pm$1.2} & 1.80\tiny{$\pm$0.09} \\
  $2^{8}$ & 0.855\tiny{$\pm$0.30} & 0.742\tiny{$\pm$0.48} & 0.99988\tiny{$\pm$7.7} & 0.99988\tiny{$\pm$7.9} & 0.010\tiny{$\pm$5.0} & 0.0486\tiny{$\pm$2.9} & 1.88\tiny{$\pm$0.17} \\
  $2^{9}$ & 0.859\tiny{$\pm$0.24} & 0.749\tiny{$\pm$0.30} & 0.99977\tiny{$\pm$6.5} & 0.99977\tiny{$\pm$6.7} & 0.015\tiny{$\pm$3.7} & 0.0489\tiny{$\pm$2.4} & 1.93\tiny{$\pm$0.08} \\
  $2^{10}$ & 0.866\tiny{$\pm$0.10} & 0.763\tiny{$\pm$0.17} & 0.99956\tiny{$\pm$12} & 0.99956\tiny{$\pm$12} & 0.023\tiny{$\pm$4.7} & 0.0488\tiny{$\pm$0.91} & 1.93\tiny{$\pm$0.06} \\
  $2^{11}$ & 0.869\tiny{$\pm$0.03} & 0.768\tiny{$\pm$0.05} & 0.99904\tiny{$\pm$8.1} & 0.99904\tiny{$\pm$7.9} & 0.035\tiny{$\pm$1.5} & 0.0491\tiny{$\pm$0.48} & 1.88\tiny{$\pm$0.07} \\
  $2^{12\,\dagger}$ & 0.869\tiny{$\pm$0.04} & 0.768\tiny{$\pm$0.02} & 0.99945\tiny{$\pm$1.7} & 0.99944\tiny{$\pm$1.9} & 0.026\tiny{$\pm$0.90} & 0.0489\tiny{$\pm$0.14} & 1.91\tiny{$\pm$0.11} \\
\bottomrule
\end{tabular}}
\end{table}

\begin{table}[t]
\centering
\caption{\textbf{Scaling Experiment (VICReg)} (mean $\pm$ std, 5 seeds). VICReg matches SIGReg on linear identifiability across all dimensions; orthogonality error grows similarly with $N$. The whitening loss is stable and small throughout.}
\label{tab:scaling-vicreg}
\resizebox{\textwidth}{!}{%
\begin{tabular}{r cc cc c cc}
\toprule
 \multicolumn{1}{c}{\textbf{Latents}} & \multicolumn{2}{c}{\textbf{Mixing difficulty}} & \multicolumn{2}{c}{\textbf{Linear identifiability}} & \multicolumn{1}{c}{\textbf{Orthogonality}} & \multicolumn{2}{c}{\textbf{VICReg losses}} \\
\cmidrule(lr){1-1} \cmidrule(lr){2-3} \cmidrule(lr){4-5} \cmidrule(lr){6-6} \cmidrule(lr){7-8}
$N$ & $R^2(z \to x)$ & $R^2(x \to z)$ & $R^2(z \to h)$ & $R^2(h \to z)$ & $\|\hat Q^\top \hat Q - I\|_F / \sqrt{N}$ & Align & Whitening \\
 & {\scriptsize $\pm$std\,$\times 10^{-3}$} & {\scriptsize $\pm$std\,$\times 10^{-3}$} & {\scriptsize $\pm$std\,$\times 10^{-7}$} & {\scriptsize $\pm$std\,$\times 10^{-7}$} & {\scriptsize $\pm$std\,$\times 10^{-5}$} & {\scriptsize $\pm$std\,$\times 10^{-4}$} & {\scriptsize $\pm$std\,$\times 10^{-4}$} \\
\midrule
  $2^{1}$ & 0.871\tiny{$\pm$0.66} & 0.782\tiny{$\pm$1.7} & 0.99999\tiny{$\pm$8.4} & 0.99999\tiny{$\pm$8.4} & 0.001\tiny{$\pm$19} & 0.0491\tiny{$\pm$24} & 0.0022\tiny{$\pm$13} \\
  $2^{2}$ & 0.865\tiny{$\pm$15} & 0.726\tiny{$\pm$24} & 0.99998\tiny{$\pm$54} & 0.99998\tiny{$\pm$54} & 0.001\tiny{$\pm$65} & 0.0476\tiny{$\pm$11} & 0.0041\tiny{$\pm$13} \\
  $2^{3}$ & 0.869\tiny{$\pm$4.4} & 0.731\tiny{$\pm$8.1} & 0.99998\tiny{$\pm$4.9} & 0.99998\tiny{$\pm$4.8} & 0.002\tiny{$\pm$31} & 0.0476\tiny{$\pm$13} & 0.0045\tiny{$\pm$12} \\
  $2^{4}$ & 0.860\tiny{$\pm$6.9} & 0.733\tiny{$\pm$5.9} & 0.99998\tiny{$\pm$4.7} & 0.99998\tiny{$\pm$4.6} & 0.002\tiny{$\pm$12} & 0.0476\tiny{$\pm$3.6} & 0.0036\tiny{$\pm$2.8} \\
  $2^{5}$ & 0.855\tiny{$\pm$1.5} & 0.736\tiny{$\pm$1.9} & 0.99998\tiny{$\pm$9.4} & 0.99998\tiny{$\pm$9.4} & 0.003\tiny{$\pm$3.6} & 0.0477\tiny{$\pm$6.7} & 0.0035\tiny{$\pm$3.3} \\
  $2^{6}$ & 0.853\tiny{$\pm$1.7} & 0.737\tiny{$\pm$1.4} & 0.99996\tiny{$\pm$8.2} & 0.99996\tiny{$\pm$8.1} & 0.004\tiny{$\pm$9.0} & 0.0488\tiny{$\pm$2.1} & 0.0035\tiny{$\pm$1.8} \\
  $2^{7}$ & 0.853\tiny{$\pm$0.87} & 0.739\tiny{$\pm$0.66} & 0.99994\tiny{$\pm$7.1} & 0.99994\tiny{$\pm$7.2} & 0.006\tiny{$\pm$9.5} & 0.0487\tiny{$\pm$0.99} & 0.0035\tiny{$\pm$0.75} \\
  $2^{8}$ & 0.855\tiny{$\pm$0.25} & 0.742\tiny{$\pm$0.49} & 0.99988\tiny{$\pm$7.2} & 0.99988\tiny{$\pm$7.2} & 0.009\tiny{$\pm$4.1} & 0.0486\tiny{$\pm$2.9} & 0.0035\tiny{$\pm$0.39} \\
  $2^{9}$ & 0.858\tiny{$\pm$0.23} & 0.749\tiny{$\pm$0.30} & 0.99978\tiny{$\pm$7.2} & 0.99978\tiny{$\pm$6.9} & 0.014\tiny{$\pm$3.7} & 0.0489\tiny{$\pm$2.4} & 0.0035\tiny{$\pm$0.29} \\
  $2^{10}$ & 0.866\tiny{$\pm$0.10} & 0.763\tiny{$\pm$0.17} & 0.99958\tiny{$\pm$11} & 0.99958\tiny{$\pm$11} & 0.022\tiny{$\pm$4.5} & 0.0488\tiny{$\pm$0.91} & 0.0035\tiny{$\pm$0.06} \\
  $2^{11}$ & 0.869\tiny{$\pm$0.03} & 0.768\tiny{$\pm$0.05} & 0.99909\tiny{$\pm$7.5} & 0.99909\tiny{$\pm$7.5} & 0.033\tiny{$\pm$1.3} & 0.0490\tiny{$\pm$0.47} & 0.0035\tiny{$\pm$0.11} \\
\bottomrule
\end{tabular}}
\end{table}

\begin{table}[t]
\centering
\caption{\textbf{Scaling Experiment (InfoNCE)} (mean $\pm$ std, 5 seeds). InfoNCE matches the batch-statistic methods at low $N$ but degrades at scale under a fixed Gaussian kernel width. 
The InfoNCE loss column shows the saturation directly: at $N \geq 32$, both positive- and negative-pair similarities $\exp(-\|\cdot\|^2/\sigma^2)$ underflow to zero, killing the gradient. The non-monotone $R^2$ across $N$ (note the spike at $N{=}16$) is the fingerprint of this kernel-width mismatch rather than a property of the InfoNCE objective itself.
}
\label{tab:scaling-infonce}
\resizebox{\textwidth}{!}{%
\begin{tabular}{r cc cc c cc}
\toprule
 \multicolumn{1}{c}{\textbf{Latents}} & \multicolumn{2}{c}{\textbf{Mixing difficulty}} & \multicolumn{2}{c}{\textbf{Linear identifiability}} & \multicolumn{1}{c}{\textbf{Orthogonality}} & \multicolumn{2}{c}{\textbf{InfoNCE losses}} \\
\cmidrule(lr){1-1} \cmidrule(lr){2-3} \cmidrule(lr){4-5} \cmidrule(lr){6-6} \cmidrule(lr){7-8}
$N$ & $R^2(z \to x)$ & $R^2(x \to z)$ & $R^2(z \to h)$ & $R^2(h \to z)$ & $\|\hat Q^\top \hat Q - I\|_F / \sqrt{N}$ & Align & InfoNCE \\
 & {\scriptsize $\pm$std\,$\times 10^{-3}$} & {\scriptsize $\pm$std\,$\times 10^{-3}$} & {\scriptsize $\pm$std\,$\times 10^{-3}$} & {\scriptsize $\pm$std\,$\times 10^{-3}$} & {\scriptsize $\pm$std} & {\scriptsize $\pm$std\,$\times 10^{-3}$} & {\scriptsize $\pm$std} \\
\midrule
  $2^{1}$ & 0.870\tiny{$\pm$0.92} & 0.781\tiny{$\pm$1.4} & 0.96030\tiny{$\pm$2.4} & 0.95096\tiny{$\pm$1.6} & 0.505\tiny{$\pm$0.06} & 0.0653\tiny{$\pm$5.1} & 4.529\tiny{$\pm$0.03} \\
  $2^{2}$ & 0.855\tiny{$\pm$11} & 0.730\tiny{$\pm$23} & 0.93874\tiny{$\pm$8.1} & 0.91087\tiny{$\pm$8.2} & 0.908\tiny{$\pm$0.12} & 0.0866\tiny{$\pm$5.0} & 3.557\tiny{$\pm$0.03} \\
  $2^{3}$ & 0.870\tiny{$\pm$2.0} & 0.724\tiny{$\pm$12} & 0.92503\tiny{$\pm$25} & 0.88681\tiny{$\pm$42} & 0.876\tiny{$\pm$0.11} & 0.0883\tiny{$\pm$11} & 1.811\tiny{$\pm$0.05} \\
  $2^{4}$ & 0.858\tiny{$\pm$6.4} & 0.736\tiny{$\pm$6.8} & 0.99988\tiny{$\pm$0.01} & 0.99988\tiny{$\pm$0.01} & 0.010\tiny{$\pm$0.00} & 0.0489\tiny{$\pm$1.2} & 0.156\tiny{$\pm$0.01} \\
  $2^{5}$ & 0.853\tiny{$\pm$2.9} & 0.737\tiny{$\pm$2.8} & 0.92206\tiny{$\pm$18} & 0.90780\tiny{$\pm$26} & 0.393\tiny{$\pm$0.07} & 0.0738\tiny{$\pm$6.5} & 0.000\tiny{$\pm$0.00} \\
  $2^{6}$ & 0.853\tiny{$\pm$1.7} & 0.737\tiny{$\pm$1.4} & 0.77351\tiny{$\pm$0.82} & 0.64849\tiny{$\pm$3.1} & 1.756\tiny{$\pm$0.04} & 0.2123\tiny{$\pm$3.5} & 0.000\tiny{$\pm$0.00} \\
  $2^{7}$ & 0.853\tiny{$\pm$0.87} & 0.739\tiny{$\pm$0.66} & 0.77232\tiny{$\pm$5.2} & 0.56695\tiny{$\pm$6.6} & 4.875\tiny{$\pm$0.08} & 0.3803\tiny{$\pm$6.2} & 0.000\tiny{$\pm$0.00} \\
  $2^{8}$ & 0.855\tiny{$\pm$0.25} & 0.742\tiny{$\pm$0.49} & 0.83727\tiny{$\pm$0.51} & 0.69658\tiny{$\pm$0.49} & 3.922\tiny{$\pm$0.04} & 0.2519\tiny{$\pm$2.5} & 0.000\tiny{$\pm$0.00} \\
  $2^{9}$ & 0.858\tiny{$\pm$0.23} & 0.749\tiny{$\pm$0.30} & 0.84131\tiny{$\pm$0.15} & 0.70439\tiny{$\pm$0.26} & 3.946\tiny{$\pm$0.00} & 0.2507\tiny{$\pm$0.95} & 0.000\tiny{$\pm$0.00} \\
  $2^{10}$ & 0.866\tiny{$\pm$0.10} & 0.763\tiny{$\pm$0.17} & 0.84972\tiny{$\pm$0.18} & 0.72024\tiny{$\pm$0.20} & 4.003\tiny{$\pm$0.00} & 0.2507\tiny{$\pm$1.1} & 0.000\tiny{$\pm$0.00} \\
  $2^{11}$ & 0.869\tiny{$\pm$0.03} & 0.768\tiny{$\pm$0.05} & 0.85343\tiny{$\pm$0.07} & 0.72714\tiny{$\pm$0.10} & 4.028\tiny{$\pm$0.00} & 0.2514\tiny{$\pm$0.84} & 0.000\tiny{$\pm$0.00} \\
\bottomrule
\end{tabular}}
\end{table}

We test whether identifiability holds at scale by sweeping the latent dimension $N \in \{2^1, \ldots, 2^{12}\}$ with 5 seeds each; for $N > 1250$ the evaluation set grows to $8N$ points.
The mixing function is a RealNVP-style coupling layer architecture (4 layers with orthogonal weight matrices) \citep{dinh2016density}, and the encoder is a \emph{matched} inverse-coupling-layer architecture with learnable weights.
This architectural matching removes encoder expressivity as a confounder: any failure of identifiability is due to the optimization landscape, not the encoder's function class.
For $N \leq 32$, we train $K=3$ encoders per seed and select the one with lowest final loss; for $N > 32$, all runs converge to equivalent solutions ($K=1$).

Tabs.~\ref{tab:scaling-sigreg}--\ref{tab:scaling-infonce} report the per-method results. The mixing difficulty is consistently nonlinear across dimensions ($R^2(x \to z) \approx 0.73$--$0.78$) for all methods, since the RealNVP mixing is shared. The three methods differ in their failure modes:

\textbf{SIGReg} (Tab.~\ref{tab:scaling-sigreg}). Linear identifiability is essentially perfect at all dimensions ($R^2 > 0.999$, at $N{=}4096$ with batch size 1024). Alignment and SIGReg losses are stable across $N$. The only quantity that drifts is the orthogonality error $\|\hat Q^\top \hat Q - I\|_F / \sqrt{N}$, which grows from $\sim 10^{-5}$ at $N{=}2$ to $\sim 2 \times 10^{-2}$ at $N{=}2048$. This is the slow degradation predicted by Thm.~\ref{thm:approx}: at fixed alignment quality, the bound on the linear-part distortion grows with the dimension being whitened.

\textbf{VICReg} (Tab.~\ref{tab:scaling-vicreg}). Numerically indistinguishable from SIGReg across the entire sweep, on both linear identifiability and orthogonality error. The whitening loss is small ($\sim 3.5 \times 10^{-3}$) and stable. Confirms that for Gaussian latents at the population optimum, second-moment whitening is sufficient (the Gaussianity assumption on $h(z)$ in Thm.~\ref{thm:main} can be relaxed to whitening, see Sec.~\ref{sec:converse}).

\textbf{InfoNCE} (Tab.~\ref{tab:scaling-infonce}). Qualitatively different. $R^2$ is non-monotone in $N$: high at $N{=}16$ ($\sim 0.9999$), then collapsing to $\sim 0.57$--$0.73$ for $N \geq 64$. Orthogonality error is one to three orders of magnitude larger than for the batch-statistic methods. The InfoNCE loss column reveals the mechanism: at $N \geq 32$, the loss saturates at zero, indicating that the Gaussian kernel $\exp(-\|\cdot\|^2/\sigma^2)$ underflows for both positive and negative pairs once $\|z\|^2 \approx N$ at $\sigma{=}1$, killing the gradient. The non-monotone $R^2$ is the fingerprint of this kernel-width mismatch, not a property of the InfoNCE objective itself; per-dimension kernel tuning would presumably restore performance, at the cost of a hyperparameter sweep absent from the batch-statistic methods.

These three patterns make the main-text story precise: the practical gap between methods is not in their theoretical identifiability guarantees (Thm.~\ref{thm:main} applies to all), but in how each method's optimization behaves at scale.

\begin{figure}[h]
    \centering
    \includegraphics[width=0.9\textwidth]{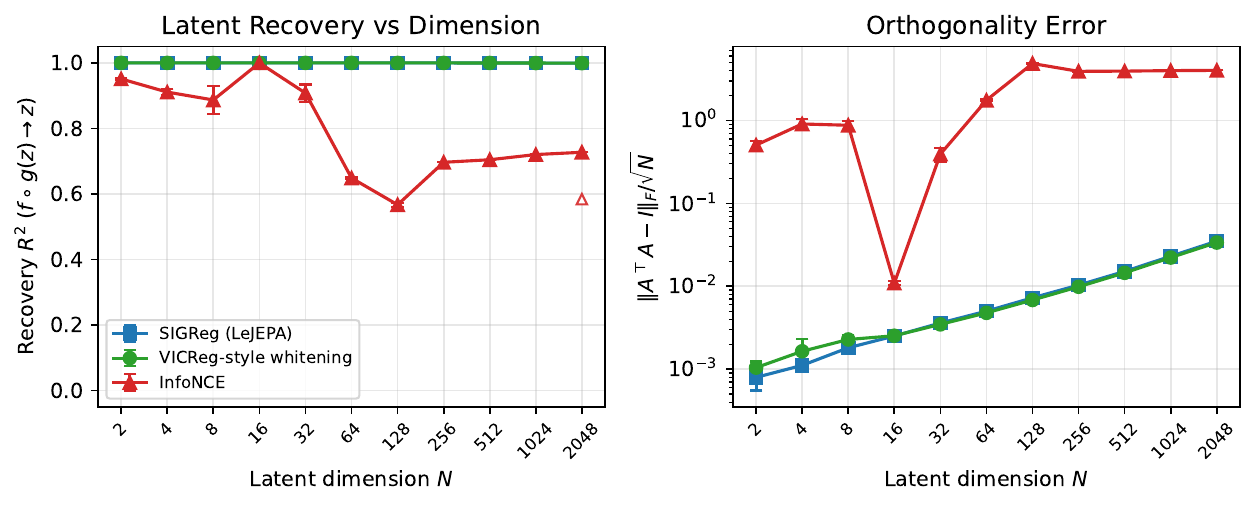}
    \caption{
        \textbf{Scaling with latent dimension} (5 seeds, batch size 256).
        \textbf{Left:} Recovery $R^2(h \to z)$; open marker: held-out recovery for InfoNCE at $N{=}2048$.
        \textbf{Right:} Normalized orthogonality error of the fitted linear map (log scale), which grows gradually with $N$ for SIGReg and VICReg.
        $N{=}4096$ is shown in Tab.~\ref{tab:scaling-4096}.
    }
    \label{fig:scaling_modes}
\end{figure}

\paragraph{$N = 2^{12}$.}
At the default batch size of 256, all three objectives fail at $N{=}4096$ (Tab.~\ref{tab:scaling-4096}, Fig.~\ref{fig:scaling_modes}): the alignment loss stays at five times its value for $N \leq 2048$ and recovery drops to $0.65$--$0.73$. This is an estimation limit rather than an identifiability one. A batch of 256 is smaller than the dimension, so the batch statistics that SIGReg and VICReg constrain are rank-deficient; raising the batch size to 1024 restores SIGReg to $R^2 = 0.9994$ (3 seeds), whereas quadrupling the number of SIGReg slices at batch size 256 has no effect. We have not determined the exact transition batch size.

\begin{table}[h]
\centering
\caption{\textbf{Scaling diagnostic at $N{=}4096$.} Recovery, orthogonality error, and final alignment loss (mean $\pm$ std where more than one seed).}
\label{tab:scaling-4096}
\small
\begin{tabular}{l cc c ccc}
\toprule
Objective & Batch & Slices & Seeds & $R^2(h \to z)$ & $\|\hat Q^\top \hat Q - I\|_F / \sqrt{N}$ & Align \\
\midrule
SIGReg  & 256  & 256  & 5 & 0.663\tiny{$\pm$0.001} & 2.51\tiny{$\pm$0.15} & 0.272 \\
SIGReg  & 256  & 1024 & 1 & 0.663 & 2.27 & 0.270 \\
SIGReg  & 1024 & 256  & 3 & \textbf{0.99945}\tiny{$\pm$2e{-7}} & 0.027 & 0.049 \\
SIGReg  & 1024 & 1024 & 1 & \textbf{0.99945} & 0.027 & 0.049 \\
VICReg  & 256  & --   & 5 & 0.652\tiny{$\pm$0.002} & 2.09\tiny{$\pm$0.09} & 0.273 \\
InfoNCE & 256  & --   & 5 & 0.727\tiny{$\pm$0.000} & 4.03\tiny{$\pm$0.00} & 0.252 \\
\bottomrule
\end{tabular}
\end{table}

\subsection{Reacher Experiment}\label{app:reacher}

\begin{figure}[t]
    \centering
    \includegraphics[width=0.25\textwidth]{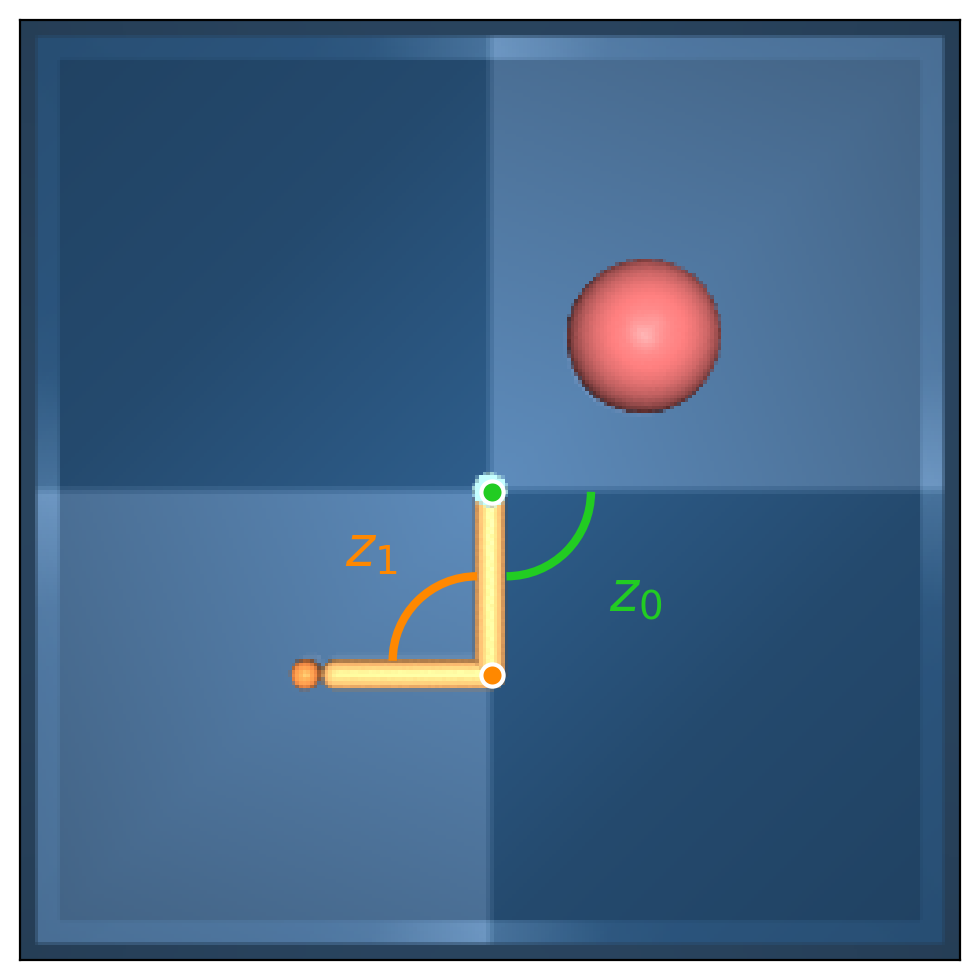}
    \caption{
        \textbf{DMC Reacher.} The latent state $z = (z_0, z_1)$ consists of two joint angles (shoulder and wrist) that fully determine the arm configuration. The nonlinear mixing $g$ is the MuJoCo rendering pipeline producing $64 \times 64$ pixel observations.
    }
    \label{fig:reacher_annotated}
\end{figure}

\paragraph{Environment and rendering.}
We use the DeepMind Control Suite Reacher (``hard'' variant)~\citep{tassa2018dmcontrol} with two revolute joints: a shoulder (unbounded) and a wrist (limited to $[-2.79, 2.79]$ rad).
The latent state is $z = (\theta_0, \theta_1) \in \mathbb{R}^2$ (Fig.~\ref{fig:reacher_annotated}).
Images are rendered at $64 \times 64$ pixels via MuJoCo with EGL headless rendering.
The target position is fixed at $(0.1, 0.1)$ across all experiments to eliminate it as a confound.
For each experimental condition, we pre-render 100{,}000 image pairs and 10{,}000 evaluation images (i.i.d.\ Gaussian samples, shared across all conditions).
Images are stored as uint8 and normalized per-channel at training time.

\textbf{OU condition.}
For each $\rho \in \{0.3, 0.5, 0.7, 0.8, 0.9, 0.95, 0.99\}$, we sample $z \sim \mathcal{N}(0, I_2)$, generate $z'$ via~\eqref{eq:dgp}, and render.
The theory's assumptions (Gaussian marginals, isotropic OU transitions, additive noise) are exactly satisfied.
Wrapping beyond $\pm\pi$ occurs for $<0.2\%$ of samples.

\begin{figure}[t]
    \centering
    \includegraphics[width=\textwidth]{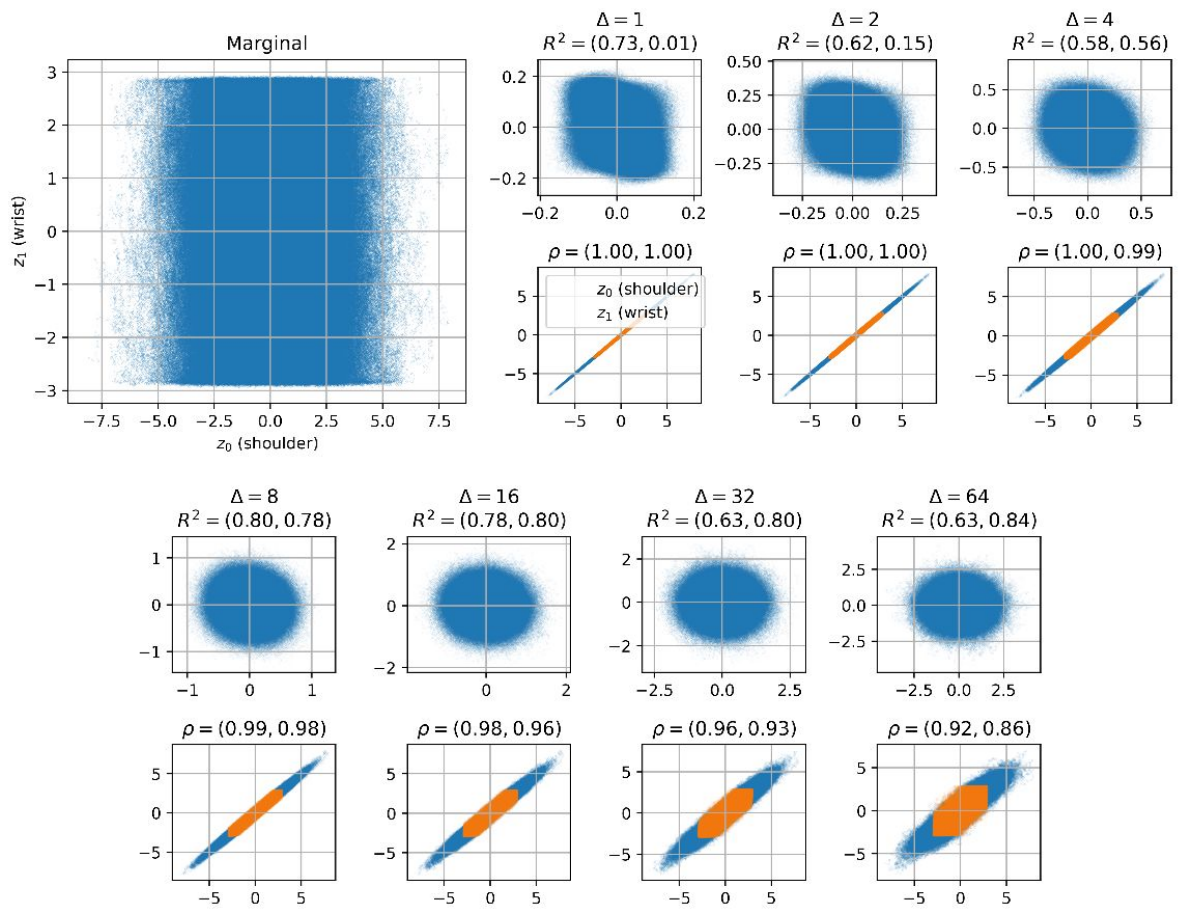}
    \caption{
        \textbf{Reacher trajectory latent distributions across temporal strides $\delta$.}
        \textbf{Left:} Stationary marginal of $(z_0, z_1)$; shoulder is broad, wrist is nearly bimodal.
        \textbf{Top row:} 2D transition differences $z_{t+\delta} - z_t$, with the per-dimension $R^2$ of the best-tuned encoder.
        \textbf{Bottom row:} Autocorrelation scatter $z_t$ vs.\ $z_{t+\delta}$, per dimension, with Pearson $\rho$ annotated.
        Small $\delta$: $\rho \approx 1$, transition is trivial, alignment carries no signal.
        Large $\delta$: $\rho$ has dropped but the difference distribution inherits the non-Gaussian structure of the stationary marginals.
        Intermediate $\delta$: both $\rho$ is informative and the differences are effectively Gaussian, and identifiability is highest.
    }
    \label{fig:traj_distributions}
\end{figure}

\textbf{Trajectory condition.}
We use 10{,}000 episodes (201 steps each) from the LeWorldModel~\citep{maes2026leworldmodel} dataset, collected by a trained SAC policy.
For each temporal stride $\delta \in \{1, 2, 4, 8, 16, 32, 64\}$, we subsample 10 pairs per episode (100{,}000 total), extract the joint angles $(\theta_0^t, \theta_1^t)$ and $(\theta_0^{t+\delta}, \theta_1^{t+\delta})$, and re-render with the fixed target.
The marginal distributions (Figure~\ref{fig:traj_distributions}) violate the Gaussian assumption: the shoulder is broad and heavy-tailed ($13.6\%$ beyond $\pm\pi$), while the wrist is nearly uniform over $[-\pi, \pi]$.
Autocorrelations are anisotropic (e.g., at $\delta=8$: $\rho_0 = 0.992$, $\rho_1 = 0.982$).

\paragraph{Encoder architecture.}
We use a CNN with BatchNorm:
\begin{align*}
    &\texttt{Conv}(3 \to 32, 4\times4, s{=}2) \to \texttt{BN} \to \texttt{GELU} \to
    \texttt{Conv}(32 \to 64) \to \texttt{BN} \to \texttt{GELU} \\
    &\to \texttt{Conv}(64 \to 128) \to \texttt{BN} \to \texttt{GELU} \to
    \texttt{Conv}(128 \to 256) \to \texttt{BN} \to \texttt{GELU} \\
    &\to \texttt{AvgPool}(4) \to \texttt{Flatten} \to \texttt{Linear}(256, 256) \to \texttt{BN} \to \texttt{GELU} \to \texttt{Linear}(256, 2).
\end{align*}
This architecture has ${\sim}1.1$M parameters.
BatchNorm was critical for stable training: without it, ${\sim}36\%$ of runs collapsed to zero-variance outputs.
Gradient clipping ($\|\nabla\|_{\max} = 1.0$) is applied after each backward pass.

\paragraph{Training.}
We train with AdamW (lr $= 3 \times 10^{-3}$, weight decay $10^{-4}$) and cosine annealing for 100 epochs, batch size 256, with SIGReg using 256 random slices.
The loss is $\mathcal{L} = \lambda \mathcal{L}_{\mathrm{SIG}} + (1 - \lambda) \mathcal{L}_{\mathrm{inv}}$.
We sweep $\lambda \in \{10^{-3}, 5 \times 10^{-3}, 10^{-2}, 5 \times 10^{-2}\}$ with 3 seeds per setting, reporting the best $\lambda$ per condition (mean $\pm$ std over seeds).
The final model (after 100 epochs) is always used; no early stopping or checkpoint selection is applied.

\paragraph{Evaluation.}
We fit a linear regression from embeddings to joint angles on 10{,}000 training images and score on 10{,}000 held-out evaluation images (proper train/test split).
We report bidirectional $R^2$, per-dimension $R^2$, and an $R^2$ diagnostic for $\sin/\cos$ targets $(\sin\theta_0, \cos\theta_0, \sin\theta_1, \cos\theta_1)$.
The $\sin/\cos$ diagnostic tests whether the encoder learns a trigonometric representation of the cyclic angles; negative values indicate the 2D embedding encodes raw angles (linear in $\theta$) rather than their trigonometric functions.

\paragraph{Latent distributions.}
Figure~\ref{fig:traj_distributions} shows the stationary joint-angle distribution and, for each stride $\delta$, the 2D transition-difference distribution together with the per-dimension autocorrelation scatter.
The stationary marginal (leftmost panel) reveals a striking asymmetry: the shoulder ($z_0$) is broad and mildly platykurtic, while the wrist ($z_1$) is nearly bimodal, reflecting a policy-induced preference for two joint-limit configurations.
The transition-difference panels (top row) grow from tight, narrow clouds at small $\delta$ to fully-developed isotropic clouds at $\delta = 64$.
The autocorrelation scatter (bottom row) confirms that both joints have $\rho \approx 1$ at $\delta = 1$ (the transition is nearly trivial) and decorrelate monotonically, with the wrist decorrelating faster.
The per-dimension $R^2$ values annotated on each panel already hint at the core phenomenon examined in the next paragraph: both small and large $\delta$ are bad, and identifiability peaks in an intermediate regime.

\begin{figure}[t]
    \centering
    \includegraphics[width=\textwidth]{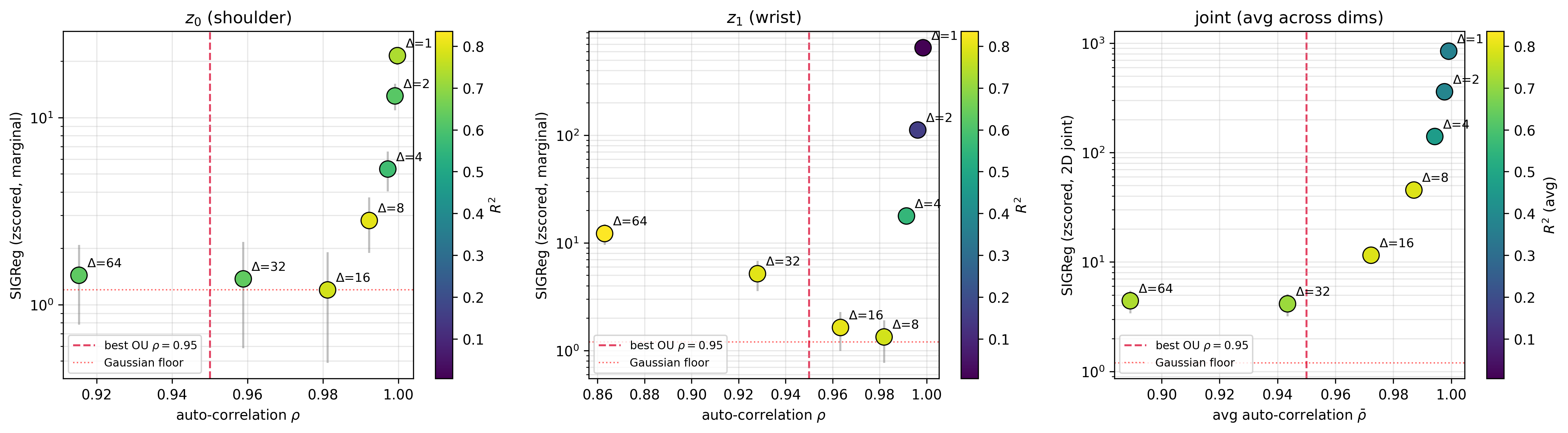}
    \caption{
        \textbf{Identifiability requires both approximate Gaussianity and non-trivial autocorrelation.}
        For each temporal stride $\delta$, we plot SIGReg (zscored, averaged over random projections and subsamples, error bars show one standard deviation) against the Pearson autocorrelation $\rho$ of the transition-difference distribution, colored by the corresponding identifiability $R^2$ (see Table~\ref{tab:reacher}).
        \textbf{Left:} shoulder ($z_0$, 1D marginal).
        \textbf{Middle:} wrist ($z_1$, 1D marginal).
        \textbf{Right:} 2D joint; $x$-axis is the average $\bar{\rho}$ across dimensions.
        Dotted red line: SIGReg floor from matched iid $\mathcal{N}(0, I)$ samples ($\approx 1.2$).
        Dashed red line: best-performing OU autocorrelation ($\rho = 0.95$).
        Small $\delta$ (top right): $\rho \approx 1$ yields no spectral gap, and small differences inherit the non-Gaussian stationary shape; both constraints fail.
        Large $\delta$ (far left): the wrist difference starts resolving the bimodal stationary, so SIGReg climbs again.
        Intermediate $\delta \in \{8, 16\}$: simultaneously near the Gaussian floor and in a regime of informative $\rho$; identifiability $R^2$ peaks here.
    }
    \label{fig:rho_vs_sigreg}
\end{figure}

\paragraph{Gaussianity and autocorrelation jointly determine identifiability.}
Thm.~\ref{thm:main} requires two conditions on the data-generating process: Gaussian latent marginals and a non-trivial autocorrelation $\rho \in (0, 1)$. Thm.~\ref{thm:approx} relaxes these to approximate whitening and a non-vanishing spectral gap.
The trajectory condition satisfies neither exactly, and the temporal stride $\delta$ controls how close each condition comes to being met.
To disentangle these two effects, we measure the (zscored) SIGReg value of each transition-difference distribution and of their 2D joint, and plot it against the Pearson autocorrelation, colored by identifiability $R^2$ (Figure~\ref{fig:rho_vs_sigreg}).
We compare to a finite-sample floor obtained by evaluating SIGReg on iid standard Gaussian samples of matched size; this floor is approximately $1.2$, so any value near $1$--$2$ is indistinguishable from Gaussian at this sample size.

Three regimes are visible.
At $\delta = 1, 2$ (top right of each panel), $\rho$ is essentially $1$ and SIGReg is two or three orders of magnitude above the floor: the transition is too trivial to carry an identifiability signal \emph{and} the differences inherit the non-Gaussian structure of the stationary marginals, because at this stride the differences $z_{t+\delta} - z_t$ are tiny perturbations that cannot Gaussianize via any averaging.
At $\delta = 64$ (far left), $\rho$ has dropped to $0.86$--$0.92$ and the shoulder differences are near the Gaussian floor, but the wrist differences climb again because the increments at this stride start to resolve the bimodal stationary distribution (they approach a symmetrized sum of two draws from a bimodal law).
At $\delta = 8, 16$, \emph{both} constraints are satisfied simultaneously: $\rho \in [0.96, 0.99]$ gives a meaningful spectral gap $2\rho(1-\rho)$, and SIGReg drops to within a few times the floor for both marginals and the joint.
This is precisely the regime where $R^2$ is maximal, as shown by the yellow points clustered in the bottom-middle region of each panel.
The best OU $\rho$ value (vertical dashed line, $\rho = 0.95$) sits near these high-$R^2$ points, confirming that the trajectory condition most closely approaches the theoretical assumptions at intermediate stride.

The joint panel (rightmost) tells a consistent story at slightly higher absolute values: the 2D SIGReg of the transition-difference distribution is always larger than either marginal because it additionally penalizes the cross-dimension dependence that neither marginal can see.
The $R^2$ ordering is the same as in the per-dimension panels.

\begin{figure}[t]
    \centering
    \includegraphics[width=0.6\textwidth]{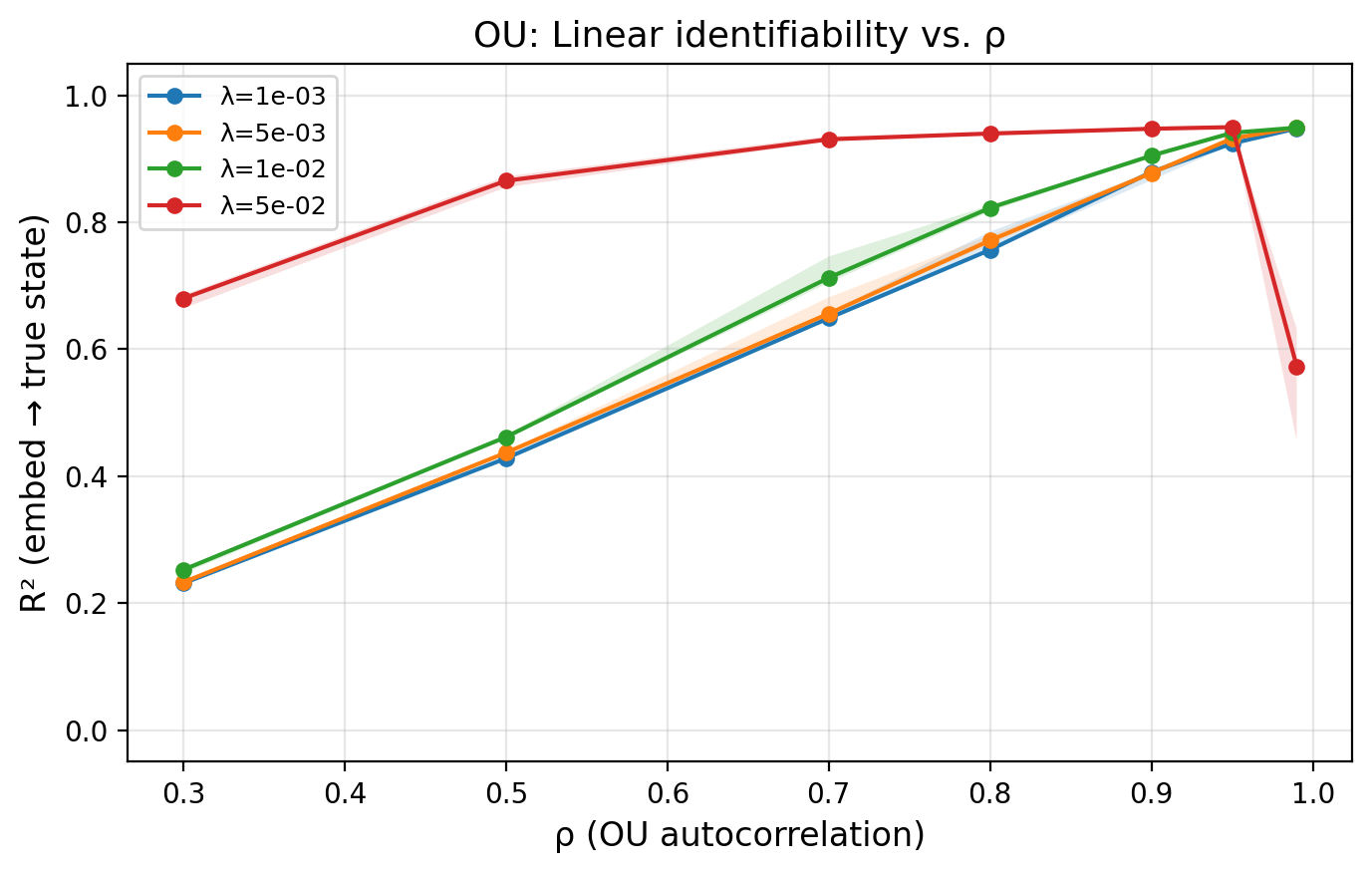}
    \includegraphics[width=0.6\textwidth]{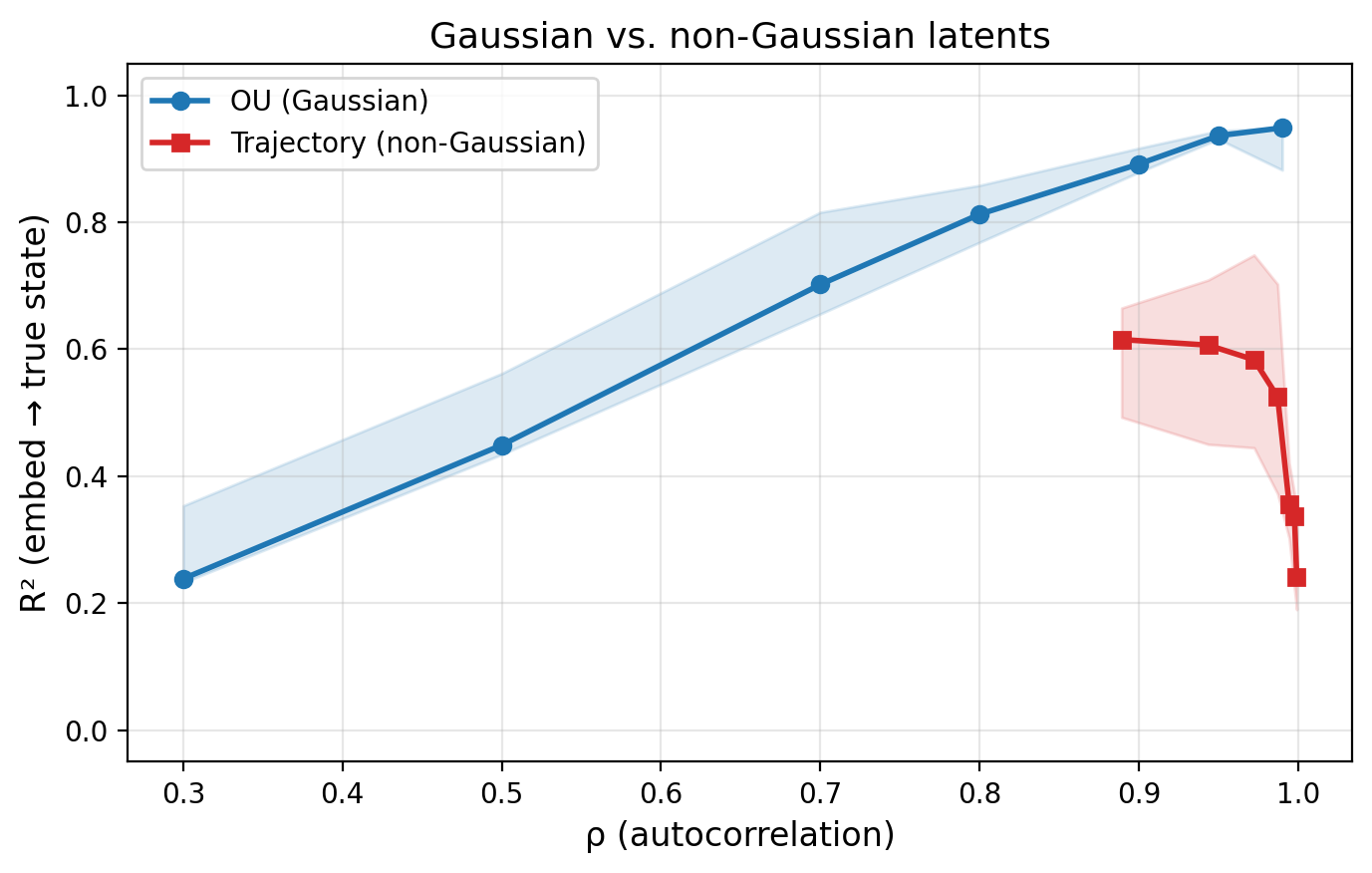}
    \caption{
        \textbf{Left:} OU identifiability vs.\ $\rho$ for different $\lambda$ values. $R^2$ increases monotonically, with all $\lambda$ values converging at high $\rho$. Higher $\lambda$ (stronger SIGReg) helps at low $\rho$ where the alignment signal is weak.
        \textbf{Right:} Gaussian (OU) vs.\ trajectory data at matched autocorrelation $\rho$. At the same $\rho$, Gaussian latents achieve substantially higher $R^2$, directly validating the converse theorem: non-Gaussian marginals reduce identifiability.
    }
    \label{fig:reacher_r2}
\end{figure}

\paragraph{Additional results on OU and Trajectory conditions.}
Fig.~\ref{fig:reacher_r2} directly contrasts the two: at matched $\rho$, Gaussian (OU) latents achieve substantially higher $R^2$ than SAC-policy trajectories (right panel); an empirical instance of Thm.~\ref{thm:converse} on raw pixels. Within OU alone (left), $R^2$ rises monotonically in $\rho$, and stronger $\lambda$ helps only at low $\rho$ where the alignment signal is weak.

\begin{figure}[t]
    \centering
    \includegraphics[width=0.6\textwidth]{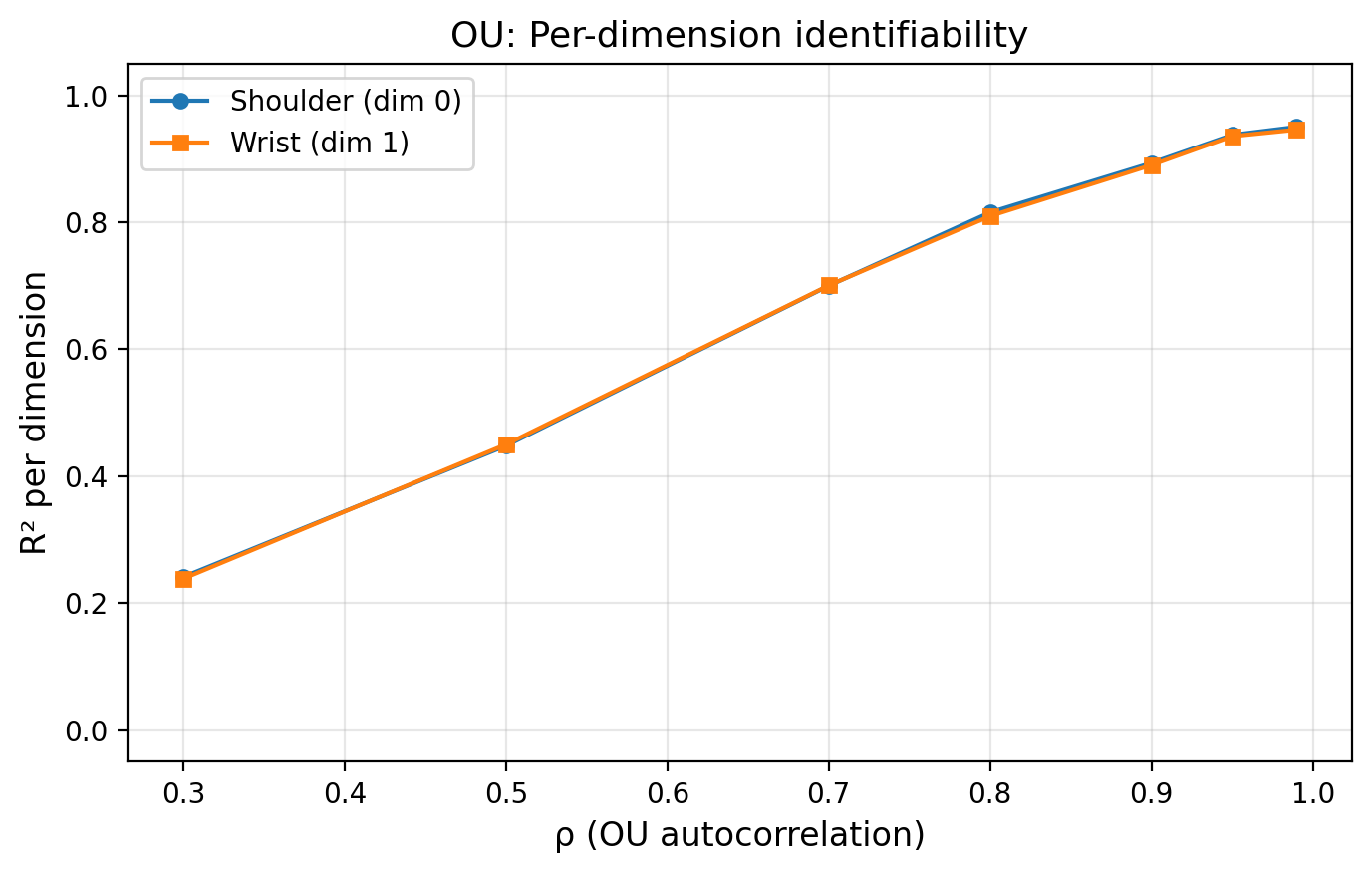}
    \includegraphics[width=0.6\textwidth]{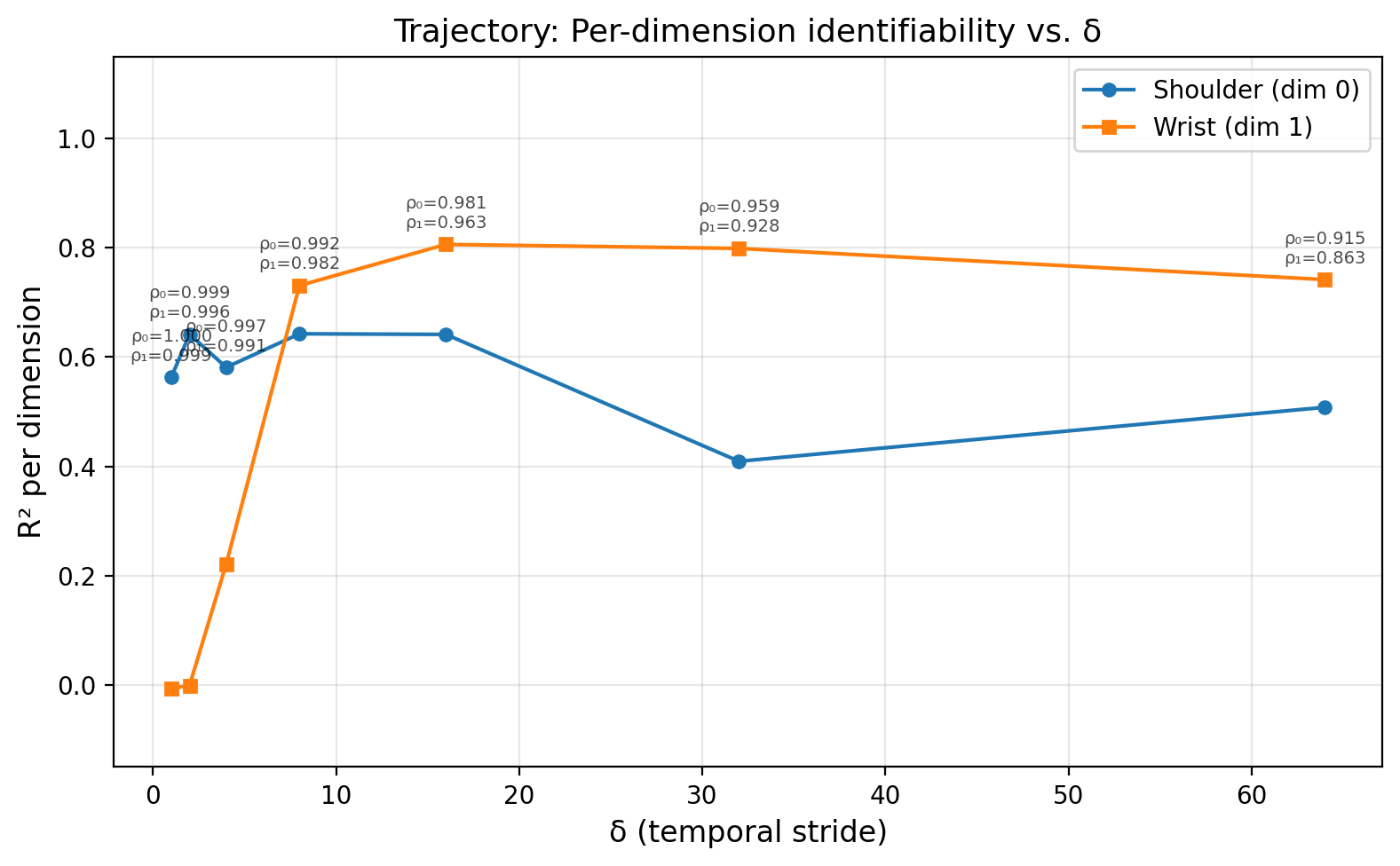}
    \caption{
        \textbf{Per-dimension identifiability.}
        \textbf{Left:} OU condition; shoulder and wrist are recovered symmetrically, consistent with the isotropic transition.
        \textbf{Right:} Trajectory condition; massive asymmetry: the wrist ($R^2 \approx 0$ at $\delta = 1$) recovers only at larger $\delta$ where temporal variation provides learning signal; the shoulder is consistently easier but degrades at large $\delta$ due to wrapping beyond $\pm\pi$.
        Per-dimension $\rho$ values are annotated.
    }
    \label{fig:reacher_appendix_perdim}
\end{figure}

Fig.~\ref{fig:reacher_appendix_perdim} resolves identifiability by joint. OU recovers shoulder and wrist symmetrically, consistent with the isotropy of the Gaussian OU transition (see App.~\ref{app:prior-work} for why isotropic transitions are necessary for the simultaneous approach). Trajectory recovery is massively asymmetric: the wrist, driven by a nearly-bimodal policy-induced marginal and a faster autocorrelation, is recovered poorly at small $\delta$ and improves only once $\delta$ is large enough to provide informative temporal variation; the shoulder degrades at large $\delta$ due to wrapping beyond $\pm\pi$. This is the empirical signature predicted by the Sturm--Liouville analysis when isotropy fails.

\begin{figure}[t]
    \centering
    \includegraphics[width=0.6\textwidth]{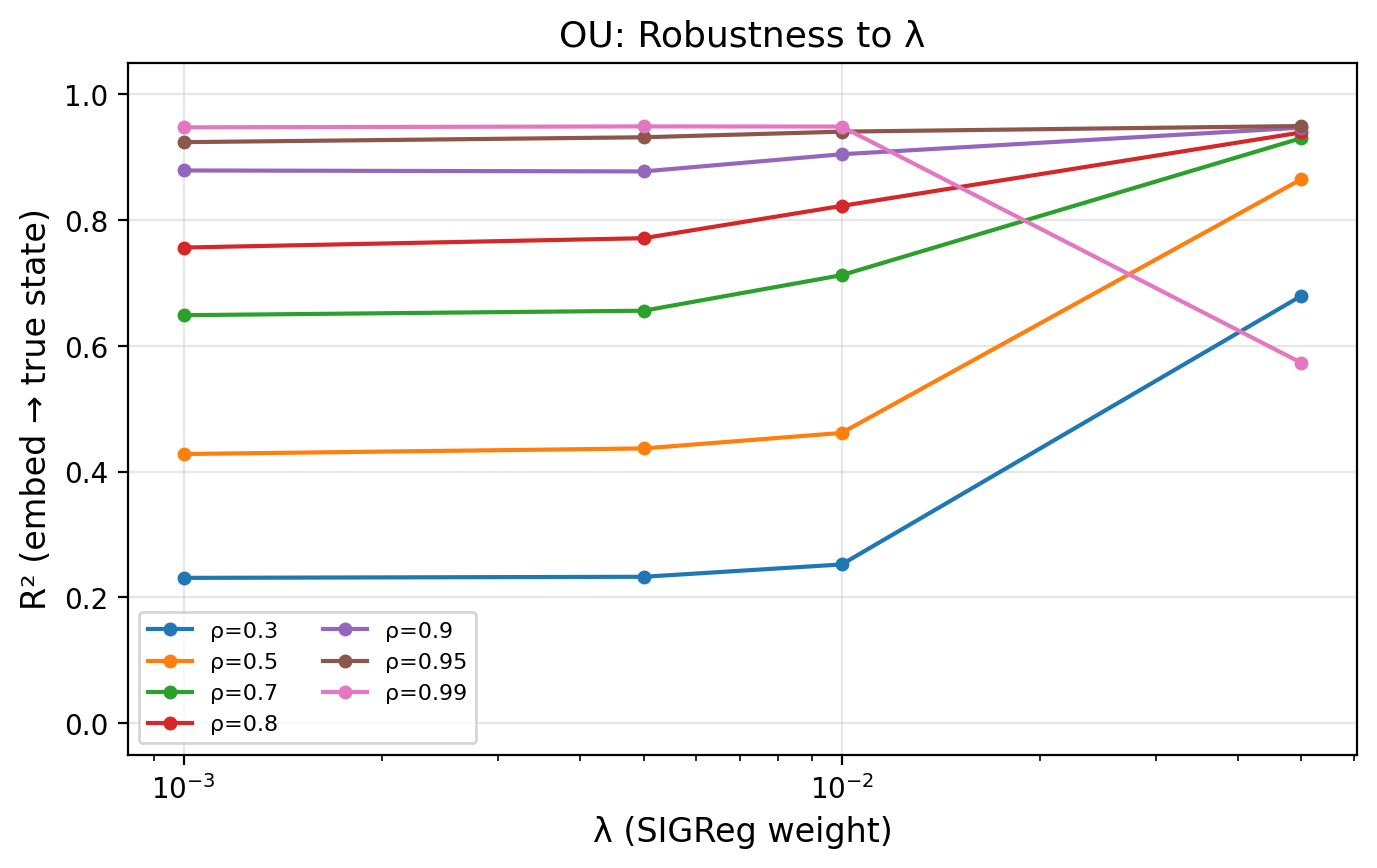}
    \includegraphics[width=0.6\textwidth]{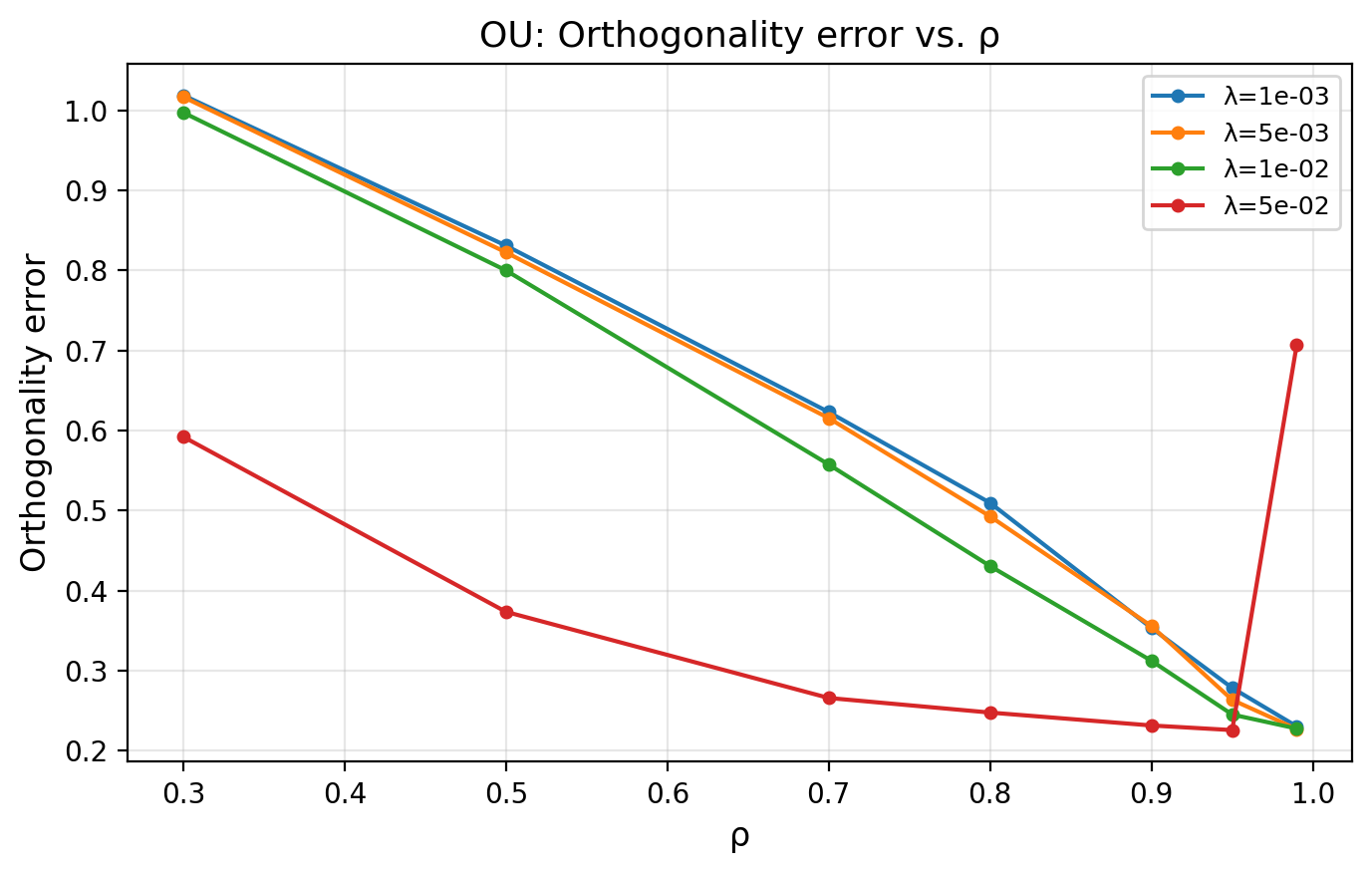}
    \caption{
        \textbf{Left:} $\lambda$ robustness in the OU condition. For high $\rho$, identifiability is stable across $\lambda$; for low $\rho$, stronger regularization ($\lambda = 5 \times 10^{-2}$) compensates for the weak alignment signal. At $\rho = 0.99$, the highest $\lambda$ degrades slightly as SIGReg begins to dominate alignment.
        \textbf{Right:} Orthogonality error decreases monotonically with $\rho$, consistent with the approximate bound (Thm.~\ref{thm:approx}).
    }
    \label{fig:reacher_lambda}
\end{figure}

Fig.~\ref{fig:reacher_lambda} connects back to the approximate bound. \textbf{Left:} OU identifiability is robust to $\lambda$ at high $\rho$; only at $\rho = 0.99$ with the strongest $\lambda$ does SIGReg begin to dominate alignment, mirroring the collapse regime of the grid search (Fig.~\ref{fig:grid}). \textbf{Right:} the fitted orthogonality error $\|\hat Q^\top \hat Q - I\|_F / \sqrt{N}$ decreases monotonically with $\rho$, tracking the tightening spectral gap $2\rho(1-\rho)$ that governs Thm.~\ref{thm:approx}.

\subsection{Planning Experiment}\label{app:planning-exp}

\begin{figure}[t]
    \centering
    \includegraphics[width=\textwidth]{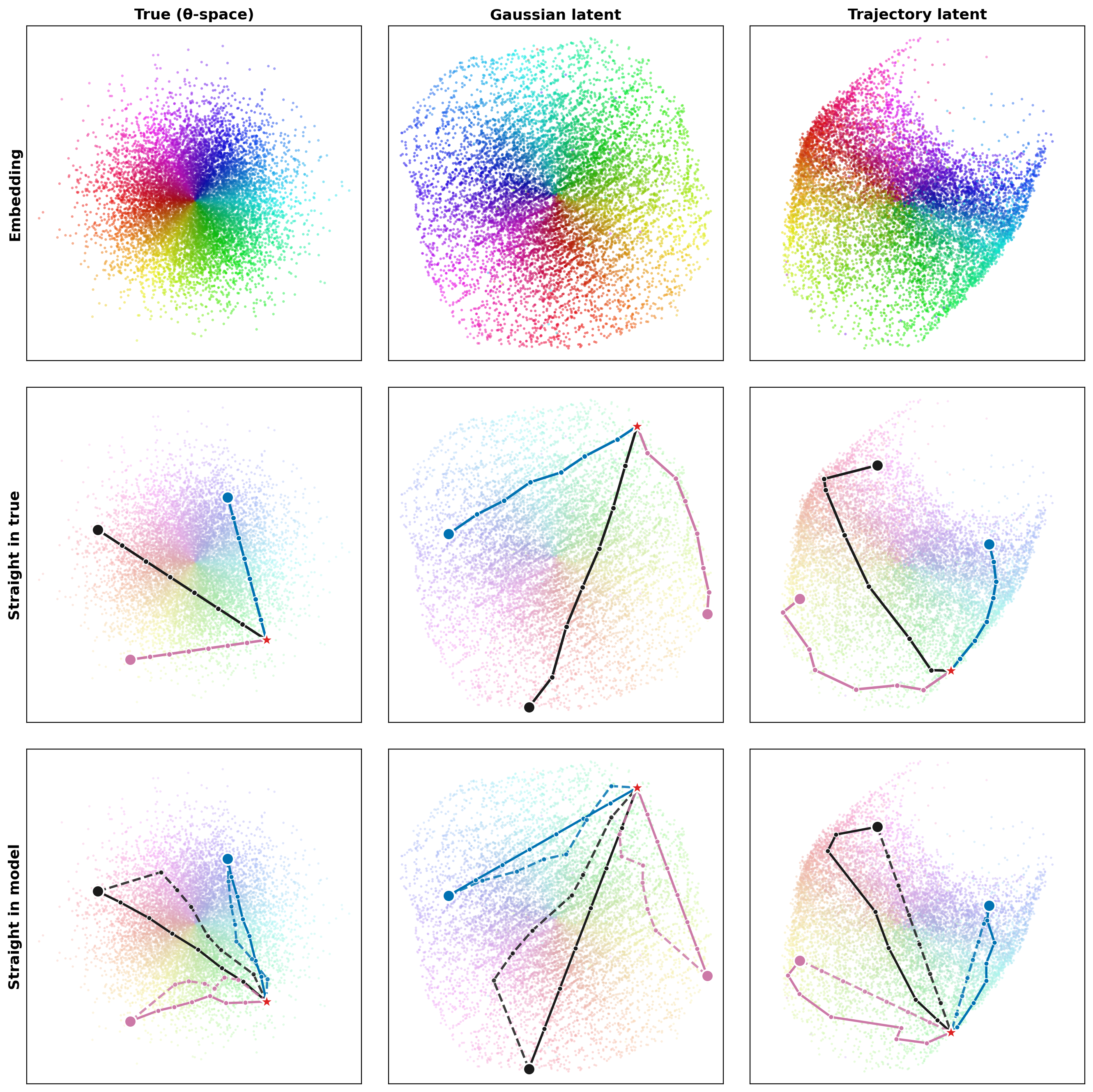}
    \caption{
        \textbf{Planning: embedding and straight-line paths.}
        Columns: true $\theta$-space (left), Gaussian/OU encoder (center), Trajectory encoder (right).
        \textbf{Top row:} scatter of eval-set embeddings, colored by true $\theta$-space polar angle. The Gaussian encoder is an approximate rotation of the true latent; the Trajectory encoder is visibly warped.
        \textbf{Middle row:} three example trajectories that are straight in the \emph{true} joint space, rendered in each representation. They remain approximately straight in the Gaussian embedding but curve in the Trajectory embedding.
        \textbf{Bottom row:} the complementary test; straight lines in each model's latent, decoded back to joint space. Gaussian-encoder plans stay near-straight in $\theta$-space; Trajectory-encoder plans curve.
        Together, the middle and bottom rows confirm Fig.~\ref{fig:planning}: only linear identifiability turns straight-latent plans into straight joint-space paths.
    }
    \label{fig:planning_scatter}
\end{figure}

The planning experiment (Sec.~\ref{sec:planning-exp}, Fig.~\ref{fig:planning}) evaluates straight-line latent interpolation across all trained Reacher encoders (App.~\ref{app:reacher}). The qualitative panels (Fig.~\ref{fig:planning} left) compare the best Gaussian encoder (OU condition at $\rho = 0.99$) against the best Trajectory encoder (stride $\delta = 8$); the quantitative panels (middle, right) aggregate over the full set.

\paragraph{Setup.}
We construct a retrieval library of $10{,}000$ held-out evaluation frames paired with their joint-angle labels. Given a start frame $x_s$ and a goal frame $x_g$, we encode both to latents $y_s = f(x_s)$ and $y_g = f(x_g)$ and form the straight-line latent interpolant $y_t = (1-t)\, y_s + t\, y_g$ for $t \in \{0, \tfrac{1}{T}, \ldots, 1\}$ with $T = 15$. Each $y_t$ is decoded back to a frame by $k$-nearest-neighbor retrieval ($k = 1$, Euclidean distance) against the library's embeddings, yielding a decoded joint-space path $\{\hat{\theta}_t\}$. We evaluate this path against the ideal straight-in-joint-space trajectory connecting $\theta_s$ and $\theta_g$.

\paragraph{Metric.}
We sample $K = 30$ start--goal pairs uniformly from the eval library (shared seed across encoders) and report \textbf{path length}: $\big(\sum_t \|\hat{\theta}_{t+1} - \hat{\theta}_t\|\big) / \|\theta_g - \theta_s\|$, the $\theta$-space arc length of the decoded path divided by the chord length. Ideal value $1$ (straight chord); larger values indicate the encoder warps joint-space geometry. We also track mean orthogonal deviation from the chord as a complementary diagnostic.

\paragraph{Interpretation.}
Under an affine encoder $h(z) = Qz$, the latent straight line maps to the true straight line in joint space and the metric attains its ideal value; deviations measure how much the encoder warps joint-space geometry. Fig.~\ref{fig:planning_scatter} decomposes this geometrically.

\subsection{Misspecified Latent Dimension}\label{app:dims}

\paragraph{Setup.}
The true latent dimension $N$ is fixed and the encoder output dimension $M$ is swept. Since the matched encoder of the scaling experiment is a bijection and cannot change dimension, we use the 4-layer MLP encoder (hidden width 512 for RealNVP with $N{=}8$, 256 for the spiral with $N{=}2$), trained with SIGReg ($\lambda = 10^{-3}$, $\rho = 0.95$, batch size 256, 20{,}000 steps, 5 seeds) and evaluated on 20{,}000 held-out points. For $M \neq N$, recovery $R^2(h \to z)$ regresses the $N$ latents on the $M$-dimensional embedding and encoding $R^2(z \to h)$ the reverse. The participation ratio $\mathrm{PR} = (\sum_i \lambda_i)^2 / \sum_i \lambda_i^2$ is computed from the eigenvalues $\lambda_i$ of the $M \times M$ embedding covariance $C$; equivalently, $\mathrm{PR} = (\mathrm{tr}\,C)^2 / (\varepsilon^2 + 2\,\mathrm{tr}\,C - M)$ with $\varepsilon = \|C - I_M\|_F$.

\begin{figure}[h]
    \centering
    \includegraphics[width=0.85\textwidth]{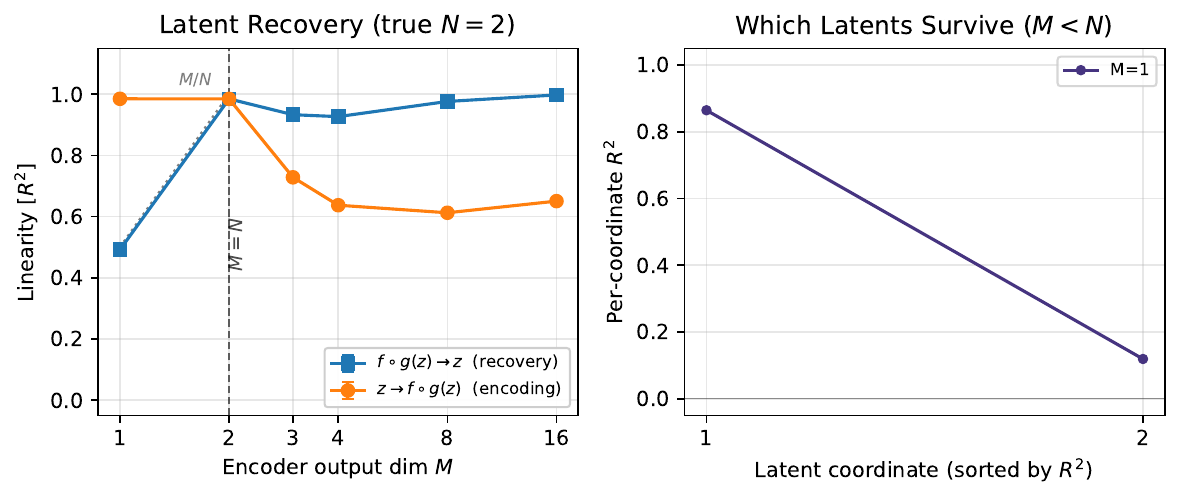}
    \caption{
        \textbf{Misspecified latent dimension, spiral mixing} (true $N{=}2$, 5 seeds; compare Fig.~\ref{fig:dims}).
        \textbf{Left:} Recovery tracks $M/N$ below $N$ and saturates above; unlike RealNVP, encoding $R^2(z \to h)$ drops for $M > N$ because the surplus dimensions carry nonlinear functions of $z$.
        \textbf{Right:} At $M{=}1$, the encoder keeps one latent variable and discards the other.
    }
    \label{fig:dims_spiral}
\end{figure}

\paragraph{Results.}
Tab.~\ref{tab:dims} and Figs.~\ref{fig:dims},~\ref{fig:dims_spiral} give both settings. Below $N$, both settings behave identically: recovery is close to $M/N$ and encoding is near-perfect, so the embedding is an isometric copy of an $M$-dimensional subspace of the latents, and $\mathrm{PR} = M$ exactly because SIGReg fills every output dimension. Above $N$ the two settings differ. On RealNVP the surplus dimensions collapse, $\mathrm{PR}$ stays at $N$ up to $M{=}16$, and recovery and encoding stay near $0.99$; at $M{=}32, 64$ some seeds lose a latent ($\mathrm{PR} < N$, larger seed variance). On the spiral the surplus dimensions are filled with nonlinear functions of $z$: recovery stays high, encoding drops to $0.61$--$0.73$, and $\mathrm{PR}$ keeps growing to about $4$. In both settings, however, $\mathrm{PR} = M$ holds exactly up to $M{=}N$ and fails from $M{=}N{+}1$ on, which is the basis of the estimator in Sec.~\ref{sec:beyond}.

\begin{table}[h]
\centering
\caption{\textbf{Misspecified latent dimension} (mean over 5 seeds; std $\leq 0.01$ unless given). PR: participation ratio of $\mathrm{Cov}(h)$. The last $M$ with $\mathrm{PR} \approx M$ (bold) equals the true $N$ in both settings.}
\label{tab:dims}
\small
\begin{tabular}{r ccc @{\hspace{24pt}} r ccc}
\toprule
\multicolumn{4}{c}{\textbf{RealNVP, $N{=}8$}} & \multicolumn{4}{c}{\textbf{Spiral, $N{=}2$}} \\
\cmidrule(r{12pt}){1-4} \cmidrule(l){5-8}
$M$ & $R^2(h\to z)$ & $R^2(z \to h)$ & PR & $M$ & $R^2(h\to z)$ & $R^2(z \to h)$ & PR \\
\midrule
2  & 0.242 & 0.971 & 2.00 & 1  & 0.492 & 0.985 & 1.00 \\
4  & 0.491 & 0.982 & 4.00 & \textbf{2} & 0.985 & 0.985 & \textbf{2.00} \\
6  & 0.740 & 0.986 & 6.00 & 3  & 0.933 & 0.728 & 2.83 \\
\textbf{8} & 0.987 & 0.987 & \textbf{7.99} & 4 & 0.927 & 0.637 & 3.49 \\
12 & 0.988 & 0.988 & 7.99 & 8  & 0.976 & 0.613 & 3.98 \\
16 & 0.989 & 0.989 & 7.99 & 16 & 0.998 & 0.651 & 3.80 \\
32 & 0.949\tiny{$\pm$0.06} & 0.988 & 7.19\tiny{$\pm$0.75} & & & & \\
64 & 0.977\tiny{$\pm$0.02} & 0.986 & 7.71\tiny{$\pm$0.39} & & & & \\
\bottomrule
\end{tabular}
\end{table}

\subsection{Predictors}\label{app:pred}

\paragraph{Setup.}
LeJEPA's objective (Sec.~\ref{sec:objective}) aligns the two embeddings directly. JEPAs with a predictor instead align $h(z')$ with a learned prediction $P(h(z))$, i.e., $\mathcal{L}_{\mathrm{inv}} = \mathbb{E}\big[\|h(z') - P(h(z))\|^2\big]$, trained jointly with the encoder. We compare three arms that are identical in every other respect: no predictor ($P = \mathrm{id}$, Eq.~\ref{eq:loss-expand}), a linear predictor ($P \in \mathbb{R}^{M \times M}$, no bias, random initialization), and an MLP predictor (two hidden layers of the encoder's width with GELU; $271$k parameters for RealNVP). All arms use positive pairs $(z, z')$ from the OU transition~\eqref{eq:dgp} with $\rho = 0.95$, the 4-layer MLP encoder (hidden width 512 for RealNVP with $N{=}8$, 256 for the spiral with $N{=}2$), SIGReg with $\lambda = 4 \times 10^{-3}$, batch size 256, 20{,}000 steps, and 5 seeds.

\begin{figure}[h]
    \centering
    \includegraphics[width=0.85\textwidth]{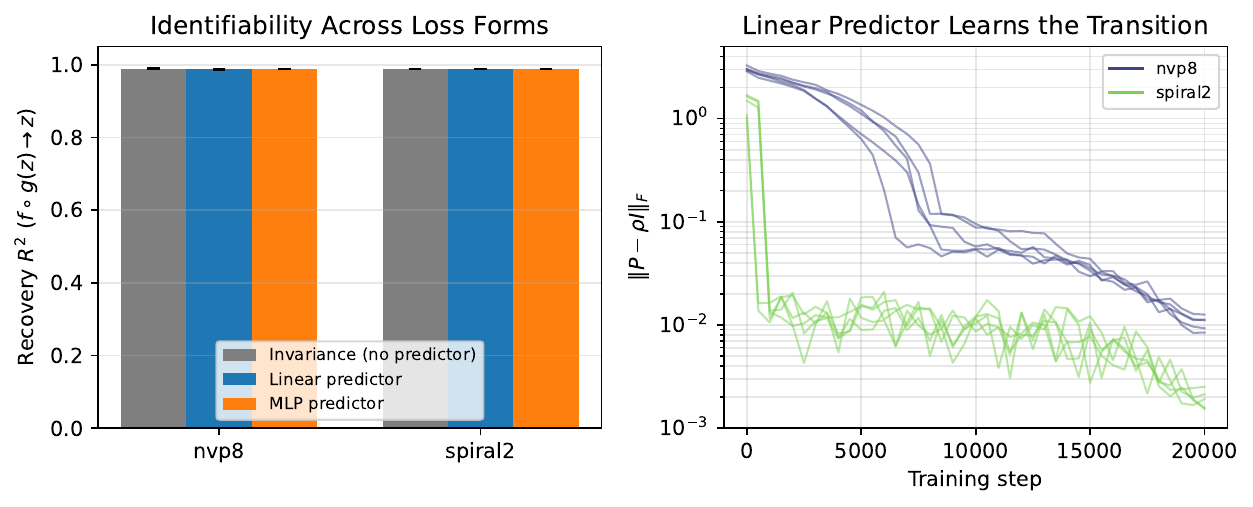}
    \caption{
        \textbf{Predictors.}
        \textbf{Left:} Recovery for the three arms (mean $\pm$ std, 5 seeds).
        \textbf{Right:} Distance of the linear predictor from the true transition operator, $\|P - \rho I\|_F$, during training (lines are seeds).
    }
    \label{fig:pred}
\end{figure}

\paragraph{Results.}
All three arms reach the same recovery in both settings (Tab.~\ref{tab:pred}, Fig.~\ref{fig:pred}). For the linear predictor, the theory suggests a specific solution: at $h(z) = Qz$ with $Q \in O(N)$, the least-squares predictor is $\mathrm{Cov}(h(z'), h(z))\,\mathrm{Cov}(h(z))^{-1} = \rho I$, the true transition operator in the learned coordinates (Thm.~\ref{thm:main}). The learned $P$ converges to it from a random initialization, $\|P - \rho I\|_F = 0.011$ (RealNVP) and $0.002$ (spiral). We do not have a proof that the predictive objective selects $h(z) = Qz$, so this is consistency rather than a theorem. For the MLP predictor no such statement is possible: an expressive predictor can in principle absorb a nonlinear distortion of $h$ without changing the loss. That gradient descent does not find such solutions here is an empirical observation, which a full treatment of action-conditioned predictors \citep{maes2026leworldmodel} would need to explain.

\begin{table}[h]
\centering
\caption{\textbf{Predictors.} Recovery $R^2(h \to z)$, mean $\pm$ std over 5 seeds.}
\label{tab:pred}
\small
\begin{tabular}{l cc}
\toprule
Predictor & RealNVP ($N{=}8$) & Spiral ($N{=}2$) \\
\midrule
None (Eq.~\ref{eq:loss-expand}) & 0.989\tiny{$\pm$0.001} & 0.989\tiny{$\pm$0.000} \\
Linear & 0.987\tiny{$\pm$0.001} & 0.989\tiny{$\pm$0.000} \\
MLP & 0.988\tiny{$\pm$0.001} & 0.988\tiny{$\pm$0.000} \\
\bottomrule
\end{tabular}
\end{table}

\subsection{Training Dynamics}\label{app:dynamics}

\begin{figure}[h]
    \centering
    \includegraphics[width=0.55\textwidth]{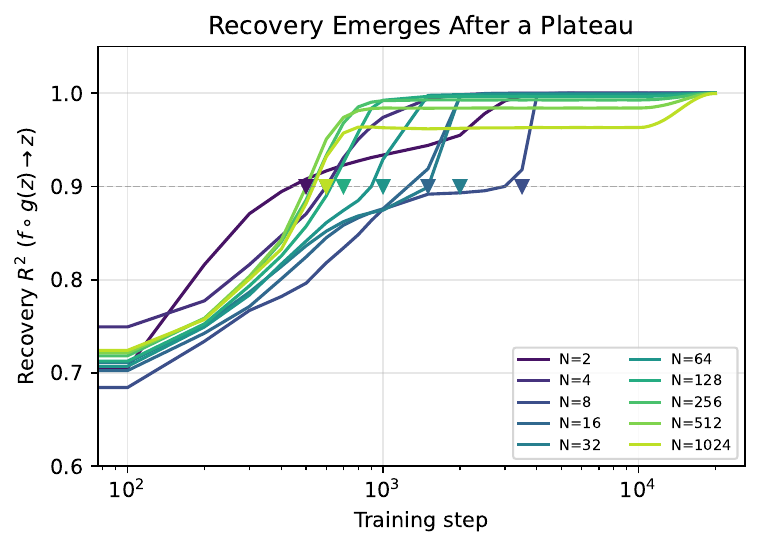}
    \caption{
        \textbf{Recovery during training} (SIGReg scaling runs, seed 0, $N \leq 1024$).
        Triangles mark the escape step, where $R^2(h \to z)$ first exceeds $0.9$ (dashed line).
    }
    \label{fig:dynamics}
\end{figure}

Our theorems characterize the optimum, not the trajectory that reaches it. Fig.~\ref{fig:dynamics} shows recovery during training for the scaling runs (SIGReg, matched encoder, batch size 256). Recovery first rises slowly and then jumps to its final value within a few hundred steps, and Tab.~\ref{tab:dynamics} lists the first step at which $R^2(h \to z) > 0.9$. This escape step is not monotone in $N$: it grows up to $N \approx 8$--$32$, where it also varies most across seeds, and then shortens and becomes nearly seed-independent for $N \geq 128$. At $N{=}4096$ and batch size 256, recovery never crosses $0.9$ within 20{,}000 steps (App.~\ref{app:scaling}). Explaining these dynamics, for instance by tracking per-coordinate recovery during training, is left to future work.

\begin{table}[h]
\centering
\caption{\textbf{Escape step}, the first evaluation step with $R^2(h \to z) > 0.9$ (median and range over 5 seeds; SIGReg scaling runs, batch size 256).}
\label{tab:dynamics}
\resizebox{\textwidth}{!}{%
\begin{tabular}{l ccccccccccc}
\toprule
$N$ & 2 & 4 & 8 & 16 & 32 & 64 & 128 & 256 & 512 & 1024 & 2048 \\
\midrule
median & 500 & 800 & 1500 & 1500 & 1500 & 1000 & 700 & 600 & 600 & 600 & 1000 \\
range & 200--600 & 600--1500 & 900--3500 & 700--2000 & 800--3000 & 700--3000 & 600--800 & 600 & 500--600 & 600 & 1000 \\
\bottomrule
\end{tabular}}
\end{table}

\subsection{ImageNet-X Linear Probes}\label{app:imagenetx}

\paragraph{Encoders.}
LeJEPA, DINO \citep{caron2021emerging}, and iBOT \citep{zhou2022ibot} are ViT-B/16 models trained on ImageNet-1k with a shared codebase and schedule, so that the objective is the only variable among them. DINOv2 \citep{oquab2024dinov2} and DINOv3 \citep{simeoni2025dinov3} (LVD-142M and LVD-1689M, the latter distilled from a 7B teacher) are reference points with two to three orders of magnitude more data and are not part of the matched comparison. Two controls separate the objective from the architecture and the images: the same ViT-B/16 at random initialization (\textit{init}) and a random linear map of the raw pixels (\textit{pixels}).

\paragraph{Analysis space.}
All analyses use 64 dimensions. For LeJEPA this is the output of the projector, where both loss terms attach and therefore the space the theory is about. For every other encoder it is the top 64 principal components of the 768-dimensional CLS embedding; since linear decodability is invariant to invertible linear maps, nothing linearly decodable within this subspace is lost. The subspace is matched in nominal dimension, not in retained information: LeJEPA's backbone has fewer effective dimensions than DINO's or iBOT's, so the projection discards proportionally more from the latter.

\paragraph{Probes.}
We probe the 17 binary factors annotated by ImageNet-X \citep{idrissi2023imagenetx} on 43{,}603 ImageNet validation images, following \citet[Fig.~3]{reizinger2024cross}. For each factor we split 70/30, balance both classes by subsampling, hold out 20\% of the training split for early stopping, and report balanced test accuracy over 5 repetitions. Probes are full-batch AdamW (learning rate $10^{-2}$, weight decay $10^{-3}$, early stopping with patience 300 steps); the MLP probe has one hidden layer of width 256 and is otherwise identical to the linear probe.

\paragraph{Results.}
Tab.~\ref{tab:imagenetx-full} lists all factors. Balanced sample sizes span three orders of magnitude, so per-factor comparisons are only meaningful for the nine factors with $N > 1000$, which give the main-text means (Tab.~\ref{tab:imagenetx}). LeJEPA has the highest accuracy on eight of these nine, both among the matched IN-1k encoders and against DINOv2 and DINOv3; the exception is \textit{smaller}. Both controls stay well below every trained encoder, so the probes do not overfit. MLP probes improve on linear ones by only $0.01$--$0.03$ for every encoder (Tab.~\ref{tab:imagenetx-gauss}); this gap is bounded by the headroom below the Bayes rate of noisy human annotations and supports no cross-model claim.

\paragraph{Gaussianity.}
Tab.~\ref{tab:imagenetx-gauss} reports the sketched Epps--Pulley statistic of LeJEPA \citep[Alg.~1]{balestriero2025lejepa} against $\mathcal{N}(0, I)$, divided by the sample size, and the participation ratio of the covariance, normalized by the dimension. The claim-bearing comparison is LeJEPA's projector output against a true Gaussian sample of the same size, since that is where SIGReg was applied: its distance is $0.0049$, two orders of magnitude above the sampling floor ($3 \times 10^{-5}$) but fifty times below LeJEPA's own backbone ($0.25$), with nearly isotropic covariance (normalized participation ratio $0.91$). SIGReg is therefore met approximately, the regime of Thm.~\ref{thm:approx}. For the other encoders the coordinates are principal components, which Gaussianize marginals by the central limit theorem regardless of the underlying distribution, so their statistics are not directly comparable.

\begin{table}[h]
\centering
\caption{\textbf{ImageNet-X linear probe accuracy, all 17 factors} (balanced accuracy, mean over 5 repetitions; chance $0.5$). $N$: balanced sample size. Bold: above chance at a Bonferroni-corrected $p < 0.05$ over all factors and models. Factors above the line ($N > 1000$) enter the main-text mean.}
\label{tab:imagenetx-full}
\small
\setlength{\tabcolsep}{4pt}
\begin{tabular}{l r ccc cc cc}
\toprule
 & & \multicolumn{3}{c}{IN-1k, matched} & \multicolumn{2}{c}{reference} & \multicolumn{2}{c}{controls} \\
\cmidrule(lr){3-5} \cmidrule(lr){6-7} \cmidrule(lr){8-9}
Factor & $N$ & LeJEPA & DINO & iBOT & DINOv2 & DINOv3 & init & pixels \\
\midrule
class            & 43500 & \textbf{0.865} & \textbf{0.666} & \textbf{0.741} & \textbf{0.856} & \textbf{0.813} & \textbf{0.620} & \textbf{0.585} \\
shape            & 1276  & \textbf{0.737} & 0.590 & \textbf{0.661} & \textbf{0.709} & \textbf{0.657} & 0.552 & 0.575 \\
subcategory      & 1118  & \textbf{0.720} & \textbf{0.604} & \textbf{0.648} & \textbf{0.682} & 0.614 & 0.554 & 0.502 \\
pattern          & 11354 & \textbf{0.674} & \textbf{0.593} & \textbf{0.618} & \textbf{0.658} & \textbf{0.653} & 0.536 & 0.536 \\
smaller          & 2738  & \textbf{0.651} & \textbf{0.618} & \textbf{0.663} & \textbf{0.627} & \textbf{0.692} & 0.529 & 0.538 \\
partial view     & 1208  & \textbf{0.629} & 0.543 & 0.592 & 0.595 & 0.606 & 0.507 & 0.506 \\
color            & 11642 & \textbf{0.616} & \textbf{0.571} & \textbf{0.580} & \textbf{0.599} & \textbf{0.606} & 0.544 & 0.526 \\
pose             & 28404 & \textbf{0.605} & \textbf{0.559} & \textbf{0.583} & \textbf{0.598} & \textbf{0.589} & 0.530 & 0.516 \\
background       & 27898 & \textbf{0.594} & \textbf{0.559} & \textbf{0.570} & \textbf{0.576} & \textbf{0.578} & \textbf{0.532} & 0.517 \\
\midrule
darker           & 230   & 0.809 & 0.714 & 0.751 & 0.643 & 0.620 & 0.700 & 0.520 \\
person blocking  & 112   & 0.800 & 0.647 & \textbf{0.771} & 0.671 & 0.712 & 0.482 & 0.588 \\
texture          & 542   & \textbf{0.664} & 0.619 & 0.643 & \textbf{0.698} & 0.625 & 0.548 & 0.521 \\
larger           & 284   & 0.614 & 0.551 & 0.549 & 0.493 & 0.558 & 0.528 & 0.493 \\
style            & 84    & 0.600 & 0.562 & 0.638 & 0.500 & 0.608 & 0.438 & 0.500 \\
brighter         & 84    & 0.592 & 0.523 & 0.485 & 0.492 & 0.523 & 0.562 & 0.562 \\
multiple objects & 76    & 0.555 & 0.445 & 0.627 & 0.491 & 0.536 & 0.473 & 0.491 \\
object blocking  & 152   & 0.552 & 0.565 & 0.435 & 0.574 & 0.513 & 0.491 & 0.474 \\
\midrule
mean ($N > 1000$) &      & \textbf{0.677} & 0.589 & 0.628 & 0.656 & 0.645 & 0.545 & 0.533 \\
mean (all 17)     &      & \textbf{0.663} & 0.584 & 0.621 & 0.615 & 0.618 & 0.537 & 0.526 \\
\bottomrule
\end{tabular}
\end{table}

\begin{table}[h]
\centering
\caption{\textbf{ImageNet-X probe summary and Gaussianity diagnostics} in the 64-dimensional analysis space. EP: sketched Epps--Pulley distance to $\mathcal{N}(0, I)$ divided by the sample size (lower is more Gaussian). PR$/K$: participation ratio of the covariance divided by the dimension ($1$ for isotropic). For LeJEPA the coordinates are the projector output; for all other rows, principal components.}
\label{tab:imagenetx-gauss}
\small
\begin{tabular}{l ccc cc c}
\toprule
 & Linear & MLP & Gap & EP & PR$/K$ & Excess kurtosis \\
\midrule
LeJEPA (projector)          & 0.663 & 0.684 & 0.021 & 0.0049  & 0.91 & 0.39 \\
LeJEPA backbone (768-d)     & --    & --    & --    & 0.2534  & 0.08 & 0.38 \\
DINO                        & 0.584 & 0.617 & 0.033 & 0.5330  & 0.99 & 0.04 \\
iBOT                        & 0.621 & 0.632 & 0.011 & 0.6083  & 0.98 & 0.03 \\
DINOv2                      & 0.615 & 0.645 & 0.030 & 0.4712  & 0.92 & 0.82 \\
DINOv3                      & 0.618 & 0.643 & 0.025 & 0.0033  & 0.94 & 0.65 \\
init                        & 0.537 & 0.554 & 0.017 & 0.5311  & 0.06 & 1.18 \\
pixels                      & 0.526 & 0.540 & 0.014 & 0.5874  & 0.14 & 0.99 \\
\midrule
$\mathcal{N}(0, I)$ sample  & --    & --    & --    & 0.00003 & 1.00 & 0.00 \\
\bottomrule
\end{tabular}
\end{table}